%% file: main.tex
\documentclass{article}

\usepackage{microtype}
\usepackage{graphicx}
\usepackage{subcaption}
\usepackage{booktabs} % for professional tables

\usepackage{amsthm}

\usepackage{url}
\usepackage{multirow}
\usepackage[table,xcdraw]{xcolor}
\usepackage[normalem]{ulem}
\useunder{\uline}{\ul}{}
\usepackage{makecell}            
\usepackage{array}               
\usepackage{amsmath} 

\usepackage{mathtools}
\usepackage{arydshln}
\usepackage{wrapfig,lipsum}
\usepackage{enumitem}

\usepackage{hyperref} 
\hypersetup
{
    colorlinks = true,       % 启用彩色链接（false时为带框链接）
    citecolor  = blue!60!black,  
    urlcolor   = cyan!60!black,  
    linkcolor  = blue!60!black,  
    breaklinks = true        % 允许长URL自动换行
}

\usepackage[accepted]{icml2026}

\usepackage{amsmath}
\usepackage{amssymb}
\usepackage{mathtools}
\usepackage{amsthm}

\usepackage{enumitem}

\usepackage[capitalize,noabbrev]{cleveref}

\theoremstyle{plain}
\newtheorem{theorem}{Theorem}[section]

\newtheorem{lemma}[theorem]{Lemma}
\newtheorem{corollary}[theorem]{Corollary}
\theoremstyle{definition}
\newtheorem{definition}[theorem]{Definition}

\theoremstyle{remark}
\newtheorem{remark}[theorem]{Remark}

\usepackage[textsize=tiny]{todonotes}

\icmltitlerunning{Calibrating Uncertainty for Zero-Shot Adversarial CLIP}

\begin{document}

\twocolumn[
  \icmltitle{Calibrating Uncertainty for Zero-Shot Adversarial CLIP}




  \begin{icmlauthorlist}
    \icmlauthor{Wenjing Lu}{sjtu,rikenaip}
    \icmlauthor{Zerui Tao}{rikenaip}
    \icmlauthor{Yuning Qiu}{rikenaip}
    \icmlauthor{Dongping Zhang}{rikenaip,gdut}
    \icmlauthor{Yang Yang}{sjtu}
    \icmlauthor{Qibin Zhao}{rikenaip}
  \end{icmlauthorlist}

  \icmlaffiliation{sjtu}{AGI Institute, School of Computer Science, Shanghai Jiao Tong University, Shanghai, China.}
  \icmlaffiliation{rikenaip}{RIKEN AIP, Tokyo, Japan}
  \icmlaffiliation{gdut}{School of Automation, Guangdong University of Technology, Guangzhou, China}

  \icmlcorrespondingauthor{Yang Yang}{yangyang@cs.sjtu.edu.cn}
  \icmlcorrespondingauthor{Qibin Zhao}{qibinzhao@riken.jp}

  \icmlkeywords{Machine Learning, ICML}

  \vskip 0.3in
]



\printAffiliationsAndNotice{}  % no special notice (required even if empty)

\begin{abstract}
CLIP delivers strong zero-shot classification but remains highly vulnerable to adversarial attacks. Prior adversarial fine-tuning work primarily matches predicted logits between clean and adversarial examples, which overlooks uncertainty calibration and may degrade the zero-shot generalization. 
A common expectation in reliable uncertainty estimation is that predictive uncertainty should increase as inputs become more difficult or shift away from the training distribution. However, we frequently observe the opposite in the adversarial setting: perturbations not only degrade accuracy but also suppress uncertainty, leading to severe miscalibration and over-confidence. This reveals a critical reliability gap beyond robustness. To bridge this gap, we propose an adversarial fine-tuning objective for CLIP considering both accuracy and uncertainty. By reparameterizing CLIP outputs as the concentration parameters of a Dirichlet distribution, we propose a unified representation that captures relative semantic structure and confidence magnitude. 
% Our objective aligns these distributions holistically under perturbations,
This enables holistic distribution alignment under perturbations,
moving beyond single-logit anchoring and restoring calibrated uncertainty. 
Experiments across multiple zero-shot benchmarks demonstrate that our method significantly improves uncertainty calibration and achieves competitive adversarial robustness while preserving clean accuracy.
% Code is available at \href{https://github.com/VivienLu/UCAT}{https://github.com/VivienLu/UCAT}.
\end{abstract}

\input{main/1_intro}

\input{main/2_relatedwork}
\input{main/3_pre}
\input{main/4-Prove}
\input{main/5_loss}
\input{main/6_exp}
\input{main/7_con}

\bibliography{ref}
\bibliographystyle{icml2026}

%%%%%%%%%%%%%%%%%%%%%%%%%%%%%%%%%%%%%%%%%%%%%%%%%%%%%%%%%%%%%%%%%%%%%%%%%%%%%%%
%%%%%%%%%%%%%%%%%%%%%%%%%%%%%%%%%%%%%%%%%%%%%%%%%%%%%%%%%%%%%%%%%%%%%%%%%%%%%%%
% APPENDIX
%%%%%%%%%%%%%%%%%%%%%%%%%%%%%%%%%%%%%%%%%%%%%%%%%%%%%%%%%%%%%%%%%%%%%%%%%%%%%%%
%%%%%%%%%%%%%%%%%%%%%%%%%%%%%%%%%%%%%%%%%%%%%%%%%%%%%%%%%%%%%%%%%%%%%%%%%%%%%%%
\newpage
\appendix
\onecolumn
% % \input{Appendix/1-RelatedWork}
\input{Appendix/2-Imp}
\input{Appendix/3-Provement}
\input{Appendix/4-PU}
\input{Appendix/5-Exp}
% % \input{Appendix/6-Limitations}

%%%%%%%%%%%%%%%%%%%%%%%%%%%%%%%%%%%%%%%%%%%%%%%%%%%%%%%%%%%%%%%%%%%%%%%%%%%%%%%
%%%%%%%%%%%%%%%%%%%%%%%%%%%%%%%%%%%%%%%%%%%%%%%%%%%%%%%%%%%%%%%%%%%%%%%%%%%%%%%

\end{document}

%% file: main/1_intro.tex
\section{Introduction}

\begin{figure}[t]
    \centering
    \begin{subfigure}[htbp]{0.95\linewidth}
        \includegraphics[width=\linewidth]{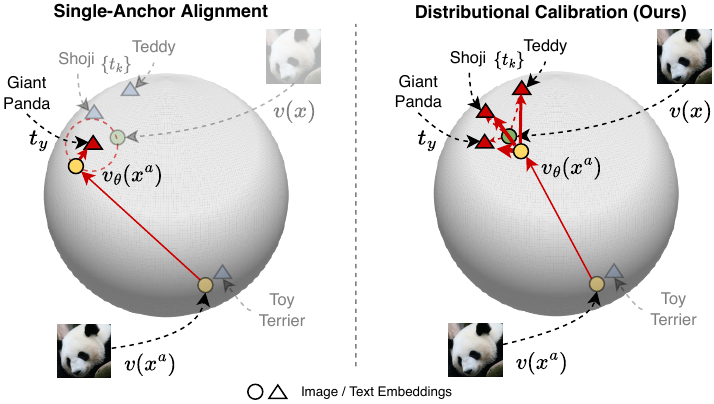}
        \caption{Single-anchor alignment vs. distributional calibration.}
        \label{fig:scheme}
    \end{subfigure}
    \begin{subfigure}[htbp]{0.95\linewidth}
        \includegraphics[width=\linewidth]{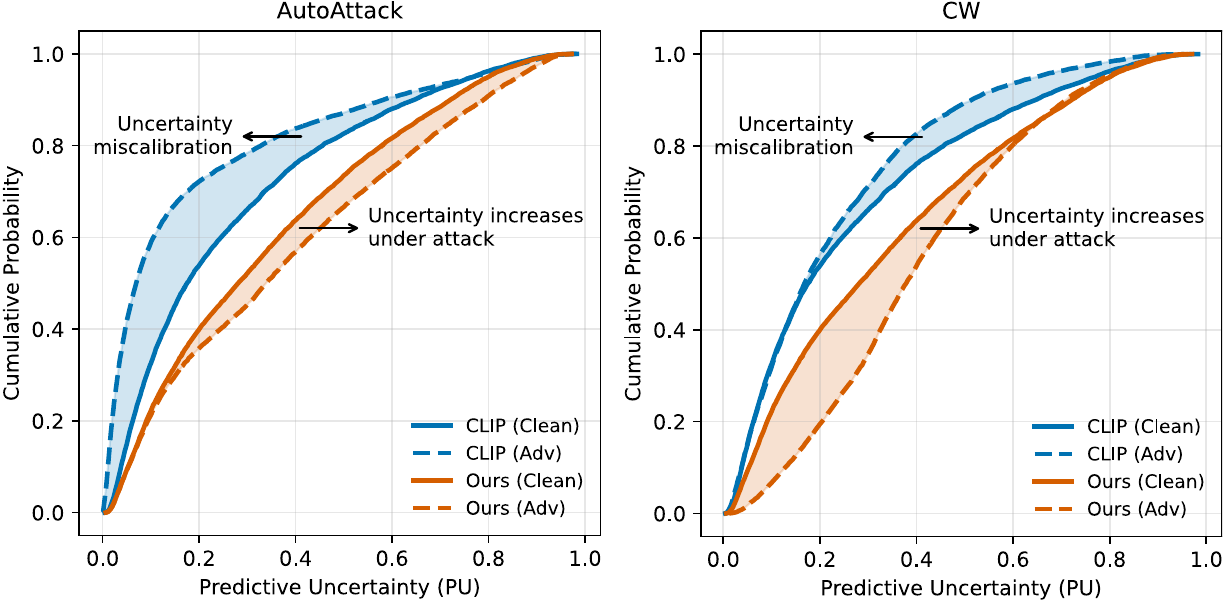}
        \caption{Predictive uncertainty CDFs under AutoAttack and CW.}
        \label{fig:pu_bar}
    \end{subfigure}
    \caption{\textbf{Distributional alignment and uncertainty miscalibration.} (a) Prior ZSAR methods pull an adversarial image feature $v(x^{a})$ toward the text prototype $t_y$ (\emph{single-anchor}), without explicitly preserving its relative similarities to other text prototypes $\{t_k\}$.  Our approach aligns predictive \emph{distributions}
    to preserve inter-class semantics and evidence strength.
    (b) Cumulative distribution functions (CDFs) of predictive uncertainty on CIFAR-10 for clean (solid) and adversarial (dashed) samples under AutoAttack and CW ($\epsilon=1/255$), comparing CLIP (blue) vs. Ours (orange). 
    % Full results on 16 datasets are in Appendix~\ref{app:puaueu}.
    }
    \label{fig:intro}
\end{figure}

%%%%%%%%%%%%%%%%%% Background

Contrastive language-image pretraining (CLIP)~\citep{radford2021learning} has become a widely adopted vision–language model, achieving strong zero-shot recognition by comparing image features with text prompts in a shared embedding space. Its scalability~\citep{jia2021scaling} and adaptability through prompting or ensembling~\citep{zhou2022learning,wortsman2022robust} have established it as a foundation model for open-world scenarios where labeled data are scarce. Although CLIP demonstrates impressive generalization ability, it is highly vulnerable to adversarial attacks: tiny pixel-level perturbations, often imperceptible to humans, can cause confident misclassifications and severe drops in performance~\citep{goodfellow2014explaining,kurakin2018adversarial,madry2017towards}. This contrast between strong zero-shot generalization and fragile robustness motivates the study of adversarial reliability in vision–language models.

Recent efforts on \emph{zero-shot adversarial robustness} (ZSAR) aim to enhance CLIP’s resistance to adversarial perturbations while preserving zero-shot generalization~\citep{mao2022understanding,schlarmann2024robust,xing2025clip,zhang2025clipure}.  
Formally, the task assumes that only the image encoder is adversarially fine-tuned, while the text encoder remains fixed and provides stable semantic anchors. 
Existing methods fine-tune the attacked encoder on labeled data to balance clean accuracy and adversarial robustness, and then evaluate transferability to unseen zero-shot datasets~\citep{yu2024text,wang2024pre,li2024language}. 
A common strategy is to align adversarial features directly to the ground-truth text embedding, which provides strong discriminative supervision but disregards the relative geometry among neighboring classes.
As illustrated in Figure~\ref{fig:scheme}, the adversarial alignment is enforced only toward the ground-truth text embedding, pulling features along an unconstrained direction and disregarding the relative geometry of neighboring embeddings.
However, these relations are essential as they encode inherent data ambiguity, such as semantic overlap between categories or the presence of multiple objects within a single image. Such ambiguity can be naturally interpreted as a form of predictive \textit{uncertainty}. This single-anchor alignment provides strong discriminative supervision but neglects the underlying uncertainty structure, which can limit generalization under adversarial perturbations.

While previous methods mostly focus on aligning the predicted logits, we argue that they overlook an essential phenomenon, that is, a systematic miscalibration in CLIP’s predictive uncertainty under adversarial perturbations.
% Figure~\ref{fig:pu_bar} compares entropy-based uncertainty on clean (solid) and adversarial (striped) inputs across multiple datasets. Strikingly, in many cases, the uncertainty of adversarial predictions is lower than that of clean predictions, 
Figure~\ref{fig:pu_bar} illustrates this issue on a representative benchmark, where CLIP can exhibit \emph{lower} predictive uncertainty on adversarially perturbed inputs than on their clean counterparts. We observe the same trend across multiple datasets (see Appendix~\ref{app:puaueu} for full results), 
challenging the widely held expectation that uncertainty should increase with input difficulty or distributional shift~\citep{guo2017calibration,hendrycks2016baseline,ovadia2019can}. This anomaly indicates that CLIP not only fails to maintain robustness but also produces spuriously confident predictions when attacked. Such behavior highlights a critical reliability gap beyond accuracy, underscoring the need to calibrate uncertainty in adversarial fine-tuning.

To address both the structural and calibration issues, we propose an Uncertainty-Calibrated Adversarial fine-Tuning (UCAT) framework for CLIP. UCAT operates by regularizing entire Dirichlet distributions rather than anchoring to a single class, thereby preserving inter-class semantic relations while calibrating the overall strength of predictive evidence. This is achieved by reparameterizing CLIP’s logits as concentration parameters of a Dirichlet distribution, yielding a unified representation for holistic alignment under perturbations. The quantitative effect of UCAT is shown in Figure~\ref{fig:pu_bar}: compared to vanilla CLIP, our fine-tuned model achieves calibrated uncertainty levels, restoring a consistent ordering: \textit{original CLIP w/ clean img. $<$ fine-tuned CLIP w/ clean img. $<$ fine-tuned CLIP w/ adversarial img.}, which faithfully reflects increasing input difficulty.
The main contributions of this work can be summarized as follows:
% \begin{itemize}\itemsep0pt \parsep0pt \topsep2pt \partopsep0pt
% \begin{enumerate}[leftmargin=*, label=\arabic*), itemsep=0pt, parsep=0pt]
\begin{enumerate}[label=\arabic*), leftmargin=*, nosep, topsep=2pt]
    \item[1)] \textbf{Dirichlet-based formulation of CLIP. }We reformulate CLIP's logits as concentration parameters of a Dirichlet distribution, providing a theoretically justified and closed-form approach to estimate predictive uncertainty.
    \item[2)] \textbf{Uncertainty-Calibrated Adversarial fine-Tuning (UCAT). }We propose a novel uncertainty-calibrated adversarial fine-tuning method that regularizes entire Dirichlet distributions to jointly preserve inter-class relations and calibrate evidence strength.
    \item[3)] \textbf{Extensive empirical validation. }Across 16 single-label benchmarks and the multi-label dataset MS-COCO, we show that our method effectively calibrates uncertainty under attack while maintaining strong clean accuracy and competitive adversarial robustness.
% \end{itemize}
\end{enumerate}

%% file: main/2_relatedwork.tex
\section{Related Work}
\noindent\textbf{From closed-set adversarial training to open-vocabulary VLM robustness.}
Classical adversarial training (AT) targets \emph{closed-set} classification with an explicit classifier head and labeled supervision. Typical objectives jointly encourage (i) \emph{clean discriminability} and (ii) \emph{local robustness} around labeled samples, commonly via first-order min--max formulations~\citep{goodfellow2014explaining,kurakin2018adversarial,madry2017towards} or principled trade-off such as TRADES~\citep{zhang2019theoretically}. Subsequent work improves robustness and efficiency through reweighting~\citep{wang2019improving}, faster inner maximization~\citep{zhang2019you,shafahi2019adversarial,wong2020fast}, weight perturbations~\citep{wu2020adversarial}, robust overfitting analyses~\citep{rice2020overfitting}, and optimization refinements~\citep{pang2019rethinking,addepalli2022efficient,cui2024decoupled}.

For \emph{open-vocabulary} vision--language models (VLMs), large-scale image--text contrastive pretraining already provides strong zero-shot recognition via image--text matching with prompts~\citep{radford2021learning,jia2021scaling,zhou2022learning}. This makes the focus different from closed-set AT: the goal is not to learn a task-specific classifier for a fixed label space, but to improve robustness while keeping zero-shot transfer. Therefore, robustness is usually pursued by \emph{adapting the pretrained model on a limited seen dataset} while \emph{preserving the cross-modal geometry} that underpins transfer to unseen classes and datasets~\citep{wortsman2022robust,mao2022understanding}.

\noindent\textbf{Zero-shot adversarial robustness for VLMs.}
Recent work improves VLM robustness under zero-shot evaluation via adversarial fine-tuning of the image encoder~\citep{mao2022understanding,schlarmann2024robust,wang2024pre,yu2024text,li2024language,dong2025improving,dong2025robustifying}, prompt tuning~\citep{li2024one,shu2022test,sheng2025r,wang2025tapt}, and training-free test-time defenses~\citep{xing2025clip,tong2025zero,zhang2025clipure}.
Many training-time objectives adopt a \emph{single-anchor} design that mainly enforces robustness toward the ground-truth text prototype and treats other classes as negatives~\citep{schlarmann2024robust,wang2024pre,yu2024text,dong2025robustifying}, potentially under-exploiting relations among semantically related classes important for open-vocabulary evaluation.

Several methods align the clean and adversarial \emph{softmax distributions} over the seen label set~\citep{,wang2024pre,dong2025improving}, which mainly constrains \emph{relative} class preferences while largely removing absolute logit-scale effects.
However, calibration studies suggest that scale-related factors affect the reliability of zero-shot inference in open-vocabulary VLMs~\citep{levine2023enabling,murugesan2024robust}, and open-set reliability often relies on confidence scores derived from \emph{absolute} CLIP similarity values~\citep{esmaeilpour2022zero}. Motivated by this, we model logits as Dirichlet evidence to calibrate uncertainty and preserve both relative semantics and logit scale under attack.

\noindent\textbf{Uncertainty calibration.} 
Uncertainty calibration under distribution shift and adversarial perturbations has been widely studied~\citep{guo2017calibration,ovadia2019can}. Evidential formulations provide a distributional view of predictions and can encourage high uncertainty on adversarial or OOD inputs~\citep{malinin2018predictive,malinin2019reverse,ulmer2021prior}, while confidence-calibrated adversarial training explicitly regularizes confidence in closed-set settings~\citep{stutz2020confidence}.
Building on the discussion above that logit scale carries useful reliability signals in open-vocabulary VLMs, we adopt an evidential view: a Dirichlet prediction separates the \emph{relative} class structure (the shape of the Dirichlet, often associated with aleatoric uncertainty) from the \emph{overall} evidence strength (often associated with epistemic uncertainty)~\citep{sensoy2018evidential,malinin2018predictive,ulmer2021prior,ma2025estimating}. We leverage this decomposition to formulate a scale-aware alignment objective that couples distributional semantics with evidence strength, yielding calibrated confidence under attack.

%% file: main/3_pre.tex
\section{Preliminary}
% \subsection{Contrastive Learning Objective and Zero-shot Classification}
\subsection{Zero-Shot Vision-Language Models}
\noindent\textbf{Contrastive pretraining objective.}
% \paragraph{Contrastive pretraining objective}
Contrastive pretraining underlies large-scale vision–language models such as CLIP~\citep{radford2021learning}.
Let $f_{\theta}:\mathcal{X}_{\mathrm{img}}\!\to\!\mathbb{R}^d$, $g_{\phi}:\mathcal{X}_{\mathrm{txt}}\!\to\!\mathbb{R}^d$ denote the image and text encoders, where $d$ is the dimension of the embedding space. For an image–text pair $(x_i^{\mathrm{img}},x_i^{\mathrm{txt}})$, the embeddings are normalized onto the unit hypersphere $\mathbb{S}^{d-1}$:$
    v_i = f_{\theta}(x_i^{\mathrm{img}})/\|f_{\theta}(x_i^{\mathrm{img}})\|_2, 
    \quad
    t_i = g_{\phi}(x_i^{\mathrm{txt}})/\|g_{\phi}(x_i^{\mathrm{txt}})\|_2.
$
The similarity between image $i$ and text $j$ can be expressed in two directional forms: $\ell_{ij}^{v\to t} = \langle v_i, t_j \rangle/\tau$, $\ell_{ij}^{t\to v} = \langle t_i, v_j \rangle/\tau$, where $\tau>0$ is a learnable temperature parameter.
Given a batch of $N$ aligned pairs, the symmetric InfoNCE objective is
\begin{equation}\label{loss:infonce}
\mathcal{L}_{\mathrm{InfoNCE}}
=
-\frac{1}{2N}
\sum_{i=1}^N
\sum_{d\in\{v\to t,\,t\to v\}}
\log
\frac{\exp(\ell_{ii}^{d})}
{\sum_{j=1}^N \exp(\ell_{ij}^{d})}.
\end{equation}

\noindent\textbf{Zero-shot Classification.}
% \paragraph{Zero-shot Classification}
% Benefiting from its self-supervised contrastive learning objective, CLIP exhibits strong zero-shot transfer capability for open-vocabulary recognition~\citep{jia2021scaling,yao2021filip,zhai2022lit,zhou2022learning}. At inference, classification is formulated as retrieving the most relevant text prompt for a given image, where only the image-to-text similarity $\ell^{v\to t}$ is evaluated.
Under contrastive pretraining, CLIP performs zero-shot classification by matching images to text prompts in a shared embedding space via the image-to-text similarity $\ell^{v\to t}$~\citep{jia2021scaling,yao2021filip,zhai2022lit,zhou2022learning}.
Each class label $c_k$ ($k=1,\dots,C$, where $C$ is the number of candidate classes) is converted into a natural-language prompt (e.g., “This is a photo of a dog”), which is encoded and normalized to yield a class prototype $t_k \in \mathbb{S}^{d-1}$.
For a test image $x$, the normalized embedding is $v(x) = f_{\theta}(x)/\|f_{\theta}(x)\|_2,$ and the logit for class $c_k$ is $\ell_k^{v\to t}(x) = \langle v(x), t_k \rangle/\tau.$
The predictive distribution over classes is obtained via the softmax 
\begin{equation}\label{eq:clip-prob}
    p^{\text{CLIP}}(y=k \mid x) = \frac{\exp(\ell_k^{v\to t}(x))}{\sum_{j=1}^C \exp(\ell_j^{v\to t}(x))}.
\end{equation}
This formulation enables recognition of categories unseen during training, relying solely on the shared image–text embedding space. 

% \subsection{Zero-shot Adversarial Fine-tuning}
\subsection{Adversarial Attacks}
% \noindent\textbf{Adversarial Attacks.}
Adversarial attacks perturb inputs with small, often imperceptible changes to mislead a model. Given an image $x$ with label $y$, an adversarial example is constructed as
$
x^{a} = x + \delta, \|\delta\|_q \leq \epsilon,
$
% where $\epsilon$ bounds the perturbation magnitude under $\ell_q$-norm. A canonical method is \textit{Projected Gradient Descent }~\citep[PGD,][]{madry2017towards}, which iteratively updates
where $\epsilon$ bounds the perturbation magnitude under $\ell_q$-norm. A canonical method is \textit{Projected Gradient Descent }~\citep[PGD,][]{madry2017towards}. For the common $\ell_\infty$ threat model, it iteratively updates
\begin{equation}
x^{a}_{t+1} = \Pi_{B^{\infty}_\epsilon(x)} \Big(x^{a}_t + \alpha \,\mathrm{sign}\big(\nabla_x \mathcal{L}(F_\varphi(x^{a}_t), y)\big)\Big),    
\end{equation}
where $B^{\infty}_\epsilon(x)=\{x':\|x'-x\|_\infty\le \epsilon\}$, $t$ is the iteration index, $\alpha$ is the step size, $F_\varphi$ is the target model, and $\Pi_{B^{\infty}_\epsilon(x)}$ projects back to the feasible set. Intuitively, PGD takes a step that increases the loss and then clips the perturbed image to stay within the allowed budget.

% \noindent\textbf{Zero-shot Adversarial Robustness.}
% % To defend against such attacks, zero-shot adversarial fine-tuning (ZSAFT) extends adversarial training to the vision–language setting for CLIP. Instead of aligning adversarial images to one-hot labels, it leverages text embeddings as class prototypes and optimizes a text-guided contrastive loss~\citep{mao2022understanding,wang2024pre,yu2024text}. 
% To defend against such attacks, recent work extends adversarial fine-tuning from closed-set classification to the vision–language domain of CLIP~\citep{mao2022understanding,wang2024pre,yu2024text}. Unlike standard adversarial training, which aligns adversarial features with the one-hot target logit in a closed vocabulary, this paradigm addresses an open-vocabulary scenario where the text encoder remains unperturbed and provides class prototypes as reliable supervision. This framework, termed zero-shot adversarial robustness, typically adopts a text-guided contrastive loss:
% \begin{equation}\label{eq:tecoa}
% \min_\theta \; \mathbb{E}_{(x,y)} \Bigg[ 
%     - \log \frac{\exp\big(\langle v(x^{a}), t_y \rangle / \tau \big)}
%     {\sum_{j=1}^C \exp\big(\langle v(x^{a}), t_j \rangle / \tau \big)}
% \Bigg],
% \end{equation}
% where $v(x^{a})$ is the normalized adversarial image embedding. This paradigm helps preserve CLIP’s image–text alignment while improving adversarial robustness. However, the optimization is dominated by the target class $t_y$, with relations to non-target classes only implicitly captured.

\subsection{Uncertainty Estimation via Evidence}\label{sec:edl-define}

\noindent\textbf{Dirichlet Parameterization with Evidence.}  
% \paragraph{Dirichlet Parameterization with Evidence}  
In evidential deep learning (EDL), predictive uncertainty is modeled explicitly by placing a \emph{Dirichlet distribution} over class probabilities rather than predicting a single categorical distribution~\citep{sensoy2018evidential,malinin2018predictive,ulmer2021prior}.  
For a $C$-class problem, the network outputs non-negative concentration parameters $\alpha=(\alpha_1,\dots,\alpha_C)\in\mathbb{R}_+^C$, typically expressed as $\alpha_k = e_k + 1, e_k \geq 0$, where $e_k$ denotes the evidence assigned to class $k$. In the original EDL formulation, this ensures $\alpha_k \geq 1$ so that zero evidence corresponds to a uniform prior. The induced Dirichlet distribution is
\begin{equation}
\mathrm{Dir}(\pi;\alpha) = \frac{1}{B(\alpha)} \prod_{k=1}^C \pi_k^{\alpha_k - 1}, 
\quad 
B(\alpha) = \frac{\prod_{k=1}^C \Gamma(\alpha_k)}{\Gamma(\alpha_0)}, 
% \quad 
% \alpha_0 = \sum_{k=1}^C \alpha_k ,
\end{equation}
where $\pi=(\pi_1,\dots,\pi_C)$ is a probability on the $(C-1)$-simplex and $B(\alpha)$ is the polynomial Beta function. Importantly, $\alpha_0=\sum_{k=1}^C \alpha_k$ quantifies the total evidence and serves as the precision of the distribution. 

The non-negativity of $\alpha$ is typically enforced by activation functions such as ReLU, Softplus, or exponential mapping used in prior works~\citep{yoon2024uncertainty,malinin2019reverse}. In particular, under the exponential parameterization with unconstrained logits $z(x)\in\mathbb{R}^C$ and $\alpha_k(x)=\exp(z_k(x))$, the predictive categorical distribution is obtained as the expectation under the Dirichlet:
% \begin{equation}
% \label{eq:dir-exp-softmax}
% p(y=k\mid x)
% :=\mathbb{E}_{\pi\sim \mathrm{Dir}(\alpha(x))}[\pi_k]
% =\frac{\alpha_k(x)}{\alpha_0(x)}
% \overset{\alpha_k=\exp(z_k)}{=}
% \frac{\exp(z_k(x))}{\sum_{j=1}^{C}\exp(z_j(x))}.
% \end{equation}
\begin{equation}
\label{eq:dir-exp-softmax}
\begin{aligned}
p(y=k \mid x)
:&= \mathbb{E}_{\pi \sim \mathrm{Dir}(\alpha(x))}[\pi_k] \\
&= \frac{\alpha_k(x)}{\alpha_0(x)}
\overset{\alpha_k=\exp(z_k)}{=}
\frac{\exp(z_k(x))}{\sum_{j=1}^{C}\exp(z_j(x))}.
\end{aligned}
\end{equation}
% which naturally sums to one and defines a valid probability distribution. 

\noindent\textbf{Closed-Form Uncertainty Decomposition.}
% \paragraph{Closed-Form Uncertainty Decomposition}
The Dirichlet parameterization not only provides a probability distribution but also admits a closed-form decomposition of predictive uncertainty into two complementary components, aleatoric and epistemic~\citep{KIUREGHIAN2009105,kendall2017uncertainties,hullermeier2021aleatoric}.

\textit{Aleatoric uncertainty (AU)} captures ambiguity inherent in the data. In vision–language models, this may arise from factors such as semantic overlap between classes (e.g., “wolf” vs. “dog”) or noisy image–text pairs where multiple labels are plausible~\citep{ulmer2021prior,ma2025estimating,ji2023map}. Formally, AU reflects how probability mass is distributed across classes and is quantified by the expected Shannon entropy of the categorical distribution under the Dirichlet:
% \begin{equation}\label{eq:au}
% \mathrm{AU}(x) 
% = \mathbb{E}_{\pi \sim \mathrm{Dir}(\alpha)}\!\big[ H(\pi) \big] 
% = - \sum_{k=1}^C \frac{\alpha_k}{\alpha_0} 
%   \Big( \psi(\alpha_k + 1) - \psi(\alpha_0 + 1) \Big),
% \end{equation}
\begin{equation}\label{eq:au}
\begin{aligned}
\mathrm{AU}(x)
&= \mathbb{E}_{\pi \sim \mathrm{Dir}(\alpha)}\!\big[ H(\pi) \big] \\
&= - \sum_{k=1}^C \frac{\alpha_k}{\alpha_0}
\Big( \psi(\alpha_k + 1) - \psi(\alpha_0 + 1) \Big),
\end{aligned}
\end{equation}
where $\psi(\cdot)$ denotes the digamma function. 

\textit{Epistemic uncertainty (EU)} arises from limited evidence or distributional shift~\citep{hendrycks2016baseline,sensoy2018evidential}. It reflects the overall reliability of the prediction: when the total evidence $\alpha_0$ is small, the model should be considered untrustworthy. Following prior work~\citep{charpentier2020posterior,ulmer2021prior,ma2025estimating}, a widely adopted closed-form proxy is
\begin{equation}\label{eq:eu}
\mathrm{EU}(x) = \frac{C}{\alpha_0 + C},
\end{equation}
which increases as $\alpha_0$ decreases.

In summary, AU reflects ambiguity in the predictive distribution across classes, while EU captures uncertainty from insufficient evidence or distributional shift. Both can be computed directly from the Dirichlet parameters, enabling efficient uncertainty estimation in a single forward pass.

%% file: main/4-Prove.tex
\section{Dirichlet Reformulation of CLIP}\label{sec:prove}
Comparing CLIP’s zero-shot probability in Equation~\ref{eq:clip-prob} with the Dirichlet expectation in Equation~\ref{eq:dir-exp-softmax} reveals a structural correspondence: both are softmax operations over a set of logits. This motivates a \emph{non-trivial} identification that reinterprets CLIP logits as \emph{evidence} governing a Dirichlet distribution (Definition~\ref{definition}). 
% It is considered non-trivial because it (i) satisfies the validity of Dirichlet evidence (Lemma~\ref{lemma:validity}), (ii) recovers CLIP’s predictive rule exactly under a specific calibration (Lemma~\ref{lemma:equiv-tau}), and (iii) preserves logit order while exposing a tunable temperature for calibration (Corollary~\ref{cor:scale}).
This identification is non-trivial for three reasons: (i) it satisfies the validity of Dirichlet evidence with tight bounds and strict monotonicity (Lemma~\ref{lemma:validity}); (ii) it exactly recovers CLIP’s predictive rule exactly under a specific calibration (Lemma~\ref{lemma:equiv-tau}); and (iii) preserves logit order while exposing a tunable temperature for calibration (Corollary~\ref{cor:scale}).

\begin{definition}[Concentration Parameter]\label{definition}
Let $v(x),t_k\in\mathbb{S}^{d-1}$ be unit-normalized image/text embeddings and $\ell^{v\to t}_k(x)=\langle v(x),t_k\rangle/\tau$ the CLIP logit with temperature $\tau>0$. We define Dirichlet concentration parameters by
\begin{equation}\label{eq:param-edl}
\alpha_k(x) \;=\; \exp\!\big(h(\ell^{v\to t}_k(x))\big),
\qquad
h(\ell)\;=\;\frac{\tau\,\ell+1}{\tau^\prime},
\end{equation}
where $\tau^\prime>0$ is a calibration coefficient.
\end{definition}

% \noindent\textit{Remark (Construction rationale).}
\begin{remark}[Construction rationale]
Since $\tau\,\ell^{v\to t}_k(x)=\langle v(x),t_k\rangle\in[-1,1]$, we shift the cosine similarity by $+1$ so that its range becomes $[0,2]$. A calibration coefficient $\tau^\prime>0$ is introduced to rescale. Applying the exponential guarantees positivity while preserving logit order and remaining compatible with softmax geometry.
\end{remark}

% Building on this definition, we first establish validity, then prove exact equivalence to CLIP in a special case, and finally extend the result to the general setting.

\begin{lemma}[Validity of Dirichlet Evidence]\label{lemma:validity}
Under Definition~\ref{definition}, for all $k$:
\begin{enumerate}\itemsep0pt
\item $\alpha_k(x)\ge 1$ and $\alpha_k(x)\in[1,\exp(2/\tau^\prime)]$;
\item $\alpha=\exp(h(\ell))$ is strictly increasing.
\end{enumerate}
\end{lemma}

% \noindent\emph{Remark ($\alpha_k\ge 1$ in EDL).}
\begin{remark}[$\alpha_k\ge 1$ in EDL]
As introduced in Section~\ref{sec:edl-define}, the classical EDL formulation enforces $\alpha_k \geq 1$ by parameterizing $\alpha_k = e_k + 1$ with non-negative evidence~\citep{sensoy2018evidential,sensoy2020uncertainty}. 
We adopt the same restriction for two reasons: (i) digamma- and trigamma-based uncertainty measures become unstable as $\alpha_k$ approaches $0$~\citep{minka2000estimating}, and (ii) Dirichlet distributions with $\alpha_k<1$ produce corner-seeking samples~\citep{telgarsky2013dirichlet}, concentrating on a few classes even under weak evidence. This violates the common principle that uncertainty should grow as inputs become harder or deviate from the training distribution.
% We adopt the same restriction for two reasons. First, uncertainty measures that involve digamma and trigamma functions become numerically unstable as $\alpha_k$ approaches $0$~\citep{minka2000estimating}. Second, Dirichlet distributions with parameters below one generate sparse, corner-seeking samples~\citep{telgarsky2013dirichlet}. When $\alpha_k < 1$, the logit distribution concentrates on a few classes even if the evidence is weak, which contradicts the goal of calibrated evidence, namely to represent this distribution in a balanced and reliable way. 
Accordingly, our reformulation guarantees $\alpha_k \geq 1$; all subsequent analysis and experiments are under this regime. Proof is provided in Appendix~\ref{app:lemma1}. 
\end{remark}

\begin{figure*}[t]
    \centering
    \includegraphics[width=0.95\linewidth]{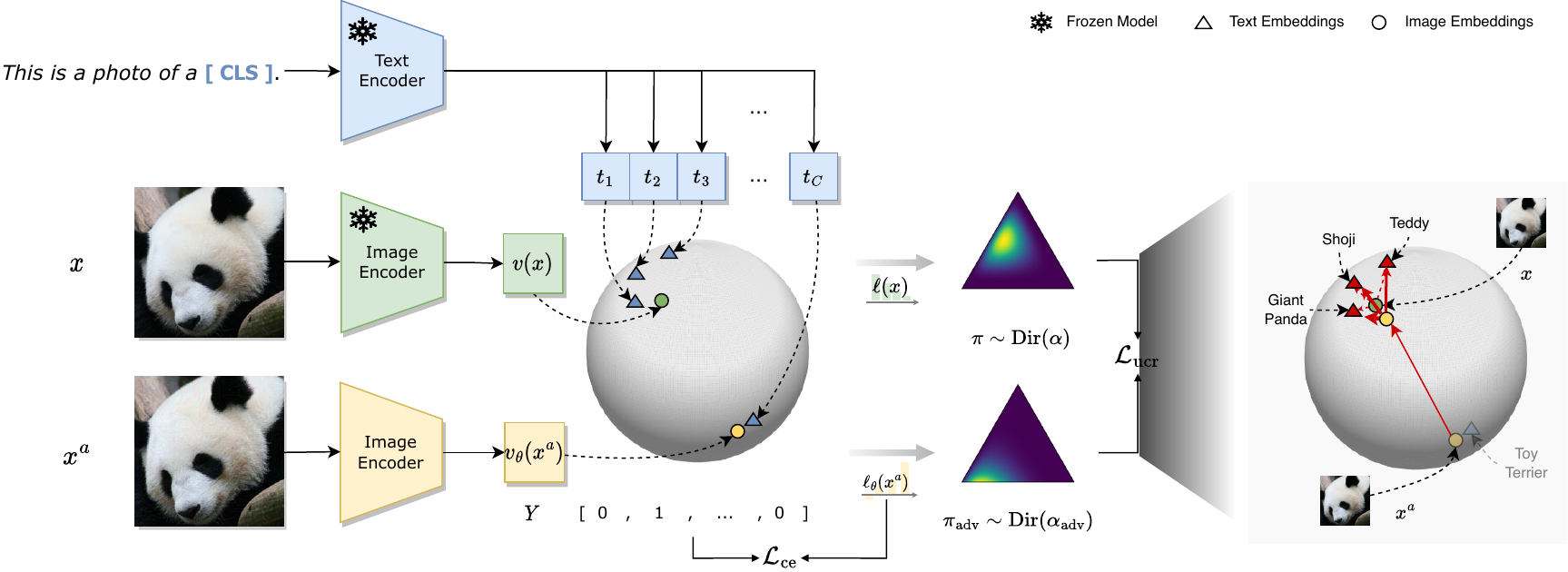}
    \caption{\textbf{Overview of our uncertainty calibration adversarial fine-tuning framework.} 
    Clean and adversarial images are encoded by CLIP’s image encoder, while text prompts are processed by the frozen text encoder. Our training objective combines the text-guided contrastive loss $\mathcal{L}_\text{ce}$ with an uncertainty calibration regularization term $\mathcal{L}_\text{ucr}$ that aligns adversarial Dirichlet distributions with the corresponding clean Dirichlet distributions, aiming to better preserve semantic relations and calibrate evidence strength.
    }
    \label{fig:framework}
\end{figure*}

\begin{lemma}[Exact Equivalence at $\tau=\tau^\prime$]\label{lemma:equiv-tau}
Let $s=\tau/\tau^\prime$. If $s=1$ (equivalently $\tau^\prime=\tau$), the Dirichlet expectation equals to CLIP’s softmax:
% \begin{equation}
% p^{\mathrm{Dir}}_k(x) \;=\; \frac{\alpha_k}{\sum_j \alpha_j}
% = \frac{\exp(h(\ell_k))}{\sum_j \exp(h(\ell_j))}
% = \operatorname{softmax}\!\big(\ell(x)\big)_k
% = p^{\mathrm{CLIP}}_k(x).
% \end{equation}
\begin{equation}
\begin{aligned}
% p^{\mathrm{Dir}}_k(x)
% &= \frac{\alpha_k}{\sum_j \alpha_j} 
% = \frac{\exp(h(\ell_k))}{\sum_j \exp(h(\ell_j))} \\
% &= \operatorname{softmax}\!\big(\ell(x)\big)_k 
% = p^{\mathrm{CLIP}}_k(x).
p^{\mathrm{Dir}}_k(x)
&= \frac{\alpha_k}{\sum_j \alpha_j} 
= \frac{\exp(h(\ell_k))}{\sum_j \exp(h(\ell_j))} 
% &= \operatorname{softmax}\!\big(\ell(x)\big)_k 
= p^{\mathrm{CLIP}}_k(x).
\end{aligned}
\end{equation}
\end{lemma}

% \noindent\textit{Remark (Significance of exact equivalence).}
\begin{remark}[Significance of exact equivalence]
Lemma~\ref{lemma:equiv-tau} shows that when $\tau^\prime=\tau$, the Dirichlet expectation coincides exactly with CLIP’s softmax prediction. This equivalence is not incidental: it demonstrates that CLIP’s original training loss in Equation~\ref{loss:infonce} 
% implicitly optimizes a Dirichlet-based model of evidence. 
{admits a Dirichlet-consistent reinterpretation.}
Hence, our reformulation is not an ad hoc construction but a faithful probabilistic interpretation of CLIP’s logits. A complete proof is provided in Appendix~\ref{app:lemma2}.  
\end{remark}

% \begin{corollary}[General form and invariances]\label{cor:scale}
% For arbitrary $\tau^\prime>0$,
% \begin{equation}
% p^{\mathrm{Dir}}(x)=\operatorname{softmax}\!\big(s\,\ell(x)\big), 
% \qquad s={\tau}/{\tau^\prime}>0.
% \end{equation}
% Hence $\arg\max_k p^{\mathrm{Dir}}_k(x)=\arg\max_k p^{\mathrm{CLIP}}_k(x)$, while the entropy of the distribution can be smoothly tuned by $s$: larger $s$ yields sharper predictions, smaller $s$ yields flatter ones.
% \end{corollary}

\begin{corollary}[General form and invariances]\label{cor:scale}
For arbitrary $\tau^\prime>0, s=\tau/\tau^\prime>0$, $p^{\mathrm{Dir}}(x)=\operatorname{softmax}\big(s\,\ell(x)\big).$ Hence 
\begin{equation}
 \arg\max_k p^{\mathrm{Dir}}_k(x)=\arg\max_k p^{\mathrm{CLIP}}_k(x),
\end{equation}
while the entropy of the distribution can be smoothly tuned by $s$: larger $s$ yields sharper predictions, smaller $s$ yields flatter ones.
\end{corollary}

% \noindent\textit{Remark (Connection to uniformity–tolerance in contrastive learning).}
\begin{remark}[Connection to uniformity–tolerance in contrastive learning]
In contrastive learning, the temperature regulates the separation strength among negatives. A \emph{smaller} softmax temperature (larger $s$) encourages \emph{uniformity} on the hypersphere by enforcing stronger separation, while a \emph{larger} temperature (smaller $s$) increases \emph{tolerance} to near-semantic neighbors~\citep{wang2020understanding,wang2021understanding}.
% We set $\tau^\prime=0.07$, yielding $s<1$ and thus softer predictions that increase tolerance to semantically related negatives. 
% {Our choice also matches the canonical temperature used in MoCo~\citep{he2020momentum} and other contrastive frameworks~\citep{radford2021learning,jia2021scaling}, which has been widely validated for stable and well-separated representations, with sensitivity analysis in Appendix~\ref{re:sec-tauprime}. Such a setting} preserves CLIP’s intrinsic semantic structure and is particularly beneficial for adversarial fine-tuning, where calibrated tolerance improves zero-shot robustness without harming the model’s original generalization ability. Proof is deferred to Appendix~\ref{app-cor:scale}.
We set $\tau^\prime=0.07$, a canonical temperature in contrastive learning~\citep{he2020momentum,radford2021learning,jia2021scaling}.
As shown in Appendix~\ref{app-cor:scale}, this implies $s<1$ and thus softer predictions with increased tolerance to semantically related negatives.
Sensitivity analysis over $\tau^\prime$ is reported in Appendix~\ref{re:sec-tauprime}.
This choice maintains the relative similarity structure induced by CLIP and is used throughout our adversarial fine-tuning experiments to control the tolerance level.
\end{remark}

\noindent\textbf{Implications of the reformulation.}
This reformulation establishes a principled mapping from CLIP logits to Dirichlet evidence.
Specifically, it enables: (i) \textbf{closed-form uncertainty decomposition} into AU/EU without sampling (Section~\ref{sec:edl-define}); (ii) \textbf{principled calibration} via $\tau^\prime$, which adjusts predictive sharpness while preserving the argmax (Corollary~\ref{cor:scale}) and thus exposes a controllable uniformity--tolerance trade-off; and (iii) \textbf{semantic fidelity}, recovering CLIP’s predictive rule in the exact-equivalence case (Lemma~\ref{lemma:equiv-tau}) while retaining both relative geometry and absolute evidence strength.

These properties directly motivate the adversarial fine-tuning objectives introduced in the next section.

%% file: main/5_loss.tex
\section{Uncertainty Calibration Adversarial Fine-tuning Objective}
\input{Tab/MultiLabel-CW-re}

To mitigate the misaligned semantics and unreliable confidence introduced by adversarial perturbations, we propose an \emph{Uncertainty Calibration Adversarial fine-Tuning (UCAT)} objective.   
The key insight builds on our reformulation: mapping CLIP logits to Dirichlet evidence yields closed-form uncertainty decomposition with principled calibration, while retaining fidelity to the semantic geometry of the embedding space. UCAT exploits this property by aligning the Dirichlet distributions of adversarial and clean samples, correcting distributional shift while simultaneously preserving \emph{semantic relations} and \emph{calibrated confidence}.

As illustrated in Figure~\ref{fig:framework}, our method adopts a CLIP-based adversarial fine-tuning pipeline with a frozen text encoder and a trainable image encoder. Clean samples $x$ and their adversarial counterparts $x^a$ (generated via $\ell_\infty$-PGD~\citep{madry2017towards}) are encoded into the joint embedding space, and their logits are reformulated as Dirichlet parameters, denoted $\alpha$ and $\alpha_{\text{adv}}$. The clean distribution $\mathrm{Dir}(\alpha)$ captures the generalized semantics from pre-training, whereas $\mathrm{Dir}(\alpha_{\text{adv}})$ may shift toward distorted or overconfident states. To correct this mismatch, we introduce an \emph{uncertainty calibration regularization} objective, defined as the {reverse} KL divergence between the two distributions:
\begin{equation}
\mathcal{L}_{\mathrm{ucr}} 
= \mathrm{KL}\!\left( \mathrm{Dir}(\alpha_{\text{adv}}) \,\|\, \mathrm{Dir}(\alpha) \right).
\end{equation}
{Our choice of the reverse KL direction follows naturally from the distribution alignment objective and significantly impacts optimization. Unlike the forward KL, which covers modes and flattens evidence, the reverse KL is mode-seeking. It preserves both relative class structure and absolute evidence strength by allowing low evidence on irrelevant classes~\citep{malinin2019reverse}.}
Since both AU and EU are closed-form functions of Dirichlet parameters (Sec.~\ref{sec:edl-define}), minimizing $\mathcal{L}_{\mathrm{ucr}}$ aligns adversarial predictions with their clean counterparts in terms of \emph{inter-class relations} (AU) and \emph{evidence magnitude} (EU).
{This alignment prevents collapse into spuriously confident errors, enhancing uncertainty estimation and zero-shot adversarial robustness.}
% , thereby preventing collapse into spuriously confident errors.  

Complementarily, the text-guided cross-entropy loss
\begin{equation}
    \mathcal{L}_{\mathrm{ce}} = - \log \frac{\exp\big(\langle v(x^{a}), t_y \rangle / \tau \big)}
    {\sum_{j=1}^C \exp\big(\langle v(x^{a}), t_j \rangle / \tau \big)},
\end{equation}
anchors adversarial embeddings to the ground-truth prototype $t_y$, providing discriminative supervision that stabilizes training and improves accuracy.  
The final objective combines both components:
\begin{equation}\label{loss:all}
\mathcal{L} = \mathcal{L}_{\mathrm{ce}} + \lambda \,\mathcal{L}_{\mathrm{ucr}},
\end{equation}
where $\lambda$ balances discriminative alignment and uncertainty calibration. This joint objective combines discriminative supervision via the cross-entropy loss with calibrated uncertainty through distributional alignment, leading to stronger zero-shot adversarial robustness.

%% file: Tab/MultiLabel-CW-re.tex
% \usepackage[normalem]{ulem}
\begin{table*}[!ht] 
\centering 
\caption{\textbf{Zero-shot adversarial robustness on multi-label dataset MS-COCO~\citep{lin2014microsoft}.} 
All ZSAR models are adversarially trained on TinyImageNet with the FARE~\citep{schlarmann2024robust} 10-step PGD setting ($\epsilon=1/255$), and evaluated under CW-100 at radii $\epsilon\in\{1/255,2/255,4/255\}$ plus clean. We report mean Average Precision (mAP), Precision (P), Recall (R), and F1-score (F1) at top-3 predictions. $H(\mathrm{F1@3})$ denotes the harmonic mean of clean and adversarial F1@3. Best and second-best are in \textbf{bold} and \underline{underline}.
}
\resizebox{\linewidth}{!}{
\begin{tabular}{l|rrrr|rrrrr|rrrrr|rrrrr}
\toprule
{\textbf{}} & \multicolumn{4}{c|}{Clean} & \multicolumn{5}{c|}{$\epsilon=1/255$}& \multicolumn{5}{c|}{$\epsilon=2/255$}& \multicolumn{5}{c}{$\epsilon=4/255$}\\

Methods & \multicolumn{1}{c}{{{mAP}}}  & \multicolumn{1}{c}{{{ P@3}}} & \multicolumn{1}{c}{{{ R@3}}} & \multicolumn{1}{c|}{{{ F1@3}}} 
& \multicolumn{1}{c}{{{mAP}}}  & \multicolumn{1}{c}{{{ P@3}}} & \multicolumn{1}{c}{{{ R@3}}} & \multicolumn{1}{c}{{{ F1@3}}} & \multicolumn{1}{c|}{H\scriptsize{(F1@3)}} 
& \multicolumn{1}{c}{{{mAP}}}  & \multicolumn{1}{c}{{{ P@3}}} & \multicolumn{1}{c}{{{ R@3}}} & \multicolumn{1}{c}{{{ F1@3}}} & \multicolumn{1}{c|}{H\scriptsize{(F1@3)}} 
& \multicolumn{1}{c}{{{mAP}}}  & \multicolumn{1}{c}{{{ P@3}}} & \multicolumn{1}{c}{{{ R@3}}} & \multicolumn{1}{c}{{{ F1@3}}} & \multicolumn{1}{c}{H\scriptsize{(F1@3)}} \\
\midrule

\rowcolor[HTML]{EDEDED}
CLIP~\citep{radford2021learning} 
& {\color[HTML]{757171} \textbf{51.96}} 
& {\color[HTML]{757171} \textbf{45.33}}  
& {\color[HTML]{757171} \textbf{46.49}}  
& {\color[HTML]{757171} \textbf{45.90}}  
& {\color[HTML]{757171} \textbf{17.72}}
& {\color[HTML]{757171} \textbf{25.21}}  
& {\color[HTML]{757171} \textbf{25.85}}  
& {\color[HTML]{757171} \textbf{25.52}}  
& {\color[HTML]{757171} \textbf{32.80}}  
& {\color[HTML]{757171} \textbf{6.67}}
& {\color[HTML]{757171} \textbf{10.07}} 
& {\color[HTML]{757171} \textbf{10.32}}  
& {\color[HTML]{757171} \textbf{10.19}}  
& {\color[HTML]{757171} \textbf{16.68}}  
& {\color[HTML]{757171} \textbf{3.77}}  
& {\color[HTML]{757171} \textbf{3.84}}  
& {\color[HTML]{757171} \textbf{3.94}}  
& {\color[HTML]{757171} \textbf{3.89}}  
& {\color[HTML]{757171} \textbf{7.17}}  \\

TeCoA~\citep{mao2022understanding}& 46.34 & 33.73& 34.57& 34.14 & 37.32 & 30.23& 30.99& 30.60 & 32.27 & 27.94 & 24.85& 25.48& 25.16 & 28.97 & \textbf{17.63} & {\ul 19.67}& {\ul 20.17}& {\ul 19.91} & {\ul 25.15} \\
FARE~\citep{schlarmann2024robust} & \textbf{49.21} & \textbf{43.83}& \textbf{44.95}& \textbf{44.37} & 29.18 & {\ul 33.45}& {\ul 34.30}& {\ul 33.86} & {\ul 38.41} & 16.18 & 22.44& 23.02& 22.72 & 30.05 & 8.12 & 13.21& 13.55& 13.38 & 20.56 \\
PMG-AFT~\citep{wang2024pre} & {\ul 48.94} & {\ul 41.77}& {\ul 42.83}& {\ul 42.29} & 29.75 & 32.32& 33.15& 32.72 & 36.89 & 17.16 & 23.37& 23.98& 23.67 & 30.35 & 8.78 & 13.45& 13.80 & 13.62 & 20.60 \\
TGA-ZSR~\citep{yu2024text} & 48.61 & 37.19& 38.13& 37.65 & \textbf{38.23} & 32.95& 33.79& 33.36 & 35.38 & {\ul29.03} & {27.70}& {28.40}& {28.04} & {32.14} & {\ul 15.18} & 18.64& 19.11& 18.87 & 25.14 \\

{Comp-TGA~\citep{yu2026complementary}} & {48.14} & {38.11} & {39.07} & {38.58} & {37.57} & {33.05} & {33.88} & {33.45} & {35.83} & {27.23} & {\ul 28.16} & {\ul 28.87} & {\ul 28.51} & {\ul 32.79} & {13.61} & {18.08} & {18.54} & {18.30} & {24.82} \\

\rowcolor[HTML]{faebd7} UCAT (Ours) & 47.14 & 41.55& 42.62& 42.07 & {\ul 37.60} & \textbf{36.58}& \textbf{37.52}& \textbf{37.04} & \textbf{39.40} & \textbf{29.33} & \textbf{31.39}& \textbf{32.20}& \textbf{31.78} & \textbf{36.21} & 14.61 & \textbf{21.33}& \textbf{21.86}& \textbf{21.59} & \textbf{28.54} \\
\bottomrule
\end{tabular}} \label{tab:multilabelCW}
\vspace{-1em}
\end{table*}

%% file: main/6_exp.tex
\section{Experiments}

\noindent\textbf{Experimental Setup.}
We adopt CLIP-B/32~\citep{radford2021learning} as the backbone and follow TeCoA's training protocol~\citep{mao2022understanding}, comparing zero-shot adversarial robustness against five baselines: CLIP~\citep{radford2021learning}, TeCoA, FARE~\citep{schlarmann2024robust}, PMG-AFT~\citep{wang2024pre}, and TGA-ZSR~\citep{yu2024text}. Training and evaluation are conducted under $\ell_\infty$ PGD regimes, including a light setting (2-step, $\epsilon=1/255$) following TeCoA and a stronger setting (10-step, $\epsilon=2/255$) following FARE. Robustness is further assessed using 100-step PGD~\citep{madry2017towards}, CW~\citep{carlini2017towards}, and AutoAttack~\citep{croce2020reliable}. We set $\lambda={10^{5}}/{\beta}$ with $\beta=2/e^{\tau'}$, and fix $\tau'=0.07$ following standard contrastive learning practices~\citep{wu2018unsupervised,he2020momentum,radford2021learning,yeh2022decoupled}. Full implementation details and datasets are provided in the Appendix~\ref{app:imp}.
Code is available at \href{https://github.com/VivienLu/UCAT}{https://github.com/VivienLu/UCAT}.

% \subsection{Experimental results}

\subsection{Robustness under Multi-Label Ambiguity}
\input{Tab/tinyImageNet-eps1-re}

To assess robustness under multi-label ambiguity, we perform zero-shot evaluation on the multi-label MS-COCO~\citep{lin2014microsoft} dataset (Table~\ref{tab:multilabelCW}). All models are fine-tuned on single-label TinyImageNet using PGD and tested directly on COCO under CW attacks that perturb multiple labels simultaneously. Across three attack strengths, our method consistently achieves the best top-$3$ precision, recall, F1, and harmonic mean of clean and adversarial F1@3, indicating a superior accuracy–robustness trade-off for the top-ranked predictions. 

Compared with single-anchor approaches that align adversarial visual features to single-category text features or original visual features, our method aligns clean and adversarial Dirichlet distributions in the evidence (similarity) space. This distributional alignment preserves both relative semantic ordering among categories and absolute evidence strength, without enforcing softmax-based cross-class normalization, thereby stabilizing semantically relevant labels near the head of the ranking under multi-label ambiguity. Consequently, improvements are more pronounced on metrics emphasizing top-$k$ reliability, while gains in mAP are more limited, as mAP evaluates fine-grained global ordering across all labels, including low-confidence and long-tail categories. Overall, these results indicate that our distributional alignment mechanism effectively stabilizes the most relevant semantic predictions under adversarial perturbations, supporting its design for robust open-vocabulary recognition in ambiguous multi-label settings.

\subsection{Cross-Dataset Zero-Shot Adversarial Robustness}

To verify the effectiveness of our approach under single-label settings, we analyze results across 16 datasets (Table~\ref{tab:tinyimagenet-eps1}).  
Our method achieves consistently strong performance, ranking best or second-best in nearly all cases.  
When trained with a single PGD regime, it generalizes effectively to multiple adversarial attacks while maintaining both the highest clean accuracy and adversarial robustness.  The only exceptions are two domain-specific datasets (PCAM~\citep{veeling2018rotation} and EuroSAT~\citep{helber2019eurosat}), which exhibit the highest predictive uncertainty (high PU in Fig.~\ref{fig:pu_bar}, high AU in Fig.~\ref{fig:au-16}, and low EU in Fig.~\ref{fig:eu-16}) and strong semantic overlap.
% , where highly specialized semantics limit the gains of our distributional alignment strategy. 
{These characteristics reflect a substantial departure from CLIP’s natural-image pre-training domain, resulting in inherently weaker clean semantic geometry. Consequently, UCAT has less reliable structure to preserve through Dirichlet alignment, naturally limiting the magnitude of improvement. Nevertheless, UCAT remains stable on these domain-shifted datasets and achieves state-of-the-art robustness on the majority of natural-image benchmarks, where CLIP provides strong clean semantic structure and our Dirichlet alignment is most effective.}
% Nevertheless, by capturing broader semantic structures and evidence strength, our method achieves state-of-the-art robustness and generalization across the full evaluation suite. 
We further extend our evaluation to larger-scale training, stronger attack settings, {and additional ablations}, with results reported in Appendix~\ref{app:exp}.
% We further extend our evaluation by training in the same manner as TeCoA~\citep{mao2022understanding} on ImageNet-1k~\citep{deng2009imagenet} with 2-step PGD at $\epsilon=1/255$ to assess performance on a larger training dataset, and by training in the same manner as FARE~\citep{schlarmann2024robust} on TinyImageNet with 10-step PGD~\citep{madry2017towards} at the same epsilon ($\epsilon=2/255$) to assess performance under a stronger attack. The results of these additional experiments are provided in the Appendix~\ref{tab:tinyimagenet-eps2,tab:imagnet}.

\subsection{Generalization across Vision--Language Backbones}
\input{Tab/VLM-main}

\begin{figure*}[htbp]
    \centering
    \begin{subfigure}[htbp]{0.29\linewidth}
        \includegraphics[width=\linewidth]{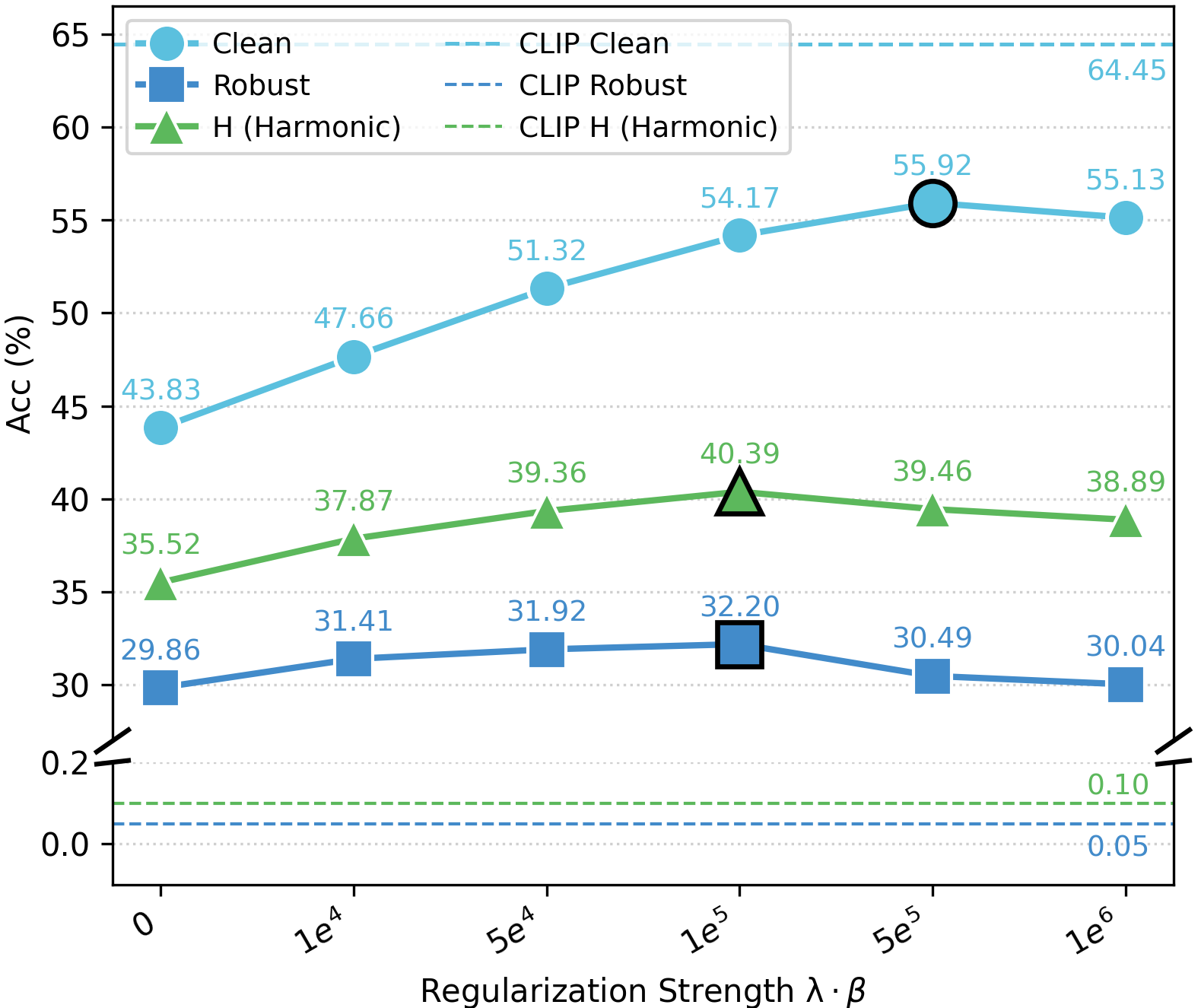}
        \caption{Sensitivity analysis of $\lambda$}
        \label{fig:sa}
    \end{subfigure}
    \begin{subfigure}[htbp]{0.29\linewidth}
        \includegraphics[width=\linewidth]{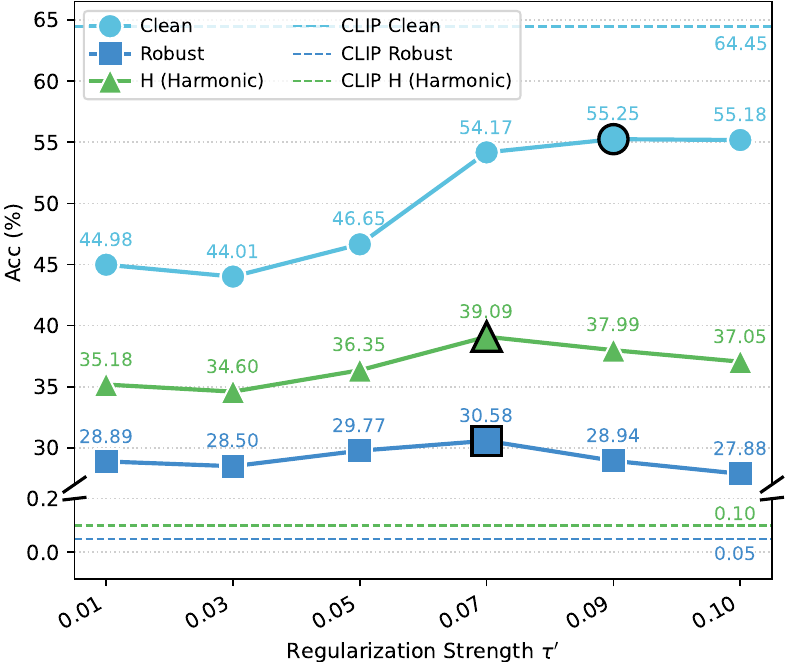}
        \caption{Sensitivity analysis of $\tau^\prime$}
        \label{fig:sa_tau}
    \end{subfigure}
    \begin{subfigure}[htbp]{0.33\linewidth}
        \includegraphics[width=\linewidth]{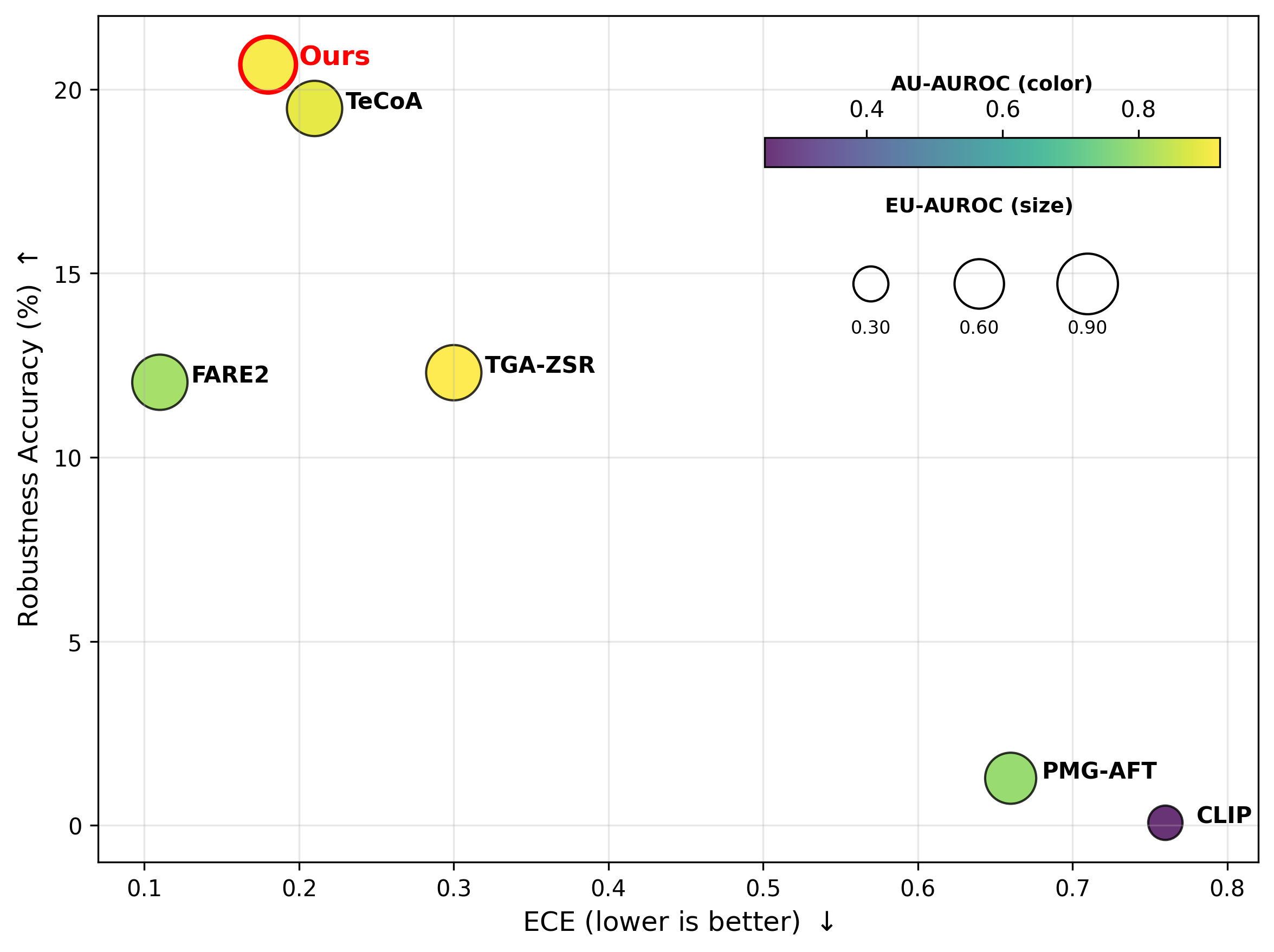}
        \caption{Calibration analysis.}
        \label{fig:auroc}
    \end{subfigure}
    \caption{
    \textbf{Parameter sensitivity and robustness--calibration trade-off.} 
    \textbf{(a) Sensitivity of regularization strength $\lambda$.} We vary $\lambda\cdot\beta\in\{{10^{4}}, 5\times10^{4}, 10^{5}, 5\times10^{5}, 10^{6}\}$ with $\beta=2/e^{\tau'}$ and $\tau'=0.07$. Models are adversarially fine-tuned with $\ell_\infty$ PGD-2 ($\epsilon=1/255$) and evaluated by $\ell_\infty$ PGD-100~\citep{madry2017towards} (same $\epsilon$). Curves report averages over 16 datasets of Clean, Robust, and their harmonic mean $H$. \textbf{(b) Sensitivity of calibration coefficient $\tau^\prime$.} Same as (a) while varying $\tau^\prime$. \textbf{(c) Strong-attack robustness vs. calibration.} Models are fine-tuned with an $\ell_\infty$ 10-step PGD attack ($\epsilon=2/255$) and evaluated with AutoAttack~\citep{croce2020reliable} at the same $\epsilon$, with all points averaged over 16 datasets. The x-axis: Expected Calibration Error (ECE; lower is better) and the y-axis is robust accuracy (higher is better). Bubble color/size encode AU-/EU-AUROC (aleatoric/epistemic uncertainty for error detection).}

    % % (a) \textbf{Sensitivity analysis of the regularization strength $\lambda$.} We evaluate $\lambda\cdot\beta \in \{{10^{4}}, {5\times10^{4}}, {10^{5}}, {5\times10^{5}}, {10^{6}}\}, {\beta}=2/{e^{\tau'}}$ on all 16 datasets, reporting averages of clean accuracy, PGD-100~\citep{madry2017towards} robustness, and their harmonic mean. 
    % (a) \textbf{$\lambda$ sensitivity.} Mean clean accuracy, PGD-100 robustness, and harmonic mean on 16 datasets vs. $\lambda\cdot\beta$ (with $\beta=2/e^{\tau'}$). 
    % (b) \textbf{$\tau^\prime$ sensitivity.} Same metrics vs. $\tau^\prime$. 
    % % (b) \textbf{Comprehensive evaluation {averaged over 16 datasets} under strong adversarial training (PGD-10, $\epsilon=2/255$) and AutoAttack~\citep{croce2020reliable} testing.}
    % (b) \textbf{Robustness vs. calibration under strong attacks.} Robust accuracy vs. ECE under PGD-10 training ($\epsilon=2/255$) and AutoAttack testing averaged over 16 datasets.
    % % X-axis shows calibration error (ECE, lower is better), while Y-axis shows robustness accuracy.
    % Bubble color indicates AU-AUROC and bubble size indicates EU-AUROC, reflecting the discriminative power of aleatoric and epistemic uncertainty.}
    \label{fig:vis}
\end{figure*}
We further evaluate whether UCAT generalizes beyond CLIP-B/32 to other \emph{contrastively pretrained} VLMs, including CLIP-B/16~\citep{radford2021learning} and SLIP-B/16~\citep{mu2022slip} (Self-supervised Language--Image Pre-training).
As shown in Table~\ref{tab:vlm-main}, UCAT consistently improves AutoAttack robustness and the clean--robust trade-off (harmonic mean, $H$) across backbones, indicating that our uncertainty-calibrated Dirichlet distribution matching is not tied to a specific CLIP variant but broadly applicable to contrastive VLMs with shared image--text embedding spaces. Full per-dataset results are provided in Appendix~\ref{app:exp}, Table~\ref{tab:re-vlm}.

\subsection{Ablation and Parameter Sensitivity}
\input{Tab/Abl}

Table~\ref{tab:abl} ablates our loss design. Starting from text-guided cross-entropy $\mathcal{L}_{\text{ce}}$, {we systematically evaluate two key design dimensions: the distribution levels (probability vs. Dirichlet) and the KL divergence directions (forward vs. reverse).} {At the probability level (rows 3–4),} introducing a KL term to align \emph{softmax} distributions {yields only} a modest gain. {This limited improvement stems from the fact that} softmax normalization discards overall magnitude information, failing to preserve absolute evidence strength. {When shifting to the Dirichlet level, comparing rows 5 and 6 clearly} reveals that the reverse KL direction significantly outperforms its forward counterpart, {validating} the necessity of its mode-seeking behavior {under adversarial perturbations}. {Consequently}, our {adopted} reverse Dirichlet KL alignment (last row) {simultaneously integrates both optimal choices to align} both relative geometry and evidence magnitude, yielding the best observed trade-off between clean and robust performance.

% Table~\ref{tab:abl} ablates our loss design. Starting from text-guided cross-entropy $\mathcal{L}_{\text{ce}}$, we {systematically ablate the alignment spaces (probability vs. Dirichlet distributions) and the KL alignment directions (forward vs. reverse).} 
% introduce a probability-level KL term (third row) to test whether aligning \emph{softmax} output distributions can improve robustness by preserving relative class geometry alone. 
% This results in a modest gain, but does not explicitly preserve absolute evidence strength, as softmax normalization discards overall magnitude information. In contrast, replacing it with our Dirichlet KL alignment (last row) aligns both relative geometry and evidence magnitude, yielding the best observed trade-off between clean and robust performance.

Performance remains stable across a broad range of $\lambda$ values, with the best trade-off achieved at ${10^{5}/\beta}$ (Fig.~\ref{fig:sa}). Sensitivity to $\tau^\prime$ is also mild around the default choice (Fig.~\ref{fig:sa_tau}). Detailed results are provided in Appendix~\ref{re:sec-tauprime}.

\subsection{Robustness and Calibration under Strong Attacks}
Figure~\ref{fig:auroc} provides a comprehensive evaluation under AutoAttack~\citep{croce2020reliable} with $\epsilon=2/255$. We report four complementary metrics.\textit{ Expected Calibration Error (ECE)} (x-axis) measures how well predicted confidence matches actual correctness (lower is better), while \textit{robustness accuracy} (y-axis) captures the ability to resist adversarial perturbations (higher is better). Bubble color denotes \textit{AU-AUROC}, reflecting how aleatoric uncertainty helps identify errors caused by class ambiguity, and bubble size denotes \textit{EU-AUROC}, reflecting how epistemic uncertainty captures errors due to insufficient evidence.
An ideal model should lie toward the top-left of the plot (high robustness, low ECE) with large and bright bubbles (high AU-AUROC and EU-AUROC). Our method is closest to this desirable region: it achieves the highest robustness accuracy, maintains lower calibration error than existing baselines, and exhibits stronger uncertainty discrimination as shown by larger and brighter bubbles. This demonstrates that our uncertainty calibration not only strengthens adversarial robustness but also improves predictive reliability under attack.

%% file: Tab/tinyImageNet-eps1-re.tex
\begin{table*}[!ht]
\centering
\caption{\textbf{Zero-shot adversarial robustness across 16 single-label datasets.}
All methods are fine-tuned on TinyImageNet following TGA-ZSR~\citep{yu2024text}, adversarial training uses 2-step PGD~\citep{madry2017towards} with $\epsilon\!=\!1/255$. \textit{Average} is the mean across datasets. \textit{H} is the harmonic mean between Clean and the corresponding robust score. Best and second-best are in \textbf{bold} and \underline{underline}.}
\resizebox{\linewidth}{!}{
\scriptsize 
\setlength{\tabcolsep}{4pt} 
\begin{tabular}{ll|*{16}{r}|rr} 
\toprule

\multicolumn{2}{l|}{Methods} &
\multicolumn{1}{c}{\rotatebox{90}{TinyImageNet}} &
\multicolumn{1}{c}{\rotatebox{90}{CIFAR-10}} &
\multicolumn{1}{c}{\rotatebox{90}{CIFAR-100}} &
\multicolumn{1}{c}{\rotatebox{90}{STL10}} &
\multicolumn{1}{c}{\rotatebox{90}{SUN397}} &
\multicolumn{1}{c}{\rotatebox{90}{Food101}} &
\multicolumn{1}{c}{\rotatebox{90}{Oxfordpets}} &
\multicolumn{1}{c}{\rotatebox{90}{Flowers102}} &
\multicolumn{1}{c}{\rotatebox{90}{DTD}} &
\multicolumn{1}{c}{\rotatebox{90}{EuroSAT}} &
\multicolumn{1}{c}{\rotatebox{90}{FGVC Aircraft}} &
\multicolumn{1}{c}{\rotatebox{90}{ImageNet}} &
\multicolumn{1}{c}{\rotatebox{90}{Caltech101}} &
\multicolumn{1}{c}{\rotatebox{90}{Caltech256}} &
\multicolumn{1}{c}{\rotatebox{90}{StanfordCars}} &
\multicolumn{1}{c}{\rotatebox{90}{PCAM}} &
\multicolumn{1}{|c}{\rotatebox{90}{Average}} &
\multicolumn{1}{c}{\rotatebox{90}{H}} \\
\midrule

%%%%%%%%%%%%%%%%%%%%%%%%%%%% Clean
\rowcolor[HTML]{EDEDED}
{\cellcolor[HTML]{FFFFFF}\multirow{7}{*}{\rotatebox{90}{Clean}}} 
 & \multicolumn{1}{l|}{CLIP~\citep{radford2021learning}}                     & {\color[HTML]{757171} \textbf{57.96}} & {\color[HTML]{757171} \textbf{88.03}} & {\color[HTML]{757171} \textbf{60.45}} & {\color[HTML]{757171} \textbf{97.03}} & {\color[HTML]{757171} \textbf{57.26}} & {\color[HTML]{757171} \textbf{83.89}} & {\color[HTML]{757171} \textbf{87.41}} & {\color[HTML]{757171} \textbf{65.49}} & {\color[HTML]{757171} \textbf{40.64}} & {\color[HTML]{757171} \textbf{42.66}} & {\color[HTML]{757171} \textbf{20.16}} & {\color[HTML]{757171} \textbf{59.15}} & {\color[HTML]{757171} \textbf{85.32}} & {\color[HTML]{757171} \textbf{81.73}} & {\color[HTML]{757171} \textbf{52.02}} & {\color[HTML]{757171} \textbf{52.08}} & {\color[HTML]{757171} \textbf{64.45}} & {\color[HTML]{757171} \textbf{}} \\
& \multicolumn{1}{l|}{TeCoA~\citep{mao2022understanding}}      & 71.24 & 67.56 & 38.26 & 85.89 & 36.01 & 28.23 & 61.30 & 32.04 & 24.95 & 16.13 & 5.19  & 32.89 & 72.16 & 59.00 & 20.28 & 50.11 & 43.83 &  \\
 & FARE~\citep{schlarmann2024robust}       & \multicolumn{1}{r}{41.86}             & \multicolumn{1}{r}{{79.81}}       & \multicolumn{1}{r}{{48.27}}       & \multicolumn{1}{r}{\textbf{94.24}}    & \multicolumn{1}{r}{{46.15}}       & \multicolumn{1}{r}{\textbf{58.90}}    & \multicolumn{1}{r}{\textbf{80.98}}    & \multicolumn{1}{r}{{\ul 47.63}}       & \multicolumn{1}{r}{23.09}             & \multicolumn{1}{r}{\textbf{24.19}}    & \multicolumn{1}{r}{\textbf{15.63}}    & \multicolumn{1}{r}{42.93}             & \multicolumn{1}{r}{78.22}             & \multicolumn{1}{r}{{\ul 72.05}}       & \multicolumn{1}{r}{\textbf{43.96}}    & \multicolumn{1}{r|}{50.02}             & 53.00 &  \\
 & PMG-AFT~\citep{wang2024pre}    & \multicolumn{1}{r}{48.60}             & \multicolumn{1}{r}{74.73}             & \multicolumn{1}{r}{43.59}             & \multicolumn{1}{r}{90.41}             & \multicolumn{1}{r}{\textbf{51.70}}    & \multicolumn{1}{r}{{\ul 56.52}}       & \multicolumn{1}{r}{{\ul 79.40}}       & \multicolumn{1}{r}{\textbf{48.43}}    & \multicolumn{1}{r}{\textbf{32.45}}    & \multicolumn{1}{r}{{21.76}}       & \multicolumn{1}{r}{{\ul 11.79}}       & \multicolumn{1}{r}{\textbf{46.74}}    & \multicolumn{1}{r}{\textbf{82.49}}    & \multicolumn{1}{r}{\textbf{73.59}}    & \multicolumn{1}{r}{{\ul 41.21}}       & \multicolumn{1}{r|}{\textbf{56.13}}    & {\ul 53.72}                           & {\ul }                           \\
 & TGA-ZSR~\citep{yu2024text} & \textbf{76.60}                        & 79.18 & 47.37 & 90.65 & 43.10 & 38.90 & 68.44 & 39.81 & 25.69 & 19.70 & 8.82  & 39.27 & 76.42 & 66.31 & 28.44 & 49.92 & 49.91 &  \\
&  {Comp-TGA~\citep{yu2026complementary}} & {75.00} & {\textbf{81.91}} & {\ul 50.67} & {90.84} & {\ul 46.45} & {44.28} & {72.04} & {41.41} & {28.35} & {\ul 23.50} & {9.12} & {42.23} & {78.96} & {69.25} & {29.55} & {49.88} & {52.09} \\

\rowcolor[HTML]{faebd7}
{\cellcolor[HTML]{FFFFFF}\multirow{-6}{*}{}} & UCAT (Ours)                       & {\ul 74.46}                           & {\ul {81.81}}                  & \textbf{54.45}                        & {\ul 91.88}                           & 41.06 & 53.58 & 74.16 & 47.57 & {\ul 31.92}                           & 19.29 & 10.95 & {\ul 43.20}                           & {\ul 82.39}                           & 71.53 & 37.32 & {\ul 51.20}                           & \textbf{54.17}                        & \textbf{}                        \\

\midrule
\rowcolor[HTML]{EDEDED}
{\cellcolor[HTML]{FFFFFF}\multirow{7}{*}{\rotatebox{90}{PGD}}} & CLIP~\citep{radford2021learning} & {\color[HTML]{757171} \textbf{0.19}}  & {\color[HTML]{757171} \textbf{9.57}}  & {\color[HTML]{757171} \textbf{3.07}}  & {\color[HTML]{757171} \textbf{23.64}} & {\color[HTML]{757171} \textbf{0.62}}  & {\color[HTML]{757171} \textbf{0.34}}  & {\color[HTML]{757171} \textbf{0.64}}  & {\color[HTML]{757171} \textbf{1.62}}  & {\color[HTML]{757171} \textbf{2.22}}  & {\color[HTML]{757171} \textbf{0.00}}  & {\color[HTML]{757171} \textbf{0.00}}  & {\color[HTML]{757171} \textbf{0.48}}  & {\color[HTML]{757171} \textbf{5.65}}  & {\color[HTML]{757171} \textbf{7.19}}  & {\color[HTML]{757171} \textbf{0.02}}  & {\color[HTML]{757171} \textbf{0.06}}  & {\color[HTML]{757171} \textbf{3.46}}  & {\color[HTML]{757171} \textbf{6.56}}\\
 & TeCoA~\citep{mao2022understanding}& {\ul 50.96}& 39.33& 21.64& 69.78& 20.07& 13.50& 37.80& 19.17& 18.30& \textbf{11.88}& 2.16 & 18.47& 56.00& 42.38& 9.33 & 46.92& 29.86& 35.52 \\
 & FARE~\citep{schlarmann2024robust} & \multicolumn{1}{r}{3.78} & \multicolumn{1}{r}{7.83} & \multicolumn{1}{r}{2.80} & \multicolumn{1}{r}{48.18}& \multicolumn{1}{r}{5.66} & \multicolumn{1}{r}{2.45} & \multicolumn{1}{r}{10.93}& \multicolumn{1}{r}{6.52} & \multicolumn{1}{r}{5.75} & \multicolumn{1}{r}{0.08} & \multicolumn{1}{r}{0.54} & \multicolumn{1}{r}{5.20} & \multicolumn{1}{r}{33.21}& \multicolumn{1}{r}{20.70}& \multicolumn{1}{r}{2.31} & \multicolumn{1}{r|}{48.97}& 12.81& 20.63 \\
 & PMG-AFT~\citep{wang2024pre} & \multicolumn{1}{r}{19.18}& \textbf{51.39}& {\ul 27.23}& {\ul 72.63}& 20.05& {\ul 16.88}& {\ul 44.59}& {\ul 26.43}& \textbf{20.05}& 11.49& {\ul 3.21} & 18.09& {\ul 61.13}& 43.46& {\ul 14.80}& \textbf{55.52}& {\ul 31.63}& {\ul 39.82}\\
 & TGA-ZSR~\citep{yu2024text} & \multicolumn{1}{r}{\textbf{50.68}} & \multicolumn{1}{r}{42.16}& \multicolumn{1}{r}{22.82}& \multicolumn{1}{r}{72.18}& \multicolumn{1}{r}{{\ul 21.57}} & \multicolumn{1}{r}{16.53}& \multicolumn{1}{r}{39.96}& \multicolumn{1}{r}{22.44}& \multicolumn{1}{r}{17.82}& \multicolumn{1}{r}{{\ul 11.75}} & \multicolumn{1}{r}{2.88} & \multicolumn{1}{r}{{\ul 20.39}} & \multicolumn{1}{r}{{ 58.05}} & \multicolumn{1}{r}{{ 46.18}} & \multicolumn{1}{r}{11.40}& \multicolumn{1}{r|}{{\ul 48.05}} & 31.55& 38.66 \\
&  {Comp-TGA~\citep{yu2026complementary}} & {49.90} & {40.44} & {22.43} & {72.56} & {\textbf{21.61}} & {16.81} & {41.56} & {23.04} & {18.88} & {11.66} & {2.70} & {20.07} & {58.80} & {\ul 46.42} & {10.50} & {44.98} & {31.40} & {39.18} \\
   
\rowcolor[HTML]{faebd7}
{\cellcolor[HTML]{FFFFFF}\multirow{-6}{*}{}} & UCAT (Ours) & 47.56& {\ul 43.81}& \textbf{25.16}& \textbf{73.83}& {20.44}& \textbf{22.86}& \textbf{45.11}& \textbf{26.79}& {\ul 19.47}& 2.99 & \textbf{3.45} & \textbf{22.22}& \textbf{65.32}& \textbf{50.47}& \textbf{15.30}& 30.37& \textbf{32.20}& \textbf{40.39}  \\

\midrule
\rowcolor[HTML]{EDEDED}
{\cellcolor[HTML]{FFFFFF}\multirow{7}{*}{\rotatebox{90}{CW}}}& CLIP~\citep{radford2021learning}& {\color[HTML]{757171} \textbf{0.14}}  & {\color[HTML]{757171} \textbf{9.91}}  & {\color[HTML]{757171} \textbf{3.34}}  & {\color[HTML]{757171} \textbf{26.01}} & {\color[HTML]{757171} \textbf{1.16}}  & {\color[HTML]{757171} \textbf{0.51}}  & {\color[HTML]{757171} \textbf{0.87}}  & {\color[HTML]{757171} \textbf{2.03}}  & {\color[HTML]{757171} \textbf{2.55}}  & {\color[HTML]{757171} \textbf{0.01}}  & {\color[HTML]{757171} \textbf{0.00}}  & {\color[HTML]{757171} \textbf{1.10}}  & {\color[HTML]{757171} \textbf{6.82}}  & {\color[HTML]{757171} \textbf{8.17}}  & {\color[HTML]{757171} \textbf{2.32}}  & {\color[HTML]{757171} \textbf{0.04}}  & {\color[HTML]{757171} \textbf{4.06}}  & {\color[HTML]{757171} \textbf{7.64}}\\
 & TeCoA~\citep{mao2022understanding}& {\ul 50.16} & 38.62 & 20.76 & 69.55 & 18.84 & 12.46 & 37.37 & 18.12 & {17.23} & \textbf{11.63} & 2.10  & 17.70 & 55.62 & 41.70 & 9.23  & {\ul 46.88} & 29.25 & 35.08  \\
 & FARE~\citep{schlarmann2024robust} & \multicolumn{1}{r}{4.10}  & \multicolumn{1}{r}{4.12}  & \multicolumn{1}{r}{2.96}  & \multicolumn{1}{r}{43.35} & \multicolumn{1}{r}{6.07}  & \multicolumn{1}{r}{3.17}  & \multicolumn{1}{r}{15.15} & \multicolumn{1}{r}{5.66}  & \multicolumn{1}{r}{4.52}  & \multicolumn{1}{r}{0.12}  & \multicolumn{1}{r}{1.11}  & \multicolumn{1}{r}{5.34}  & \multicolumn{1}{r}{32.50} & \multicolumn{1}{r}{20.85} & \multicolumn{1}{r}{4.38}  & \multicolumn{1}{r|}{48.86} & 12.64 & 20.41  \\
 & PMG-AFT~\citep{wang2024pre} & \multicolumn{1}{r}{13.16} & \multicolumn{1}{r}{42.10} & \multicolumn{1}{r}{21.31} & \multicolumn{1}{r}{65.69} & \multicolumn{1}{r}{13.12} & \multicolumn{1}{r}{11.43} & \multicolumn{1}{r}{28.05} & \multicolumn{1}{r}{17.53} & \multicolumn{1}{r}{12.55} & \multicolumn{1}{r}{8.51}  & \multicolumn{1}{r}{0.99}  & \multicolumn{1}{r}{11.72} & \multicolumn{1}{r}{52.84} & \multicolumn{1}{r}{35.68} & \multicolumn{1}{r}{7.06}  & \multicolumn{1}{r|}{14.26} & 22.25 & 31.47  \\
 & TGA-ZSR~\citep{yu2024text} & \multicolumn{1}{r}{\textbf{50.80}} & {\ul 42.24} & {\ul 22.64} & { 71.99} & {\ul 20.83} & {\ul 16.03} & {40.20} & { 21.52} & 16.97 & {\ul 11.56} & {\ul 2.85}  & {\ul 20.01} & {57.72} & {45.84} & {\ul 11.23} & \textbf{48.03} & {\ul 31.28} & {\ul 38.46}  \\

 & {Comp-TGA~\citep{yu2026complementary}} & {49.82} & {40.40} & {22.03} & {\ul 72.33} & {\textbf{21.13}} & {15.95} & {\ul 42.25} & {\ul 22.23} & {\ul 17.55} & {11.41} & {2.64} & {19.94} & {\ul 58.66} & {\ul 46.13} & {11.09} & {45.01} & {31.16} & {38.99} \\
 
\rowcolor[HTML]{faebd7}
{\cellcolor[HTML]{FFFFFF}\multirow{-6}{*}{}} & UCAT (Ours)& 47.08 & \textbf{43.30} & \textbf{23.92} & \textbf{73.55} & {19.20} & \textbf{21.68} & \textbf{45.38} & \textbf{24.95} & \textbf{17.87} & 2.41  & \textbf{3.21}  & \textbf{21.14} & \textbf{64.63} & \textbf{49.54} & \textbf{14.75} & 29.89 & \textbf{31.41} & \textbf{39.76}  \\

\midrule
\rowcolor[HTML]{EDEDED}
{\cellcolor[HTML]{FFFFFF}\multirow{7}{*}{\rotatebox{90}{Auto Attack}}} & CLIP~\citep{radford2021learning}     & {\color[HTML]{757171} \textbf{0.00}}  & {\color[HTML]{757171} \textbf{2.54}}  & {\color[HTML]{757171} \textbf{1.11}}  & {\color[HTML]{757171} \textbf{3.18}}  & {\color[HTML]{757171} \textbf{0.05}}  & {\color[HTML]{757171} \textbf{0.03}}  & {\color[HTML]{757171} \textbf{0.03}}  & {\color[HTML]{757171} \textbf{0.02}}  & {\color[HTML]{757171} \textbf{0.19}}  & {\color[HTML]{757171} \textbf{0.17}}  & {\color[HTML]{757171} \textbf{0.23}}  & {\color[HTML]{757171} \textbf{0.04}}  & {\color[HTML]{757171} \textbf{0.10}}  & {\color[HTML]{757171} \textbf{0.26}}  & {\color[HTML]{757171} \textbf{0.07}}  & {\color[HTML]{757171} \textbf{0.12}}  & {\color[HTML]{757171} \textbf{0.51}}     & {\color[HTML]{757171} \textbf{1.01}}  \\
 & TeCoA~\citep{mao2022understanding}  & \textbf{49.44}  & 37.87  & 20.45  & 69.31  & 17.41  & 12.19  & 36.58  & 17.81  & {\ul 17.29}  & \textbf{11.42}  & 1.86   & 17.19  & 54.95  & 41.19  & 8.16   & {\ul 46.79}  & 28.74  & 34.72   \\
 & FARE~\citep{schlarmann2024robust} & \multicolumn{1}{r}{0.12}  & \multicolumn{1}{r}{0.03}  & \multicolumn{1}{r}{0.21}  & \multicolumn{1}{r}{10.18} & \multicolumn{1}{r}{0.84}  & \multicolumn{1}{r}{0.19}  & \multicolumn{1}{r}{0.93}  & \multicolumn{1}{r}{0.60}  & \multicolumn{1}{r}{1.92}  & \multicolumn{1}{r}{0.07}  & \multicolumn{1}{r}{0.06}  & \multicolumn{1}{r}{0.86}  & \multicolumn{1}{r}{10.26} & \multicolumn{1}{r}{5.59}  & \multicolumn{1}{r}{0.21}  & \multicolumn{1}{r|}{5.15}  & 2.33   & 4.45    \\
 & PMG-AFT~\citep{wang2024pre}    & \multicolumn{1}{r}{8.22}  & {\ul 41.86}  & 21.18  & 65.45  & 7.95   & 7.34   & 18.94  & 12.59  & 3.13   & 7.17   & 0.51   & 7.90   & 44.91  & 28.29  & 3.22   & 7.41   & 17.88  & 26.83   \\
 & TGA-ZSR~\citep{yu2024text} & \multicolumn{1}{r}{{\ul 49.26}} & 40.92  & {\ul 21.75}  & {\ul 71.55}  & \textbf{19.88}  & {\ul 15.32}  & {\ul 38.84}  & {\ul 20.98}  & 17.02  & {\ul 11.26}  & {\ul 2.34}   & {\ul 19.12}  & {\ul 57.11}  & {\ul 45.16}  & {\ul 9.87}   & \textbf{48.00}  & {\ul 30.52}  & {\ul 37.88}    \\

  & {Comp-TGA~\citep{yu2026complementary}} & {39.84} & {30.11} & {17.23} & {65.48} & {15.70} & {12.50} & {34.10} & {18.21} & {14.84} & {10.82} & {1.74} & {15.87} & {52.61} & {40.69} & {7.38} & {42.81} & {26.24} & {34.90} \\
  
\rowcolor[HTML]{faebd7}
{\cellcolor[HTML]{FFFFFF}\multirow{-6}{*}{}} & UCAT (Ours)     & 45.80  & \textbf{42.32}  & \textbf{23.03}  & \textbf{73.15}  & {\ul 18.26}  & \textbf{20.52}  & \textbf{44.02}  & \textbf{24.54}  & \textbf{18.14}  & 2.26   & \textbf{2.61}   & \textbf{20.15}  & \textbf{63.73}  & \textbf{48.66}  & \textbf{12.60}  & 29.51  & \textbf{30.58}  & \textbf{39.09} \\
\bottomrule
\end{tabular}}
\label{tab:tinyimagenet-eps1}
\vspace{-1em}
\end{table*}

%% file: Tab/VLM-main.tex
\begin{table}[t]
\centering
\caption{\textbf{Generalization across contrastive VLM backbones.}
We report the mean clean accuracy, AutoAttack robustness ($\ell_\infty$, $\epsilon=1/255$), and harmonic mean $H$ over 16 datasets when fine-tuning different contrastively pretrained VLMs on TinyImageNet (2-step PGD, $\epsilon=1/255$). Improvements of UCAT over the corresponding base model are shown in \textcolor{violet!70!black}{violet}.
}
\label{tab:vlm-main}
\vspace{-0.5em}
\resizebox{\linewidth}{!}{
\begin{tabular}{l|l|ccc}
\toprule
Backbone & Method & Clean & AutoAttack & $H$ \\
\midrule
 & Base & 63.72 & 0.01 & 0.02 \\
\rowcolor[HTML]{faebd7}
{\cellcolor[HTML]{FFFFFF}\multirow{-2}{*}{CLIP-B/16~\citep{radford2021learning}}} & +UCAT & 52.91 & 30.54 {\textcolor{violet!70!black}{\scriptsize (+30.53)}} & 39.05 {\textcolor{violet!70!black}{\scriptsize (+39.03)}} \\
\midrule
 & Base & 64.42 & 5.58 & 10.28 \\
\rowcolor[HTML]{faebd7}
{\cellcolor[HTML]{FFFFFF}\multirow{-2}{*}{CLIP-B/32~\citep{radford2021learning}}} & +UCAT & 54.17 & 30.58 {\textcolor{violet!70!black}{\scriptsize (+25.00)}} & 39.09 {\textcolor{violet!70!black}{\scriptsize (+28.81)}} \\
\midrule
 & Base & 46.03 & 0.02 & 0.04 \\
\rowcolor[HTML]{faebd7}
{\cellcolor[HTML]{FFFFFF}\multirow{-2}{*}{SLIP-B/16~\citep{mu2022slip}}} & +UCAT & 38.37 & 20.40 {\textcolor{violet!70!black}{\scriptsize (+20.38)}} & 26.68 {\textcolor{violet!70!black}{\scriptsize (+26.64)}} \\
\bottomrule
\end{tabular}}
\vspace{-1em}
\end{table}

%% file: Tab/Abl.tex
\begin{table}[t]
\caption{\textbf{Ablation study.} Trained on TinyImageNet with 1-step PGD and evaluated under 100-step PGD, CW, and AutoAttack (AA) with $\epsilon=1/255$. 
% We compare $\mathcal{L}_{\text{ce}}$, probability-level KL, and Dirichlet-level KL alignment. 
Results are averaged over 16 datasets. 
Best and second-best are in \textbf{bold} and \underline{underline}.}
\label{tab:abl}
\centering
\resizebox{\linewidth}{!}{
\begin{tabular}{l|rrrr}
\toprule
Methods & Clean & PGD & CW & AA \\
\midrule
\rowcolor[HTML]{EDEDED}
CLIP & {\color[HTML]{757171} \textbf{64.45}} & {\color[HTML]{757171} \textbf{0.05}} & {\color[HTML]{757171} \textbf{4.06}} & {\color[HTML]{757171} \textbf{0.51}} \\
$\mathcal{L}_{\text{ce}}$ & 43.83 & 29.86 & 29.25 & 28.74 \\
{$\mathcal{L}_{\text{ce}}$ + $\text{KL}(p(x)\|p(x^a))$} & {45.03} & {\ul 30.12} & {\ul 29.61} & {\ul 29.13} \\
$\mathcal{L}_{\text{ce}}$ + $\text{KL}(p(x^a)\|p(x))$ & {\ul 45.05} & {29.98} & {29.28} & {28.80} \\
{$\mathcal{L}_{\text{ce}}$ + $\text{KL}(\text{Dir}(\alpha)\|\text{Dir}(\alpha_{adv}))$} & 36.72 &	25.01 &	24.66	& 24.36 \\
\rowcolor[HTML]{faebd7}
$\mathcal{L}_{\text{ce}}$ + $\text{KL}(\text{Dir}(\alpha_{adv})\|\text{Dir}(\alpha))$ & \textbf{54.17} & \textbf{32.20} & \textbf{31.41} & \textbf{30.58} \\
\bottomrule
\end{tabular}} 
\vspace{-1em}
\end{table}
% \vspace{-0.5em}
% \end{wraptable}

%% file: main/7_con.tex
\section{Conclusion}

In this paper, we identified that adversarial perturbations in zero-shot CLIP not only reduce accuracy but also often suppress predictive uncertainty, leading to severe miscalibration. To address this, we reformulated CLIP logits as Dirichlet concentration parameters, yielding a representation that preserves both semantic structure and confidence strength. Building on this foundation, we introduced an uncertainty calibration adversarial finetuning method that aligns the Dirichlet distributions of clean and perturbed samples, ensuring robustness preservation and calibrated uncertainty. Extensive experiments demonstrate that our approach improves adversarial robustness, handles data ambiguity, and provides reliable uncertainty estimates. Beyond CLIP, our contrastive-theoretic perspective suggests a principled way to analyze and extend uncertainty modeling to other contrastive learning frameworks.

\section*{Impact Statement}
This work focuses on methodological advances in uncertainty modeling and robustness evaluation. The proposed approach is intended for research purposes and aims to improve understanding of model behavior under adversarial perturbations and multi-label ambiguity. While such insights may inform the design of more reliable learning systems, any deployment in real-world applications should be accompanied by thorough risk assessment and additional safeguards.

\section*{Acknowledgment}

% This work is supported in part by the National Natural Science Foundation
% of China (No. 62272300) and the JSPS Bilateral Program (No. JPJSBP120257420).

This work was supported by the National Key R\&D Program of China (No. 2023YFC2811500), the National Natural Science Foundation of China (No. 62272300), and the JSPS Bilateral Program (No. JPJSBP120257420).

%% file: Appendix/2-Imp.tex
\section{Implementation Details} \label{app:imp}

\noindent\textbf{Datasets and evaluation suite.}
We use the same zero-shot evaluation suite as prior zero-shot adversarial robustness (ZSAR) works (e.g., \citet{mao2022understanding}).
Specifically, we evaluate on ImageNet/tinyImageNet\citep{deng2009imagenet}, CIFAR10/100~\citep{krizhevsky2009learning}, STL10~\citep{coates2011analysis}, Caltech101~\citep{fei2004learning}, Caltech256~\citep{griffin2007caltech}, OxfordPets~\citep{6248092}, StanfordCars~\citep{krause20133d}, Food101~\citep{bossard2014food}, Flowers102~\citep{nilsback2008automated}, FGVC-Aircraft~\citep{maji2013fine}, SUN397~\citep{xiao2010sun}, DTD~\citep{cimpoi2014describing}, and two domain-specialized sets PCAM~\citep{veeling2018rotation} and EuroSAT~\citep{helber2019eurosat}. To further assess robustness under semantic ambiguity, we additionally include the multi-label dataset MS-COCO~\citep{lin2014microsoft}.

\noindent\textbf{Backbone and training setup.}
We adopt CLIP-B/32~\citep{radford2021learning} as the backbone and follow TeCoA’s optimizer and training schedule~\citep{mao2022understanding}, using a batch size of 256 and 10 training epochs unless otherwise stated. We benchmark five methods: CLIP~\citep{radford2021learning}, TeCoA~\citep{mao2022understanding}, FARE~\citep{schlarmann2024robust}, PMG-AFT~\citep{wang2024pre}, TGA-ZSR~\citep{yu2024text}, and {Comp-TGA~\citep{yu2026complementary}}.  

\noindent\textbf{Adversarial attacks.}
For \textit{training-time attacks,}
we adopt two regimes: (i) a light regime following TeCoA~\citep{mao2022understanding}, using $\ell_\infty$ PGD-2 with $\varepsilon=1/255$; and (ii) a stronger regime following FARE~\citep{schlarmann2024robust}, using $\ell_\infty$ PGD-10 with $\varepsilon=2/255$, {where both regimes maintain a constant step size of $\alpha=1/255$}.  
For \textit{evaluation attacks,} robustness is further assessed using $\ell_\infty$ PGD-100~\citep{madry2017towards} (with the same $\varepsilon$ as the training regime and {a fixed step size of $\alpha=1/255$}), CW-100~\citep{carlini2017towards}, and AutoAttack~\citep{croce2020reliable} (the rand version ensembling APGD-CE and APGD-DLR).

\noindent\textbf{Loss weights.}
We set $\lambda=10^{5}/\beta$ with $\beta=2/e^{\tau'}$, where $\tau'=0.07$ follows standard contrastive learning practices~\citep{wu2018unsupervised,he2020momentum,radford2021learning,yeh2022decoupled}. Here $\beta$ corresponds to the upper bound of the mapping function $h(\ell)$ that converts logits $\ell$ into non-negative evidence. Using this bound guarantees that $\lambda$ remains numerically stable across different temperature values, preventing uncontrolled scaling when $\tau'$ varies.

%% file: Appendix/3-Provement.tex
\section{Proofs}\label{app:prove}

% \subsection{Lemma 1: Validity of Dirichlet Evidence}\label{app:lemma1}

% \begin{lemma}[Validity of Dirichlet Evidence]\label{app-lemma:validity}
% Under Definition~\ref{definition}, for all $k$:
% \begin{enumerate}\itemsep0pt
% \item $\alpha_k(x)\ge 1$ and $\alpha_k(x)\in[1,\exp(2/\tau')]$;
% \item $\alpha=\exp(h(\ell))$ is strictly increasing.
% \end{enumerate}
% \end{lemma}
\subsection{Proof of Lemma~\ref{lemma:validity} (Validity of Dirichlet Evidence)}\label{app:lemma1}

% \noindent\textit{Proof.}
\begin{proof}
Since $\|v(x)\|_2=\|t_k\|_2=1$, we have $\langle v(x),t_k\rangle\in[-1,1]$.
By the logit definition, $\tau\,\ell_k^{v\to t}(x)=\langle v(x),t_k\rangle\in[-1,1]$.
Therefore,
\[
h(\ell_k^{v\to t}(x))=\frac{\tau\,\ell_k^{v\to t}(x)+1}{\tau'}\in\Big[\,0,\ \frac{2}{\tau'}\,\Big].
\]
Exponentiating yields
\[
\alpha_k(x)=\exp\!\big(h(\ell_k^{v\to t}(x))\big)\in\big[\,e^{0},\ e^{2/\tau'}\,\big]=\big[\,1,\ \exp(2/\tau')\,\big],
\]
and both endpoints are attainable when $\langle v(x),t_k\rangle=-1$ and $+1$, respectively.

For monotonicity, differentiate $\alpha_k(x)$ with respect to $\ell_k^{v\to t}(x)$:
\[
\frac{d\,\alpha_k(x)}{d\,\ell_k^{v\to t}(x)}
=\frac{\tau}{\tau'}\,\exp\!\Big(\frac{\tau\,\ell_k^{v\to t}(x)+1}{\tau'}\Big)
=\frac{\tau}{\tau'}\,\alpha_k(x)
>0,
\]
since $\tau>0$, $\tau'>0$, and $\alpha_k(x)>0$. Hence $\alpha_k$ is strictly
increasing in $\ell_k^{v\to t}$, which preserves both strict and non-strict
order between any pair of logits.
\end{proof}

% \subsection{Lemma 2: Consistency with Dirichlet Expectations}\label{app:lemma2}
% \begin{lemma}[Exact Equivalence at $\tau=\tau'$]\label{app-lemma:equiv-tau}
% Let $s=\tau/\tau'$. If $s=1$ (equivalently $\tau'=\tau$), the Dirichlet expectation equals to CLIP’s softmax:
% \[
% p^{\mathrm{Dir}}_k(x) \;=\; \frac{\alpha_k}{\sum_j \alpha_j}
% = \frac{\exp(h(\ell_k))}{\sum_j \exp(h(\ell_j))}
% = \operatorname{softmax}\!\big(\ell(x)\big)_k
% = p^{\mathrm{CLIP}}_k(x).
% \]
% \end{lemma}
\subsection{Proof of Lemma~\ref{lemma:equiv-tau} (Exact Equivalence at $\tau=\tau'$)}\label{app:lemma2}

% \noindent\textit{Proof.}
\begin{proof}
From the definition of the Dirichlet expectation in Equation~\ref{eq:dir-exp-softmax},
\[
p_k^{\mathrm{Dir}}(x)
=\mathbb{E}_{\pi\sim\mathrm{Dir}(\alpha(x))}[\pi_k]
=\frac{\alpha_k(x)}{\alpha_0(x)},
\quad
\alpha_0(x)=\sum_{j=1}^C \alpha_j(x).
\]
By construction,
\[
\alpha_k(x)=\exp(h(\ell_k^{v\to t}(x))), \quad
h(\ell_k^{v\to t}(x))=\frac{\tau \ell_k^{v\to t}(x) + 1}{\tau'}
= \frac{1}{\tau'} + \frac{\tau}{\tau'} \ell_k^{v\to t}(x).
\]
Let $s = \tau/\tau' > 0$. Then
\[
p_k^{\mathrm{Dir}}(x)
= \frac{\exp(1/\tau'+s\,\ell_k^{v\to t}(x))}
       {\sum_{j=1}^C \exp(1/\tau'+s\,\ell_j^{v\to t}(x))}
= \frac{\exp(s\,\ell_k^{v\to t}(x))}
       {\sum_{j=1}^C \exp(s\,\ell_j^{v\to t}(x))}
= \operatorname{softmax}(s\,\ell^{v\to t}(x))_k,
\]
since the additive constant $1/\tau'$ cancels out. When $s=1$ (equivalently, $\tau'=\tau$), this reduces to
\[
p_k^{\mathrm{Dir}}(x)=\operatorname{softmax}(\ell^{v\to t}(x))_k,
\]
which matches exactly the original CLIP prediction $p_k^{\mathrm{CLIP}}(x)$. 
    
\end{proof}
% For each \(k\),
% \[
% \alpha_k(x)=\exp\!\Big(\tfrac{\tau}{\tau'}\ell^{v\to t}_k(x)+\tfrac{1}{\tau'}\Big)
% = e^{1/\tau'}\,\exp\!\big(s\,\ell^{v\to t}_k(x)\big).
% \]
% Thus
% \[
% p^{\text{Dir}}_k(x)
% =\frac{e^{1/\tau'}\exp(s\,\ell^{v\to t}_k(x))}{\sum_j e^{1/\tau'}\exp(s\,\ell^{v\to t}_j(x))}
% =\frac{\exp(s\,\ell^{v\to t}_k(x))}{\sum_j \exp(s\,\ell^{v\to t}_j(x))}
% = \operatorname{softmax}\!\big(s\,\ell^{v\to t}(x)\big)_k.
% \]
% Since \(p^{\text{CLIP}}_k(x)=\operatorname{softmax}(\ell^{v\to t}(x))_k\),
% \[
% \operatorname{softmax}(s\ell^{v\to t}(x) )_k=\frac{(\exp(\ell_k^{v\to t}(x)))^{s}}{\sum_j(\exp(\ell_j^{v\to t}(x)))^{s}}
% =\frac{\big(p^{\text{CLIP}}_k(x)\big)^{s}}{\sum_j \big(p^{\text{CLIP}}_j(x)\big)^{s}},
% \]
% hence \(p^{\text{Dir}}_k(x)\propto \big(p^{\text{CLIP}}_k(x)\big)^{s}\).
% % \hfill\(\square\)

\subsection{Proof of Corollary~\ref{cor:scale} (General form and invariances)}\label{app-cor:scale}
% \begin{corollary}[General form and invariances]\label{app-cor:scale}
% For arbitrary $\tau'>0, s=\tau/\tau'>0$, $p^{\mathrm{Dir}}(x)=\operatorname{softmax}\big(s\,\ell(x)\big).$ Hence 
% \[
%  \arg\max_k p^{\mathrm{Dir}}_k(x)=\arg\max_k p^{\mathrm{CLIP}}_k(x)
% \]
% while the entropy of the distribution can be smoothly tuned by $s$: larger $s$ yields sharper predictions, smaller $s$ yields flatter ones.
% \end{corollary}

% \noindent\textit{Proof.}
\begin{proof}
For any logits $\ell\in\mathbb{R}^C$ and scalar $s>0$, 
\[
    \arg\max_k \ell_k = \arg\max_k s\ell_k.
\]
Since the softmax assigns the maximum probability to the index with maximum input, we have
\[
\arg\max_k\, p^{\text{CLIP}}_k(x) = \arg\max_k\, p^{\text{Dir}}_k(x).
\]
Thus both distributions yield the same classification decision, proving the accuray invariance.

For calibaration control, observe that $p^{\text{Dir}}_k(x)=e^{s\ell_k}/\sum_j e^{s\ell_j}$ becomes increasingly peaked as $s\to\infty$, converging to a one-hot vector, and tends to the uniform distribution as $s\to 0^+$. The entropy
\[
H(p^{\text{Dir}}(x))=-\sum_k p^{\text{Dir}}_k(x)\log p^{\text{Dir}}_k(x)
\]
decreases monotonically with $s$. Thus $s$ leaves classification accuracy unchanged while directly modulating the calibration of predictive confidence.
\end{proof}

%% file: Appendix/4-PU.tex
\section{Extended Uncertainty Analysis}

\subsection{Implementation Details for Uncertainty Quantification}
Recall the decomposition of predictive uncertainty under the Dirichlet parameterization into aleatoric uncertainty (AU) and epistemic uncertainty (EU) in Section~\ref{sec:edl-define}. 
\[
\mathrm{AU}(x) 
= \mathbb{E}_{\pi \sim \mathrm{Dir}(\alpha)}\!\big[ H(\pi) \big] 
= - \sum_{k=1}^C \frac{\alpha_k}{\alpha_0} 
  \Big( \psi(\alpha_k + 1) - \psi(\alpha_0 + 1) \Big),
  \quad
  \mathrm{EU}(x) = \frac{C}{\alpha_0 + C}.
\]
Our reformulation $\alpha_k(x) = \exp\big(h(\ell^{v\to t}_k(x))\big),\,
h(\ell)=\frac{\tau\,\ell+1}{\tau'},$ adopts a linear definition of the evidence mapping $h(\ell)$, for which Section~\ref{sec:prove} and Appendix~\ref{app:prove} have established the theoretical equivalence between CLIP logits and Dirichlet distributions.

In practice, however, the learnable temperature coefficient $\tau$ may become very small during training (e.g., $\tau=0.01$), which leads to excessively large logits after exponentiation and renders the raw uncertainty values numerically unstable. To address this, we introduce an additional activation $h'(\ell)=\operatorname{softplus}(h(\ell))$, which is commonly adopted in EDL to smooth the outputs and map them into a numerically stable range suitable for analysis~\cite{sensoy2018evidential,malinin2018predictive}. 

Moreover, when $\tau$ is too small (e.g., $\tau=0.01$), EU degenerates towards $0$ and AU coincides with PU. To avoid this issue, we adopt $\tau=0.07$ for computing EU, while keeping $\tau=0.01$ for AU. This choice is theoretically acceptable: both the softplus mapping and the rescaling by $\tau$ affect only the magnitude of uncertainty values, not their ordering. As a result, the reliability of AUROC evaluation, which depends only on ranking, is unaffected. For ECE, we use PU directly computed from probabilities, which is independent of $\tau$ and activation adjustments. 

These practical adjustments ensure stable and meaningful AU/EU quantification without altering the comparative reliability of our uncertainty metrics.

\subsection{Additional Visualizations of Uncertainty}\label{app:puaueu}
% To complement the main results, we provide extended visualizations of predictive uncertainty under adversarial attacks. 
% Figure~\ref{fig:uncertainty-drop} reports the degradation of accuracy and predictive uncertainty (PU) across 16 datasets under three strong white-box attacks (PGD, CW, AutoAttack). 
% Figures~\ref{fig:au-16} and~\ref{fig:eu-16} further decompose the uncertainty into aleatoric and epistemic components, respectively, comparing CLIP with our method on both clean and adversarial samples. 
% These results illustrate how adversarial perturbations simultaneously reduce accuracy and distort uncertainty, while our method consistently provides more reliable AU/EU estimates across diverse datasets, thereby achieving effective uncertainty calibration.

To complement the main results, we provide extended visualizations that follow the analysis chain from \emph{overall predictive uncertainty} to its \emph{aleatoric/epistemic} decomposition.

\begin{figure}[htbp]
    \centering
    \includegraphics[width=\linewidth]{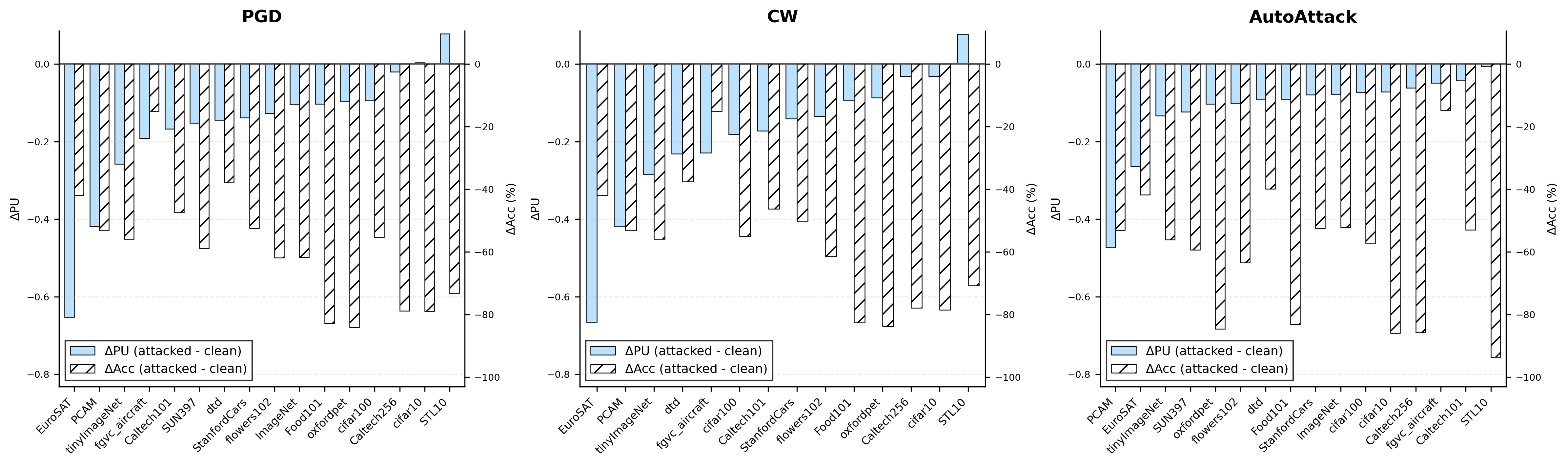}
    \caption{Effect of strong white-box attacks ($\epsilon = 1/255$, 100 steps) on accuracy and predictive uncertainty across 16 datasets. Each panel shows the change under a single attack type (left: PGD, center: CW, right: AutoAttack); for each dataset the filled light bars plot $\Delta\mathrm{PU}=\mathrm{PU}_{\text{attacked}}-\mathrm{PU}_{\text{clean}}$ (left axis) and the hatched bars plot $\Delta\mathrm{Acc}=\mathrm{Acc}_{\text{attacked}}-\mathrm{Acc}_{\text{clean}}$ in percentage points (right axis). Negative values therefore indicate decreases caused by the attack. Results demonstrate that all three attacks induce simultaneous drops in accuracy and predictive uncertainty on most datasets, with the magnitude of degradation varying by dataset and attack.
}
    \label{fig:uncertainty-drop}
\end{figure}

\noindent\textbf{Attack-wise robustness vs. predictive uncertainty shift.}
Figure~\ref{fig:uncertainty-drop} reports the joint change in accuracy and predictive uncertainty (PU) under three strong white-box attacks (PGD, CW, AutoAttack). The consistently negative $\Delta\mathrm{PU}$ for CLIP indicates that miscalibration is not tied to a specific attack, motivating uncertainty-aware regularization.

\begin{figure}[htbp]
    \centering
    \includegraphics[width=\linewidth]{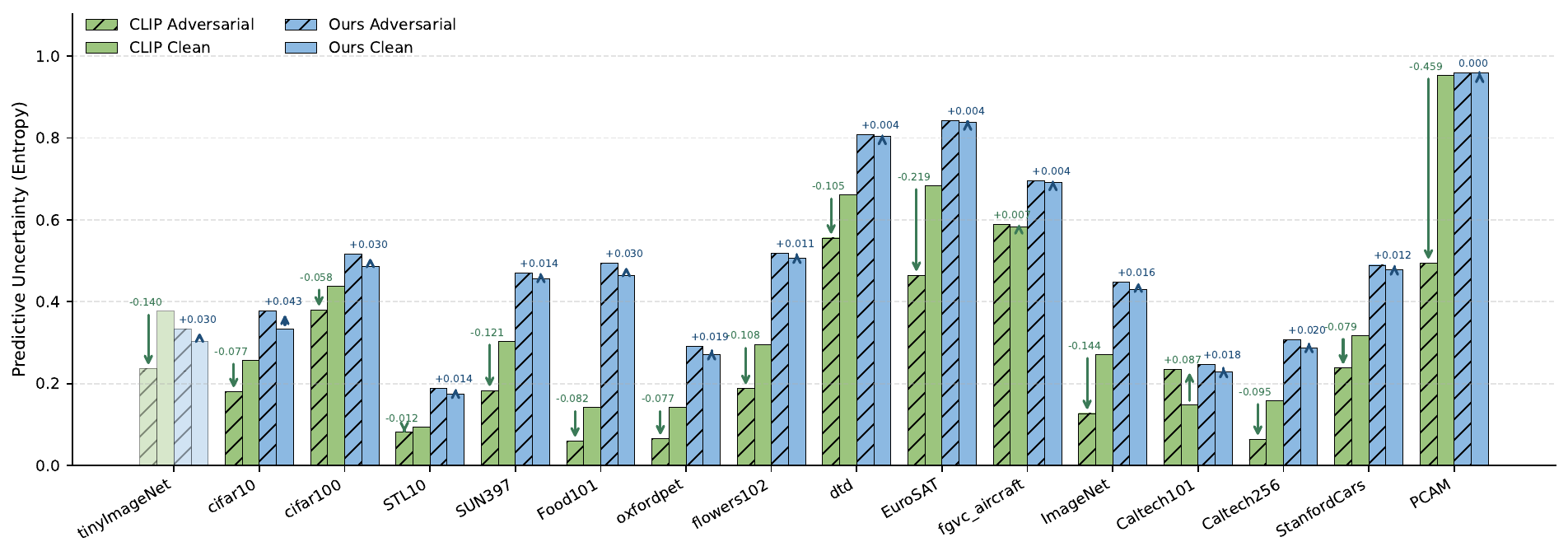}
    \caption{\textbf{Predictive uncertainty on 16 datasets}. \textcolor[HTML]{9cc57e}{CLIP} shows reduced entropy on adversarial inputs, whereas our method \textcolor[HTML]{8cb9e2}{UCAT} restores calibrated uncertainty. Arrows and numbers show uncertainty change (direction, magnitude). }
    \label{fig:pu_16}
\end{figure}

\noindent\textbf{Predictive uncertainty (PU).}
Figure~\ref{fig:pu_16} summarizes PU (entropy) on 16 datasets for clean and adversarial inputs. Across most datasets, CLIP exhibits \emph{uncertainty suppression} under attack (lower entropy on adversarial inputs), which manifests as over-confident yet incorrect predictions.
Our method mitigates this effect and yields a more consistent ordering between clean and adversarial uncertainty by explicitly regularizing evidence scale in addition to relative class structure.

\begin{figure}[htbp]
    \centering
    \includegraphics[width=\linewidth]{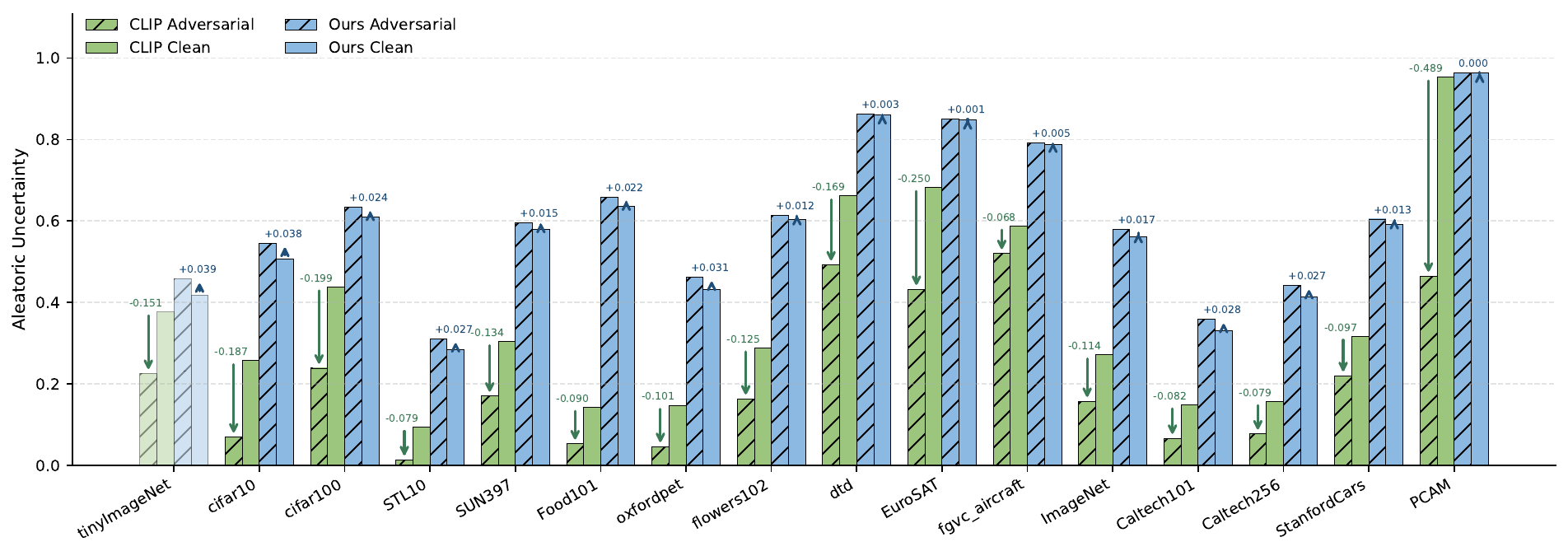}
    \caption{Comparison of \textbf{aleatoric uncertainty} on clean and adversarial samples across 16 datasets between CLIP and our method, adversarially trained on tinyImageNet under 10-step PGD with $\epsilon=2/255$.}
    \label{fig:au-16}
\end{figure}

\begin{figure}[htbp]
    \centering
    \includegraphics[width=\linewidth]{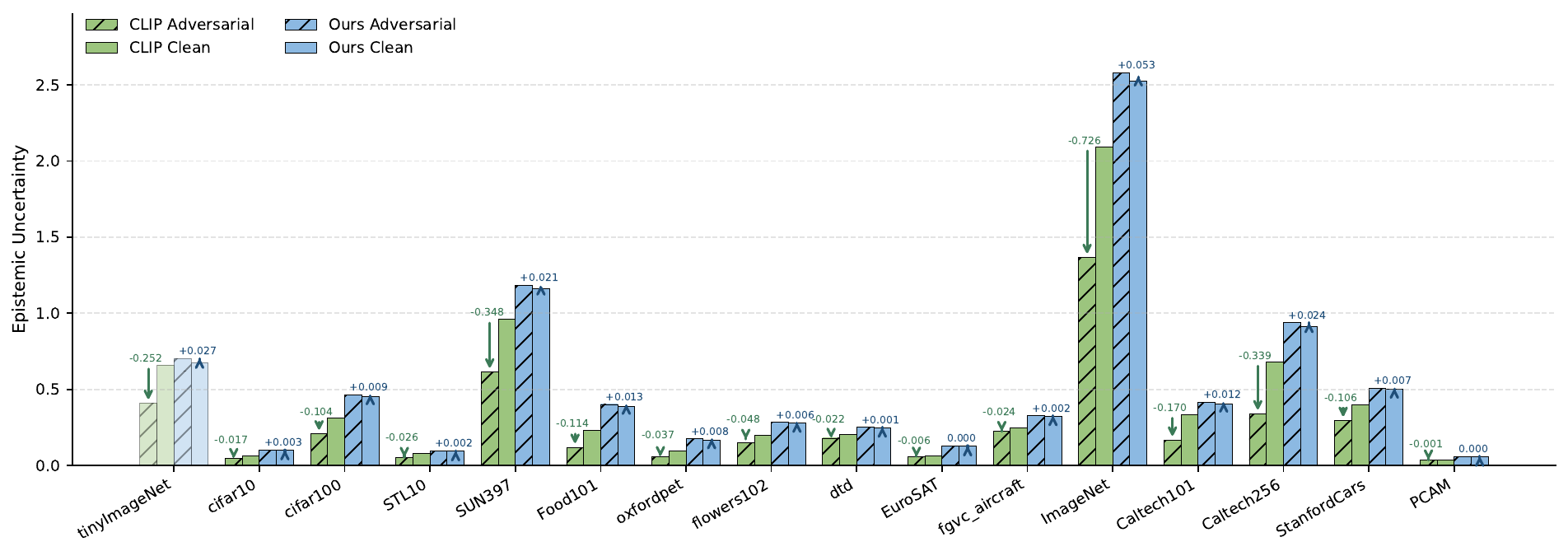}
    \caption{Comparison of \textbf{epistemic uncertainty} on clean and adversarial samples across 16 datasets between original CLIP and our method, adversarially trained on tinyImageNet under 10-step PGD with $\epsilon=2/255$.}
    \label{fig:eu-16}
\end{figure}

\noindent\textbf{Decomposition into AU/EU.}
Finally, Figures~\ref{fig:au-16} and~\ref{fig:eu-16} decompose uncertainty into aleatoric (AU) and epistemic (EU) components. Together, these plots show that adversarial perturbations can distort both class ambiguity (AU) and evidence strength (EU), whereas our method yields more reliable AU/EU behaviors across datasets.

\clearpage

%% file: Appendix/5-Exp.tex
\section{{Evaluation under Stronger Attacks, Larger Datasets, and Additional Ablations}}\label{app:exp}
% {We conduct four extended evaluations to further assess the generality, stability, and architectural robustness of UCAT beyond the main experiments.}

\subsection{{Multi-target PGD evaluation on multi-label MS-COCO}}\label{app:multilabel-pgd}
Here we additionally evaluate a stronger \emph{multi-label} PGD attack on MS-COCO.
Specifically, instead of attacking a single label at a time, we jointly maximize \texttt{BCEWithLogitsLoss} over all target labels, resulting in a \emph{joint multi-objective} adversarial perturbation that simultaneously degrades multiple semantic predictions.
As reported in Table~\ref{tab:multilabelPGD}, UCAT remains competitive and achieves the best (or tied-best) clean--robust trade-off in $H(\mathrm{F1@3})$ under moderate-to-large perturbation budgets (e.g., $\epsilon\in{2/255,4/255}$), suggesting that distributional (Dirichlet) alignment better preserves a stable multi-label neighborhood under stronger joint adversarial perturbations.
\input{Tab/MultiLabel-PGD-re}

\subsection{{Training under stronger adversarial attacks}}
{TeCoA~\citep{mao2022understanding} and FARE~\citep{schlarmann2024robust} represent two widely used adversarial fine-tuning configurations on TinyImageNet, with FARE adopting a substantially stronger perturbation budget. To ensure a fair and comprehensive comparison, we evaluate UCAT under this stronger training regime as well. Specifically, we follow the FARE configuration and train models using 10-step PGD with $\epsilon=2/255$.}

\input{Tab/tinyImageNet-esp2}

{We first report zero-shot robustness under {three standard white-box attacks (\emph{i.e.}, PGD, CW, and AutoAttack)} in Table~\ref{tab:tinyimagenet-eps2}, which measures performance against substantially stronger perturbations than those used in Table~\ref{tab:tinyimagenet-eps1}. 
To further assess generalization across attack types, we additionally evaluate CAA~\citep{mao2021composite} and A$^3$~\citep{liu2022practical}, two strong adaptive attacks.
 % (Table~\ref{tab:re-aa}). 
Together, these evaluations examine UCAT’s robustness under both stronger training perturbations and a wider range of test-time threat models.}

\subsection{{Training on a larger dataset}}
Following TeCoA~\citep{mao2022understanding}, we train on ImageNet-1k with 2-step PGD at $\epsilon=1/255$ to assess performance on a larger training dataset across 15 benchmarks (tinyImageNet is excluded, as it was not reported in TeCoA’s original paper). {As shown in Table~\ref{tab:imagnet}, this setting examines whether UCAT continues to benefit from its uncertainty-calibration mechanism when trained on large-scale data.}
\input{Tab/ImageNet-eps1}

\subsection{{Effect of the calibration coefficient $\tau^\prime$}} \label{re:sec-tauprime}
{As shown in Table~\ref{tab:re-tauprime}, We perform a controlled ablation by varying the Dirichlet calibration coefficient $\tau^\prime$ and reporting clean accuracy, AutoAttack robustness, and their harmonic mean $H$ across 16 datasets. This study isolates the influence of the evidence-scaling term in our formulation and identifies $\tau^\prime=0.07$ as the best operating point.}

\input{Tab/rebuttal-tauprime}
\input{Tab/rebuttal-VLM}

\subsection{{Generalization across vision–language models}}
{We further examine whether UCAT is tied to the CLIP-B/32 backbone. We fine-tune SLIP-B16~\citep{mu2022slip}, CLIP-B/16~\citep{radford2021learning}, and CLIP-B/32~\citep{radford2021learning} on TinyImageNet using the same zero-shot adversarial robustness (ZSAR) configuration (2-step PGD, $\epsilon=1/255$) in Table~\ref{tab:re-vlm}. This evaluates the architecture-agnostic nature of our Dirichlet alignment.}

\subsection{Comparison with classical adversarial training baselines}
{
Finally, we examine whether classical adversarial training methods (AT) can serve as a baseline for zero-shot adversarial robustness (ZSAR). For fairness, all AT baselines (TRADES~\citep{zhang2019theoretically}, ACAT~\citep{addepalli2022efficient}, DKL~\citep{cui2024decoupled}) are fine-tuned on TinyImageNet using the same 2-step PGD configuration ($\epsilon = 1/255$) with their \emph{native objective, supervision signal, and optimization hyperparameters}. These models are then evaluated in a zero-shot manner on downstream ZSAR benchmarks, where no task-specific labels are used during testing. Although these AT methods were not originally developed for zero-shot scenarios, they still achieve strong performance under ZSAR (Table~\ref{tab:re-at}), demonstrating the general effectiveness of their theoretical formulations. However, compared with these supervised AT approaches, our UCAT method, which is explicitly designed to preserve open-set semantic alignment, achieves consistently better overall ZSAR performance, particularly in terms of the harmonic mean between clean and robust accuracy.
}

\input{Tab/rebuttal-at}

Overall, our method remains consistently strong across both extended settings, confirming its robustness under {larger datasets, stronger adversarial attacks, hyperparameter variations, and different contrastive VLM architectures.}

%% file: Tab/MultiLabel-PGD-re.tex
% \usepackage[normalem]{ulem}
\begin{table*}[!ht] 
\centering 
\caption{\textbf{Zero-shot adversarial robustness on multi-label dataset MS-COCO~\citep{lin2014microsoft}.} 
All ZSAR models are adversarially trained on TinyImageNet with the FARE~\citep{schlarmann2024robust} 10-step PGD setting ($\epsilon=1/255$), and evaluated under \textbf{PGD}-100 at radii $\epsilon\in\{1/255,2/255,4/255\}$ plus clean. We report mean Average Precision (mAP), Precision (P), Recall (R), and F1-score (F1) at top-3 predictions. $H(\mathrm{F1@3})$ denotes the harmonic mean of clean and adversarial F1@3. Best and second-best are in \textbf{bold} and \underline{underline}.
}
\resizebox{\linewidth}{!}{
\begin{tabular}{l|rrrr|rrrrr|rrrrr|rrrrr}
\toprule
{\textbf{}} & \multicolumn{4}{c|}{Clean} & \multicolumn{5}{c|}{$\epsilon=1/255$}& \multicolumn{5}{c|}{$\epsilon=2/255$}& \multicolumn{5}{c}{$\epsilon=4/255$}\\

Methods & \multicolumn{1}{c}{{{mAP}}}  & \multicolumn{1}{c}{{{ P@3}}} & \multicolumn{1}{c}{{{ R@3}}} & \multicolumn{1}{c|}{{{ F1@3}}} 
& \multicolumn{1}{c}{{{mAP}}}  & \multicolumn{1}{c}{{{ P@3}}} & \multicolumn{1}{c}{{{ R@3}}} & \multicolumn{1}{c}{{{ F1@3}}} & \multicolumn{1}{c|}{H\scriptsize{(F1@3)}} 
& \multicolumn{1}{c}{{{mAP}}}  & \multicolumn{1}{c}{{{ P@3}}} & \multicolumn{1}{c}{{{ R@3}}} & \multicolumn{1}{c}{{{ F1@3}}} & \multicolumn{1}{c|}{H\scriptsize{(F1@3)}} 
& \multicolumn{1}{c}{{{mAP}}}  & \multicolumn{1}{c}{{{ P@3}}} & \multicolumn{1}{c}{{{ R@3}}} & \multicolumn{1}{c}{{{ F1@3}}} & \multicolumn{1}{c}{H\scriptsize{(F1@3)}} \\
\midrule

\rowcolor[HTML]{EDEDED}
CLIP~\citep{radford2021learning} 
& {\color[HTML]{757171} \textbf{51.96}} 
& {\color[HTML]{757171} \textbf{45.33}}  
& {\color[HTML]{757171} \textbf{46.49}}  
& {\color[HTML]{757171} \textbf{45.90}} 

& {\color[HTML]{757171} \textbf{40.21}}
& {\color[HTML]{757171} \textbf{40.43}}  
& {\color[HTML]{757171} \textbf{41.45}}  
& {\color[HTML]{757171} \textbf{40.93}}  

& {\color[HTML]{757171} \textbf{43.27}}  
& {\color[HTML]{757171} \textbf{24.94}}
& {\color[HTML]{757171} \textbf{27.85}} 
& {\color[HTML]{757171} \textbf{28.54}}  

& {\color[HTML]{757171} \textbf{28.19}}  
& {\color[HTML]{757171} \textbf{34.93}}  
& {\color[HTML]{757171} \textbf{11.30}}  
& {\color[HTML]{757171} \textbf{15.55}}  

& {\color[HTML]{757171} \textbf{15.94}}  
& {\color[HTML]{757171} \textbf{15.74}}  
& {\color[HTML]{757171} \textbf{23.44}}  \\

TeCoA~\citep{mao2022understanding}& 46.34 & 33.73& 34.57& 34.14 & 38.90 & {\ul 39.14}& {\ul 40.14}& {\ul 39.63} & 36.68 & {\ul 42.10} & 30.54& 31.32& 30.92 & 32.45 & \textbf{31.29} & 25.55& 26.21& 25.87 & 29.44 \\
FARE~\citep{schlarmann2024robust} & \textbf{49.21} & \textbf{43.83}& \textbf{44.95}& \textbf{44.37} & 43.33 & 38.63& 39.61& 39.11 & \textbf{41.57} & 23.82 & 32.89& 33.74& 33.30 & {\ul 38.05} & 11.21 & 24.31& 24.93& 24.61 & 31.66 \\
PMG-AFT~\citep{wang2024pre} & {\ul 48.94} & {\ul 41.77}& {\ul 42.83}& {\ul 42.29} & \textbf{46.42} & 36.00& 36.91& 36.44 & 39.15 & 34.56 & 32.91& 33.75& 33.32 & 37.27 & 15.51 & 25.16& 25.81& 25.47 & 31.79 \\
TGA-ZSR~\citep{yu2024text} & 48.61 & 37.19& 38.13& 37.65 & {\ul 44.62} & 32.33& 33.15& 32.73 & 35.02 & \textbf{42.79} & {\ul 33.52}& {\ul 34.37}& {\ul 33.93} & 35.69 & {\ul 29.20}  & {\ul 27.31}& {\ul 28.01}& {\ul 27.65} & {\ul 31.88} \\

{Comp-TGA~\citep{yu2026complementary}} & {48.14} & {38.11} & {39.07} & {38.58} & {44.52} & {36.10} & {37.01} & {36.54} & {37.53} & {40.07} & {33.32} & {34.17} & {33.73} & {35.99} & {25.16} & {26.97} & {27.66} & {27.30} & {\ul 31.97} \\

\rowcolor[HTML]{faebd7} UCAT (Ours) & 47.14 & 41.55& 42.62& 42.07 & 44.31 & \textbf{40.10}& \textbf{41.12}& \textbf{40.60} & {\ul 41.32} & 39.88 & \textbf{37.47}& \textbf{38.43}& \textbf{37.94} & \textbf{39.90} & 28.01 & \textbf{29.95}& \textbf{30.72}& \textbf{30.32} & \textbf{35.24} \\
\bottomrule
\end{tabular}} 
\label{tab:multilabelPGD}
\vspace{-1em}
\end{table*}

%% file: Tab/tinyImageNet-esp2.tex
\begin{table}[!ht]
\centering
\caption{
% \textbf{Zero-shot adversarial robustness under 10-step PGD training.}
% All methods are fine-tuned on TinyImageNet following FARE~\citep{schlarmann2024robust}, adversarial training uses 10-step PGD~\citep{madry2017towards} with $\epsilon\!=\!2/255$. \textit{Average} is the mean across datasets; \textit{H} is the harmonic mean between Clean and the corresponding robust score. Best and second-best are in \textbf{bold} and \underline{underline}.
\textbf{Zero-shot adversarial robustness under 10-step PGD training.} All methods are fine-tuned on TinyImageNet following FARE \citep{schlarmann2024robust}, and adversarial finetuning uses 10-step PGD \citep{madry2017towards} with $\epsilon=2/255$. We report zero-shot robustness across 16 single-label datasets under five different white-box and adaptive attacks (PGD~\citep{madry2017towards},  CW~\citep{carlini2017towards}, AutoAttack~\citep{croce2020reliable}, CAA~\citep{mao2021composite} and A$^3$~\citep{liu2022practical}). \textit{H} is the harmonic mean between Clean and the corresponding robust score. Best and second-best are in \textbf{bold} and \underline{underline}.
}
\resizebox{\linewidth}{!}{
\begin{tabular}{ll|rrrrrrrrrrrrrrrr|rr}
\toprule
\multicolumn{2}{l}{Methods} &
\multicolumn{1}{|c}{\rotatebox{90}{TinyImageNet}} &
\multicolumn{1}{c}{\rotatebox{90}{CIFAR-10}} &
\multicolumn{1}{c}{\rotatebox{90}{CIFAR-100}} &
\multicolumn{1}{c}{\rotatebox{90}{STL10}} &
\multicolumn{1}{c}{\rotatebox{90}{SUN397}} &
\multicolumn{1}{c}{\rotatebox{90}{Food101}} &
\multicolumn{1}{c}{\rotatebox{90}{Oxfordpets}} &
\multicolumn{1}{c}{\rotatebox{90}{Flowers102}} &
\multicolumn{1}{c}{\rotatebox{90}{DTD}} &
\multicolumn{1}{c}{\rotatebox{90}{EuroSAT}} &
\multicolumn{1}{c}{\rotatebox{90}{FGVC Aircraft}} &
\multicolumn{1}{c}{\rotatebox{90}{ImageNet}} &
\multicolumn{1}{c}{\rotatebox{90}{Caltech101}} &
\multicolumn{1}{c}{\rotatebox{90}{Caltech256}} &
\multicolumn{1}{c}{\rotatebox{90}{StanfordCars}} &
\multicolumn{1}{c}{\rotatebox{90}{PCAM}} &
\multicolumn{1}{|c}{\rotatebox{90}{Average}} &
\multicolumn{1}{c}{\rotatebox{90}{H}} \\

\midrule
\rowcolor[HTML]{EDEDED}
{\cellcolor[HTML]{FFFFFF}\multirow{6}{*}{\rotatebox{90}{Clean}}} & \color[HTML]{757171} \textbf{CLIP}~\citep{radford2021learning}          & {\color[HTML]{757171} \textbf{57.96}} & {\color[HTML]{757171} \textbf{88.03}} & {\color[HTML]{757171} \textbf{60.45}} & {\color[HTML]{757171} \textbf{97.03}} & {\color[HTML]{757171} \textbf{57.26}} & {\color[HTML]{757171} \textbf{83.89}} & {\color[HTML]{757171} \textbf{87.41}} & {\color[HTML]{757171} \textbf{65.49}} & {\color[HTML]{757171} \textbf{40.64}} & {\color[HTML]{757171} \textbf{42.66}} & {\color[HTML]{757171} \textbf{20.16}} & {\color[HTML]{757171} \textbf{59.15}} & {\color[HTML]{757171} \textbf{85.32}} & {\color[HTML]{757171} \textbf{81.73}} & {\color[HTML]{757171} \textbf{52.02}} & {\color[HTML]{757171} \textbf{52.08}} & {\color[HTML]{757171} \textbf{64.45}} & {\color[HTML]{757171} } \\
& TeCoA~\citep{mao2022understanding}      & 63.20& 58.62& 31.75& 80.59& 25.71& 19.15& 49.25& 24.61& 17.34& 15.89& 2.88& 24.70& 63.04& 47.67& 13.11& 49.97& 36.72&    \\
& FARE~\citep{schlarmann2024robust}      & {16.92}& {40.23}& {11.96}& {64.56}& {7.89} & {8.07} & {19.24}& {11.82}& {7.93} & {12.52}& {2.55} & {7.98} & {47.14}& {27.61}& {6.06} & {\ul 50.02}& 21.41&    \\
% & PMG-AFT~\citep{wang2024pre}    & 22.16& {\ul 74.05}& 33.81& \textbf{92.75}& \textbf{55.66}& \textbf{72.69}& \textbf{83.10}& \textbf{55.86}& \textbf{28.30}& {\ul 19.83}& \textbf{17.46}& \textbf{51.46}& \textbf{80.83}& \textbf{75.22}& \textbf{43.09}& 48.72& \textbf{53.44}& {\ul }    \\
& PMG-AFT~\citep{wang2024pre}    & 46.62 & 70.37 & 38.49 & 90.34 & \textbf{52.02} & \textbf{57.11} & \textbf{79.75} & \textbf{49.36} & \textbf{32.23} & 23.43 & \textbf{12.03} & \textbf{47.70} & \textbf{82.49} & \textbf{73.65} & \textbf{41.60} & \textbf{55.87} & \textbf{53.32} &    \\

& TGA-ZSR~\citep{yu2024text} & \textbf{69.78}& \textbf{83.98}& \textbf{52.32}& \textbf{{91.36}}& {\ul 44.70}& {\ul 50.17}& {\ul 72.55}& {\ul 45.05}& {\ul 26.92}& \textbf{27.58}& {\ul 10.68}& {\ul 41.84}& {\ul 80.04}& {\ul 71.94}& {\ul 33.14}& {\ul 50.02}& {\ul 53.25}&    \\

 & {Comp-TGA~\citep{yu2026complementary}} & {66.58} & {\ul 82.25} & {\ul 49.40} & {\ul 91.35} & {41.73} & {46.89} & {70.13} & {44.14} & {26.01} & {\ul 26.96} & {8.16} & {38.31} & {78.85} & {70.33} & {32.63} & {\ul 50.02} & {51.48} & \\

\rowcolor[HTML]{faebd7}
{\cellcolor[HTML]{FFFFFF}\multirow{-6}{*}{}} & UCAT (Ours)          & {\ul 67.18}& 66.52& {41.07}& 86.73& 30.10& 36.97& 62.66& 36.69& 24.95& 19.39& 7.26& 32.61& 75.00& 60.15& 26.39& 49.66& 45.21& \textbf{} \\

\midrule
\rowcolor[HTML]{EDEDED}
{\cellcolor[HTML]{FFFFFF}\multirow{6}{*}{\rotatebox{90}{PGD}}} 
& \color[HTML]{757171} \textbf{CLIP}~\citep{radford2021learning}          & {\color[HTML]{757171} \textbf{0.00}}  & {\color[HTML]{757171} \textbf{0.94}}  & {\color[HTML]{757171} \textbf{0.28}}  & {\color[HTML]{757171} \textbf{0.45}}  & {\color[HTML]{757171} \textbf{0.00}}  & {\color[HTML]{757171} \textbf{0.00}}  & {\color[HTML]{757171} \textbf{0.00}}  & {\color[HTML]{757171} \textbf{0.00}}  & {\color[HTML]{757171} \textbf{0.11}}  & {\color[HTML]{757171} \textbf{0.00}}  & {\color[HTML]{757171} \textbf{0.00}}  & {\color[HTML]{757171} \textbf{0.00}}  & {\color[HTML]{757171} \textbf{0.77}}  & {\color[HTML]{757171} \textbf{0.19}}  & {\color[HTML]{757171} \textbf{0.00}}  & {\color[HTML]{757171} \textbf{0.00}}  & {\color[HTML]{757171} \textbf{0.00}}  & {\color[HTML]{757171} \textbf{0.00}} \\
& TeCoA~\citep{mao2022understanding}      & \textbf{35.74}& {\ul 23.57} & {\ul 14.47}& {\ul 53.50}& {\ul 10.20}& 6.58& {\ul 22.90}& 10.96& 10.75& \textbf{11.72}& 0.57& {\ul 10.53}& 40.27& 27.02& {\ul 3.61} & 49.31& {\ul 20.73}& 26.50     \\
& FARE~\citep{schlarmann2024robust}      & {8.62} & {18.19}& {5.67} & {41.10}& {3.46} & {2.88} & {7.01} & {5.22} & {5.21} & {9.29} & {{\ul 0.96}}        & {3.47} & {31.98}& {15.47}& {1.73} & {\textbf{50.02}}    & 13.14& 16.29     \\
% & PMG-AFT~\citep{wang2024pre}    & 0.12& {\ul 24.10}& 3.76& 16.34& 0.16& 0.04& 0.30& 0.23& 2.29& 0.50& 0.00& 0.23& 4.53& 2.02& 0.01& 47.90& 6.41& 11.44     \\
& PMG-AFT~\citep{wang2024pre}    & 7.08 & 13.18 & 3.63 & 55.79 & 3.39 & {\ul 7.90} & 12.81 & 4.73 & 5.11 & 4.47 & 0.00 & 4.38 & 39.65 & 26.22 & 0.44 & 0.94 & 11.86 & 19.40    \\
& TGA-ZSR~\citep{yu2024text} & 30.74& 20.17& 12.02& 51.99& 9.46& {6.69} & 20.58& {\ul 12.47}& {10.85}& {11.22}& 0.63& 10.28& {\ul 40.63}& {\ul 29.06}& 3.56& {\ul 49.97}& 20.02& {\ul 29.10}  \\

 & {Comp-TGA~\citep{yu2026complementary}} & {30.28} & {18.79} & {11.53} & {48.83} & {8.71} & {5.45} & {19.60} & {11.87} & {\ul 11.12} & {\ul 11.42} & {0.45} & {9.02} & {36.67} & {24.82} & {2.90} & {49.93} & {18.84} & {27.58} \\
 
\rowcolor[HTML]{faebd7}
{\cellcolor[HTML]{FFFFFF}\multirow{-6}{*}{}} & UCAT (Ours) & {\ul 35.38}& \textbf{25.81}& \textbf{15.67}& \textbf{58.44}& \textbf{11.48}& \textbf{11.17}& \textbf{26.82}& \textbf{15.04}& \textbf{13.94}& 4.53& \textbf{1.20} & \textbf{13.13}& \textbf{51.34}& \textbf{34.60}& \textbf{6.72} & 34.02& \textbf{22.45}& \textbf{30.01}         \\

\midrule
\rowcolor[HTML]{EDEDED}
{\cellcolor[HTML]{FFFFFF}\multirow{6}{*}{\rotatebox{90}{CW}}} 
& \color[HTML]{757171} \textbf{CLIP}~\citep{radford2021learning}          & {\color[HTML]{757171} \textbf{0.00}}  & {\color[HTML]{757171} \textbf{0.58}}  & {\color[HTML]{757171} \textbf{0.20}}  & {\color[HTML]{757171} \textbf{0.57}}  & {\color[HTML]{757171} \textbf{0.08}}  & {\color[HTML]{757171} \textbf{0.00}}  & {\color[HTML]{757171} \textbf{0.00}}  & {\color[HTML]{757171} \textbf{0.00}}  & {\color[HTML]{757171} \textbf{0.12}}  & {\color[HTML]{757171} \textbf{0.00}}  & {\color[HTML]{757171} \textbf{0.00}}  & {\color[HTML]{757171} \textbf{0.08}}  & {\color[HTML]{757171} \textbf{0.24}}  & {\color[HTML]{757171} \textbf{0.33}}  & {\color[HTML]{757171} \textbf{2.19}}  & {\color[HTML]{757171} \textbf{0.00}}  & {\color[HTML]{757171} \textbf{0.27}}  & {\color[HTML]{757171} \textbf{0.55}} \\
& TeCoA~\citep{mao2022understanding}      & {\ul 33.90}& {\ul 23.05}& {\ul 13.66}& {\ul 52.50}& 8.95& 5.74& {\ul 22.13}& 10.05& 9.42& \textbf{11.40}& {\ul 0.63} & 9.58& 39.92& 26.04& 3.10& 49.26& {\ul 19.96}& 25.86     \\
& FARE~\citep{schlarmann2024robust}      & {7.04} & {14.64}& {4.63} & {38.94}& {2.91} & {2.20} & {6.68} & {4.38} & {4.36} & {8.43} & {0.75} & {2.96} & {30.13}& {14.24}& {1.73} & {\textbf{50.02}}    & 12.13& 15.48     \\
% & PMG-AFT~\citep{wang2024pre}    & 0.02& 1.63& 0.70& 2.73& 0.09& 0.02& 0.03& 0.00& 0.32& 0.00& 0.00& 0.11& 3.03& 1.13& 1.87& 0.00& 0.73& 1.44\\
& PMG-AFT~\citep{wang2024pre}    & 1.64 & 0.27 & 0.79 & 14.04 & 1.89 & 1.15 & 2.51 & 3.09 & 4.42 & 5.83 & 0.00 & 1.88 & 19.58 & 11.41 & 2.02 & 1.03 & 4.47 & 8.25   \\
& TGA-ZSR~\citep{yu2024text} & 29.68& 20.48& 11.48& 51.53& {\ul 9.15} & {\ul 6.43} & 21.72& {\ul 12.05}& {\ul 9.63} & {11.01}& 0.60& {\ul 10.06}& {\ul 40.79}& {\ul 28.67}& {\ul 4.48} & {\ul 49.97}& 19.86& {\ul 28.93}  \\
 & {Comp-TGA~\citep{yu2026complementary}} & {27.92} & {18.56} & {10.77} & {47.96} & {7.94} & {4.83} & {19.24} & {11.14} & {9.31} & {\ul 11.23} & {0.42} & {8.47} & {36.14} & {23.90} & {3.62} & {49.93} & {18.21} & {26.90} \\
\rowcolor[HTML]{faebd7}
{\cellcolor[HTML]{FFFFFF}\multirow{-6}{*}{}} & UCAT (Ours) & \textbf{34.64}& \textbf{25.46}& \textbf{14.69}& \textbf{57.88}& \textbf{10.25}& \textbf{9.83} & \textbf{26.49}& \textbf{12.91}& \textbf{11.70}& 3.80& \textbf{1.11} & \textbf{11.95}& \textbf{50.47}& \textbf{33.40}& \textbf{6.36} & 33.09& \textbf{21.50}& \textbf{29.14}          \\

\midrule

\rowcolor[HTML]{EDEDED}
{\cellcolor[HTML]{FFFFFF}\multirow{6}{*}{\rotatebox{90}{AutoAttack}}} 
& \color[HTML]{757171} \textbf{CLIP}~\citep{radford2021learning}          & {\color[HTML]{757171} \textbf{0.00}}  & {\color[HTML]{757171} \textbf{0.05}}  & {\color[HTML]{757171} \textbf{0.14}}  & {\color[HTML]{757171} \textbf{0.00}}  & {\color[HTML]{757171} \textbf{0.02}}  & {\color[HTML]{757171} \textbf{0.03}}  & {\color[HTML]{757171} \textbf{0.03}}  & {\color[HTML]{757171} \textbf{0.02}}  & {\color[HTML]{757171} \textbf{0.13}}  & {\color[HTML]{757171} \textbf{0.17}}  & {\color[HTML]{757171} \textbf{0.23}}  & {\color[HTML]{757171} \textbf{0.03}}  & {\color[HTML]{757171} \textbf{0.02}}  & {\color[HTML]{757171} \textbf{0.06}}  & {\color[HTML]{757171} \textbf{0.07}}  & {\color[HTML]{757171} \textbf{0.12}}  & {\color[HTML]{757171} \textbf{0.07}}  & {\color[HTML]{757171} \textbf{0.14}} \\
& TeCoA~\citep{mao2022understanding}      & {\ul 32.68}& {\ul 21.94}& {\ul 13.17}& {\ul 51.93}& {\ul 8.43} & {\ul 5.54} & {\ul 21.56}& {\ul 9.92} & {\ul 9.52} & \textbf{11.36}& 0.51& {\ul 9.10} & {\ul 38.88}& {\ul 25.35}& {\ul 2.56} & {\ul 49.23}& {\ul 19.48}& {\ul 25.46}  \\
& FARE~\citep{schlarmann2024robust}      & {7.00} & {14.81}& {4.58} & {38.71}& {2.81} & {2.16} & {6.49} & {4.29} & {4.47} & {8.52} & {{\ul 0.72}}        & {2.86} & {29.84}& {14.03}& {1.34} & {\textbf{50.02}}    & 12.04& 15.41     \\
% & PMG-AFT~\citep{wang2024pre}    & 0.00& 0.96& 0.35& 0.74& 0.06& 0.05& 0.06& 0.05& 0.27& 0.03& 0.03& 0.04& 0.92& 0.26& 0.04& 0.21& 0.25& 0.51\\
 & PMG-AFT~\citep{wang2024pre}    & 0.62 & 0.14 & 0.25 & 7.58 & 0.49 & 0.40 & 0.55 & 0.94 & 2.29 & 0.04 & 0.15 & 0.56 & 12.33 & 6.19 & 0.04 & 0.37 & 2.06 & 3.96 \\
& TGA-ZSR~\citep{yu2024text} & 11.28& 6.29& 5.53& 36.76& 4.00& 3.27& 9.27& 6.18& 6.33& {8.94} & 0.18& 5.13& 30.23& 19.57& 0.96& 42.88& 12.30& 19.98     \\
 & {Comp-TGA~\citep{yu2026complementary}} & {11.84} & {6.64} & {4.94} & {34.56} & {3.95} & {2.80} & {9.02} & {5.82} & {6.60} & {\ul 9.68} & {0.12} & {4.69} & {27.71} & {16.99} & {0.87} & {43.61} & {11.87} & {19.29} \\
\rowcolor[HTML]{faebd7}
{\cellcolor[HTML]{FFFFFF}\multirow{-6}{*}{}} & UCAT (Ours) & \textbf{32.84}& \textbf{24.08}& \textbf{13.97}& \textbf{57.15}& \textbf{9.50} & \textbf{9.09} & \textbf{24.72}& \textbf{12.60}& \textbf{11.97}& 3.76& \textbf{0.78} & \textbf{11.11}& \textbf{49.44}& \textbf{32.35}& \textbf{4.71} & 32.62& \textbf{20.67}& \textbf{28.37}          \\
\midrule

\rowcolor[HTML]{EDEDED}
{\cellcolor[HTML]{FFFFFF}\multirow{-6}{*}{}}   & \color[HTML]{757171}\textbf{CLIP}~\citep{radford2021learning}   & \color[HTML]{757171}\textbf{1.90} & \color[HTML]{757171}\textbf{5.84} & \color[HTML]{757171}\textbf{0.32} & \color[HTML]{757171}\textbf{28.99} & \color[HTML]{757171}\textbf{0.63} & \color[HTML]{757171}\textbf{9.80} & \color[HTML]{757171}\textbf{5.04} & \color[HTML]{757171}\textbf{1.29} & \color[HTML]{757171}\textbf{0.05} & \color[HTML]{757171}\textbf{0.04} & \color[HTML]{757171}\textbf{0.00} & \color[HTML]{757171}\textbf{1.02} & \color[HTML]{757171}\textbf{16.63} & \color[HTML]{757171}\textbf{13.17} & \color[HTML]{757171}\textbf{0.42} & \color[HTML]{757171}\textbf{0.00} & \color[HTML]{757171}\textbf{5.32} & \color[HTML]{757171}\textbf{9.83} \\
& TeCoA~\citep{mao2022understanding}  & \textbf{35.52} & {\ul 23.40} & {\ul 14.38}& 53.25   & {\ul 9.98} & 6.49& {\ul 22.65} & {\ul 10.95}  & {\ul 10.69} & \textbf{11.62}  & 0.57& {\ul 10.33}& {\ul 40.03}  & {\ul 26.83}  & {\ul 3.43} & {\ul 49.28}  & {\ul 20.59} & {\ul 26.38}\\
& FARE~\citep{schlarmann2024robust}  & 8.50& 17.74 & 5.51& 40.86   & 3.39 & 2.80& 6.95 & 5.14  & 5.05  & 8.86& {\ul 0.93}& 3.41& 31.72 & 15.32 & 1.65& \textbf{50.02}& 12.99 & 16.17 \\
& PMG-AFT~\citep{wang2024pre}& 7.18& 13.18 & 3.79& {\ul 56.18}   & 3.37 & {\ul 8.21}& 13.63& 4.36  & 3.19  & {0.32}   & 0.00& 4.43& 39.61 & 26.36 & 0.44& {0.39}& 11.54 & 18.97 \\
& TGA-ZSR~\citep{yu2024text}& {18.64} & 10.77 & {8.41} & 42.23   & 5.78 & 4.51& 11.72& 7.81  & 7.87  & {\ul 10.54} & 0.27& 6.73& 33.71 & 22.49 & 1.62& 42.37  & 14.72 & 23.06 \\

 & {Comp-TGA~\citep{yu2026complementary}} & {10.60} & {7.58} & {5.59} & {38.86} & {4.86} & {3.37} & {9.02} & {6.34} & {7.39} & {8.88} & {0.09} & {5.65} & {29.60} & {18.75} & {1.43} & {41.75} & {12.48} & {20.10} \\
 
\rowcolor[HTML]{faebd7}
{\cellcolor[HTML]{FFFFFF}\multirow{-6}{*}{\rotatebox{90}{CAA}}} & \textbf{UCAT} & {\ul 35.12}& \textbf{25.69}  & \textbf{15.42}   & \textbf{58.33}& \textbf{11.16} & \textbf{10.91}  & \textbf{26.90}& {\textbf{14.73}}& \textbf{13.83} & 4.33& \textbf{1.08}& \textbf{12.75}   & \textbf{51.11} & \textbf{34.33} & \textbf{6.29}  & 33.45  & \textbf{22.21} & \textbf{29.79}   \\ \midrule

\rowcolor[HTML]{EDEDED}
{\cellcolor[HTML]{FFFFFF}\multirow{-6}{*}{}}   & \color[HTML]{757171}\textbf{CLIP}~\citep{radford2021learning}   & \color[HTML]{757171}\textbf{1.26} & \color[HTML]{757171}\textbf{6.47} & \color[HTML]{757171}\textbf{0.33} & \color[HTML]{757171}\textbf{30.70} & \color[HTML]{757171}\textbf{0.70} & \color[HTML]{757171}\textbf{9.73} & \color[HTML]{757171}\textbf{4.80} & \color[HTML]{757171}\textbf{1.11} & \color[HTML]{757171}\textbf{0.11} & \color[HTML]{757171}\textbf{0.10} & \color[HTML]{757171}\textbf{0.00} & \color[HTML]{757171}\textbf{1.00} & \color[HTML]{757171}\textbf{19.11} & \color[HTML]{757171}\textbf{13.56} & \color[HTML]{757171}\textbf{0.37} & \color[HTML]{757171}\textbf{0.00} & \color[HTML]{757171}\textbf{5.58} & \color[HTML]{757171}\textbf{10.28} \\
& TeCoA~\citep{mao2022understanding}  & \textbf{35.38} & {\ul 23.21} & {\ul 14.22}& 53.11   & {\ul 9.95} & 6.46& {\ul 22.57} & 10.83 & 10.59 & \textbf{11.62}  & {0.57}& {\ul 10.29} & 39.98 & 26.75 & {\ul 3.47} & 49.27  & {\ul 20.52} & 26.32 \\
\multirow{-7}{*}{} & FARE~\citep{schlarmann2024robust}  & 8.46& 17.55 & 5.49& 40.65   & 3.39 & {2.78}& 6.87 & 5.12  & 5.05  & 8.89& {\ul 0.93} & 3.40& 31.69 & 15.31 & 1.65& \textbf{50.02}& 12.95 & 16.14 \\
& PMG-AFT~\citep{wang2024pre}& 7.02& 13.16 & 3.65& {\ul 55.74}   & 3.37 & {\ul 7.90}& 12.87& 4.68  & 4.95  & 4.45& 0.00& 4.37& 39.60 & 26.18 & 0.44& 0.89   & 11.83 & 19.36 \\
& TGA-ZSR~\citep{yu2024text}& 30.02   & 19.54 & 11.59& 51.60   & 9.02 & 6.44& 19.98& {\ul 12.21}  & {\ul 10.64} & {\ul 11.13} & 0.54& 9.85& {\ul 40.07}  & {\ul 28.50}  & 3.20& {\ul 49.95}  & 19.64 & {\ul 28.70}\\

 & {Comp-TGA~\citep{yu2026complementary}} & {29.20} & {18.24} & {11.13} & {48.39} & {8.28} & {5.29} & {18.89} & {11.58} & {10.37} & {11.26} & {0.42} & {8.70} & {36.28} & {24.42} & {2.72} & {49.93} & {18.44} & {27.16} \\
\rowcolor[HTML]{faebd7}
{\cellcolor[HTML]{FFFFFF}\multirow{-6}{*}{\rotatebox{90}{A$^3$}}} & UCAT   & {\ul 34.90}& \textbf{25.44}  & \textbf{15.33}   & \textbf{58.14}& \textbf{11.08} & \textbf{10.82}  & \textbf{26.41}& \textbf{14.67} & \textbf{13.83} & 4.33& \textbf{1.08}& \textbf{12.69}   & \textbf{51.02} & \textbf{34.22} & \textbf{6.19}  & 33.34  & \textbf{22.09} & \textbf{29.68}   \\

\bottomrule
\end{tabular}}
\label{tab:tinyimagenet-eps2}
\vspace{-1em}
\end{table}

%% file: Tab/ImageNet-eps1.tex
\begin{table}[!ht]
\centering
\caption{\textbf{Zero-shot adversarial robustness across 15 datasets.}
All methods are fine-tuned on ImageNet-1k following TeCoA~\citep{mao2022understanding}, adversarial training uses 2-step PGD~\citep{madry2017towards} with $\epsilon\!=\!1/255$. \textit{Average} is the mean across datasets; \textit{H} is the harmonic mean between Clean and the corresponding robust score. Best and second-best are in \textbf{bold} and \underline{underline}.}
\resizebox{\linewidth}{!}{
\begin{tabular}{cl|rrrrrrrrrrrrrrr|rr}
\toprule

\multicolumn{2}{l}{Methods} &
\multicolumn{1}{c}{\rotatebox{90}{CIFAR-10}} &
\multicolumn{1}{c}{\rotatebox{90}{CIFAR-100}} &
\multicolumn{1}{c}{\rotatebox{90}{STL10}} &
\multicolumn{1}{c}{\rotatebox{90}{SUN397}} &
\multicolumn{1}{c}{\rotatebox{90}{Food101}} &
\multicolumn{1}{c}{\rotatebox{90}{Oxfordpets}} &
\multicolumn{1}{c}{\rotatebox{90}{Flowers102}} &
\multicolumn{1}{c}{\rotatebox{90}{DTD}} &
\multicolumn{1}{c}{\rotatebox{90}{EuroSAT}} &
\multicolumn{1}{c}{\rotatebox{90}{FGVC Aircraft}} &
\multicolumn{1}{c}{\rotatebox{90}{ImageNet}} &
\multicolumn{1}{c}{\rotatebox{90}{Caltech101}} &
\multicolumn{1}{c}{\rotatebox{90}{Caltech256}} &
\multicolumn{1}{c}{\rotatebox{90}{StanfordCars}} &
\multicolumn{1}{c}{\rotatebox{90}{PCAM}} &
\multicolumn{1}{|c}{\rotatebox{90}{Average}} &
\multicolumn{1}{c}{\rotatebox{90}{H}} \\
\midrule

& \cellcolor[HTML]{EDEDED}CLIP~\citep{radford2021learning} & \cellcolor[HTML]{EDEDED}{\color[HTML]{757171}\textbf{88.03}} & \cellcolor[HTML]{EDEDED}{\color[HTML]{757171}\textbf{60.45}} & \cellcolor[HTML]{EDEDED}{\color[HTML]{757171}\textbf{97.03}} & \cellcolor[HTML]{EDEDED}{\color[HTML]{757171}\textbf{57.26}} & \cellcolor[HTML]{EDEDED}{\color[HTML]{757171}\textbf{83.89}} & \cellcolor[HTML]{EDEDED}{\color[HTML]{757171}\textbf{87.41}} & \cellcolor[HTML]{EDEDED}{\color[HTML]{757171}\textbf{65.49}} & \cellcolor[HTML]{EDEDED}{\color[HTML]{757171}\textbf{40.64}} & \cellcolor[HTML]{EDEDED}{\color[HTML]{757171}\textbf{42.66}} & \cellcolor[HTML]{EDEDED}{\color[HTML]{757171}\textbf{20.16}} & \cellcolor[HTML]{EDEDED}{\color[HTML]{757171}\textbf{59.15}} & \cellcolor[HTML]{EDEDED}{\color[HTML]{757171}\textbf{85.32}} & \cellcolor[HTML]{EDEDED}{\color[HTML]{757171}\textbf{81.73}} & \cellcolor[HTML]{EDEDED}{\color[HTML]{757171}\textbf{52.02}} & \cellcolor[HTML]{EDEDED}{\color[HTML]{757171}\textbf{52.08}} & \cellcolor[HTML]{EDEDED}{\color[HTML]{757171}\textbf{64.89}} & \cellcolor[HTML]{EDEDED}{\color[HTML]{757171}\textbf{}}     \\
& TeCoA~\citep{mao2022understanding}         & 78.12& 49.68& 93.30& 51.28& 55.37& 81.58& 50.92& 34.15& \textbf{27.57}& 13.89& 63.87& 83.51& 76.51& 33.30& \textbf{49.01}& 56.14&    \\
& FARE~\citep{schlarmann2024robust}        & \textbf{84.75}& \textbf{59.85}& \textbf{95.69}& {\ul 53.97}   & \textbf{75.58}& \textbf{86.92}& \textbf{60.48}& {\ul 36.86}   & {\ul 24.74}   & \textbf{17.10}& \textbf{85.01}& \textbf{85.01}& \textbf{80.57}& \textbf{49.71}& 45.06& \textbf{62.75}& \textbf{}    \\
\rowcolor[HTML]{faebd7}
{\cellcolor[HTML]{FFFFFF}\multirow{-4}{*}{\rotatebox{90}{\scriptsize{Clean}}}}        & UCAT (Ours)      & {\ul 83.78}   & {\ul 58.11}   & {\ul 95.65}   & \textbf{53.98}& {\ul 68.84}   & {\ul 86.05}   & {\ul 58.30}   & \textbf{37.18}& 23.02& {\ul 15.24}   & {\ul 70.48}   & {\ul 84.64}   & {\ul 80.27}   & {\ul 44.96}   & {\ul 46.56}   & {\ul 60.47}   & {\ul }       \\
\midrule

& \cellcolor[HTML]{EDEDED}CLIP~\citep{radford2021learning} & \cellcolor[HTML]{EDEDED}{\color[HTML]{757171}\textbf{9.57}}  & \cellcolor[HTML]{EDEDED}{\color[HTML]{757171}\textbf{4.55}}  & \cellcolor[HTML]{EDEDED}{\color[HTML]{757171}\textbf{35.40}} & \cellcolor[HTML]{EDEDED}{\color[HTML]{757171}\textbf{1.02}}  & \cellcolor[HTML]{EDEDED}{\color[HTML]{757171}\textbf{3.95}}  & \cellcolor[HTML]{EDEDED}{\color[HTML]{757171}\textbf{2.72}}  & \cellcolor[HTML]{EDEDED}{\color[HTML]{757171}\textbf{1.19}}  & \cellcolor[HTML]{EDEDED}{\color[HTML]{757171}\textbf{2.50}}  & \cellcolor[HTML]{EDEDED}{\color[HTML]{757171}\textbf{0.04}}  & \cellcolor[HTML]{EDEDED}{\color[HTML]{757171}\textbf{0.00}}  & \cellcolor[HTML]{EDEDED}{\color[HTML]{757171}\textbf{1.72}}  & \cellcolor[HTML]{EDEDED}{\color[HTML]{757171}\textbf{24.63}} & \cellcolor[HTML]{EDEDED}{\color[HTML]{757171}\textbf{7.19}}  & \cellcolor[HTML]{EDEDED}{\color[HTML]{757171}\textbf{0.27}}  & \cellcolor[HTML]{EDEDED}{\color[HTML]{757171}\textbf{0.10}}  & \cellcolor[HTML]{EDEDED}{\color[HTML]{757171}\textbf{0.05}}  & \cellcolor[HTML]{EDEDED}{\color[HTML]{757171}\textbf{0.10}} \\
& TeCoA~\citep{mao2022understanding}         & \textbf{59.28}& 34.13& \textbf{83.45}& {\ul 29.81}& 27.99& {\ul 62.61}& 30.69& 22.88& \textbf{15.18}& 5.10& {\ul 41.88}& {\ul 69.07}& 59.54& 13.37& 23.87& {\ul 38.59}   & 45.74        \\
& FARE~\citep{schlarmann2024robust}        & 50.96& \textbf{28.48}& 80.88& 26.66& \textbf{34.36}& 61.43& {\ul 31.91}& {\ul 24.31}& {\ul 14.12}& {\ul 5.28}& 32.11& 68.19& {\ul 59.95}& {\ul 18.52}& {\ul 25.74}& 37.53& {\ul 46.97}        \\
\rowcolor[HTML]{faebd7}
{\cellcolor[HTML]{FFFFFF}\multirow{-4}{*}{\rotatebox{90}{\scriptsize{Auto Attack}}}}   & UCAT (Ours)      & 50.59& \textbf{28.48}& {\ul 82.09}& \textbf{29.93}& {\ul 33.72}& \textbf{67.59}& \textbf{33.26}& \textbf{24.42}& 12.65& \textbf{5.73} & \textbf{47.51}& \textbf{71.11}& \textbf{62.71}& \textbf{19.62}& \textbf{25.84}& \textbf{39.68}& \textbf{47.92}                  \\
\midrule

& \cellcolor[HTML]{EDEDED}CLIP~\citep{radford2021learning} & \cellcolor[HTML]{EDEDED}{\color[HTML]{757171}\textbf{2.54}}  & \cellcolor[HTML]{EDEDED}{\color[HTML]{757171}\textbf{1.11}}  & \cellcolor[HTML]{EDEDED}{\color[HTML]{757171}\textbf{3.18}}  & \cellcolor[HTML]{EDEDED}{\color[HTML]{757171}\textbf{0.05}}  & \cellcolor[HTML]{EDEDED}{\color[HTML]{757171}\textbf{0.03}}  & \cellcolor[HTML]{EDEDED}{\color[HTML]{757171}\textbf{0.03}}  & \cellcolor[HTML]{EDEDED}{\color[HTML]{757171}\textbf{0.02}}  & \cellcolor[HTML]{EDEDED}{\color[HTML]{757171}\textbf{0.19}}  & \cellcolor[HTML]{EDEDED}{\color[HTML]{757171}\textbf{0.17}}  & \cellcolor[HTML]{EDEDED}{\color[HTML]{757171}\textbf{0.23}}  & \cellcolor[HTML]{EDEDED}{\color[HTML]{757171}\textbf{0.04}}  & \cellcolor[HTML]{EDEDED}{\color[HTML]{757171}\textbf{0.10}}  & \cellcolor[HTML]{EDEDED}{\color[HTML]{757171}\textbf{0.26}}  & \cellcolor[HTML]{EDEDED}{\color[HTML]{757171}\textbf{0.07}}  & \cellcolor[HTML]{EDEDED}{\color[HTML]{757171}\textbf{0.12}}  & \cellcolor[HTML]{EDEDED}{\color[HTML]{757171}\textbf{0.54}}  & \cellcolor[HTML]{EDEDED}{\color[HTML]{757171}\textbf{1.08}} \\
& TeCoA~\citep{mao2022understanding}         & \textbf{58.27}& \textbf{32.57}& \textbf{83.16}& \textbf{29.03}& 25.79& {\ul 61.76}   & 28.93& 20.70& \textbf{13.26}& 4.05& \textbf{48.51}& {\ul 68.40}   & 58.59& 12.03& 24.09& {\ul 37.94}   & 45.28        \\
& FARE~\citep{schlarmann2024robust}        & {\ul 49.62}   & 25.98& 80.60& 24.77& \textbf{33.06}& 60.51& {\ul 29.55}   & {\ul 22.02}   & {\ul 12.95}   & {\ul 4.08}    & 39.81& 67.21& {\ul 58.87}   & {\ul 16.43}   & \textbf{25.56}& 36.73& {\ul 46.34}        \\
\rowcolor[HTML]{faebd7}
{\cellcolor[HTML]{FFFFFF}\multirow{-4}{*}{\rotatebox{90}{\scriptsize{PGD}}}}          & UCAT (Ours)      & 49.00& {\ul 26.42}   & {\ul 81.73}   & {\ul 27.85}   & {\ul 31.88}   & \textbf{66.86}& \textbf{30.64}& \textbf{22.45}& 10.76& \textbf{4.50} & {\ul 45.59}   & \textbf{70.12}& \textbf{61.64}& \textbf{17.40}& {\ul 25.37}   & \textbf{38.15}& \textbf{46.78}                 \\
\bottomrule
\end{tabular}}
\label{tab:imagnet}
\vspace{-1em}

\end{table}

%% file: Tab/rebuttal-tauprime.tex
\begin{table}[!ht]
\centering
\caption{\textbf{Effect of the calibration coefficient $\tau^\prime$ on zero-shot adversarial robustness across 16 single-label datasets.}
All methods are fine-tuned on TinyImageNet following TeCoA~\citep{yu2024text}, adversarial training uses 2-step PGD~\citep{madry2017towards} with $\epsilon\!=\!1/255$. \textit{Average} is the mean across datasets. \textit{H} is the harmonic mean between Clean and the corresponding robust score. Best and second-best are in \textbf{bold} and \underline{underline}.}
\resizebox{\textwidth}{!}{
\begin{tabular}{ll|*{16}{r}|rr} 
% \begin{tabular}{>{\color{blue}}l>{\color{blue}}l|*{16}{>{\color{blue}}r}|>{\color{blue}}r>{\color{blue}}r} 
\toprule

\multicolumn{2}{l|}{Methods} &
\multicolumn{1}{c}{\rotatebox{90}{TinyImageNet}} &
\multicolumn{1}{c}{\rotatebox{90}{CIFAR-10}} &
\multicolumn{1}{c}{\rotatebox{90}{CIFAR-100}} &
\multicolumn{1}{c}{\rotatebox{90}{STL10}} &
\multicolumn{1}{c}{\rotatebox{90}{SUN397}} &
\multicolumn{1}{c}{\rotatebox{90}{Food101}} &
\multicolumn{1}{c}{\rotatebox{90}{Oxfordpets}} &
\multicolumn{1}{c}{\rotatebox{90}{Flowers102}} &
\multicolumn{1}{c}{\rotatebox{90}{DTD}} &
\multicolumn{1}{c}{\rotatebox{90}{EuroSAT}} &
\multicolumn{1}{c}{\rotatebox{90}{FGVC Aircraft}} &
\multicolumn{1}{c}{\rotatebox{90}{ImageNet}} &
\multicolumn{1}{c}{\rotatebox{90}{Caltech101}} &
\multicolumn{1}{c}{\rotatebox{90}{Caltech256}} &
\multicolumn{1}{c}{\rotatebox{90}{StanfordCars}} &
\multicolumn{1}{c}{\rotatebox{90}{PCAM}} &
\multicolumn{1}{|c}{\rotatebox{90}{Average}} &
\multicolumn{1}{c}{\rotatebox{90}{H}} \\

% \multicolumn{2}{>{\color{blue}}l|}{Methods} &
% \multicolumn{1}{>{\color{blue}}c}{\rotatebox{90}{TinyImageNet}} &
% \multicolumn{1}{>{\color{blue}}c}{\rotatebox{90}{CIFAR-10}} &
% \multicolumn{1}{>{\color{blue}}c}{\rotatebox{90}{CIFAR-100}} &
% \multicolumn{1}{>{\color{blue}}c}{\rotatebox{90}{STL10}} &
% \multicolumn{1}{>{\color{blue}}c}{\rotatebox{90}{SUN397}} &
% \multicolumn{1}{>{\color{blue}}c}{\rotatebox{90}{Food101}} &
% \multicolumn{1}{>{\color{blue}}c}{\rotatebox{90}{Oxfordpets}} &
% \multicolumn{1}{>{\color{blue}}c}{\rotatebox{90}{Flowers102}} &
% \multicolumn{1}{>{\color{blue}}c}{\rotatebox{90}{DTD}} &
% \multicolumn{1}{>{\color{blue}}c}{\rotatebox{90}{EuroSAT}} &
% \multicolumn{1}{>{\color{blue}}c}{\rotatebox{90}{FGVC Aircraft}} &
% \multicolumn{1}{>{\color{blue}}c}{\rotatebox{90}{ImageNet}} &
% \multicolumn{1}{>{\color{blue}}c}{\rotatebox{90}{Caltech101}} &
% \multicolumn{1}{>{\color{blue}}c}{\rotatebox{90}{Caltech256}} &
% \multicolumn{1}{>{\color{blue}}c}{\rotatebox{90}{StanfordCars}} &
% \multicolumn{1}{>{\color{blue}}c}{\rotatebox{90}{PCAM}} &
% \multicolumn{1}{|>{\color{blue}}c}{\rotatebox{90}{Average}} &
% \multicolumn{1}{>{\color{blue}}c}{\rotatebox{90}{H}} \\
\midrule

\rowcolor[HTML]{EDEDED}

{\cellcolor[HTML]{FFFFFF}\multirow{-6}{*}{}}   & \color[HTML]{757171}\textbf{CLIP}~\citep{radford2021learning} & \color[HTML]{757171}\textbf{57.96} & \color[HTML]{757171}\textbf{88.02} & \color[HTML]{757171}\textbf{60.47} & \color[HTML]{757171}\textbf{97.03} & \color[HTML]{757171}\textbf{57.26} & \color[HTML]{757171}\textbf{83.89} & \color[HTML]{757171}\textbf{87.38} & \color[HTML]{757171}\textbf{65.52} & \color[HTML]{757171}\textbf{40.69} & \color[HTML]{757171}\textbf{42.65} & \color[HTML]{757171}\textbf{20.16} & \color[HTML]{757171}\textbf{59.15} & \color[HTML]{757171}\textbf{85.33} & \color[HTML]{757171}\textbf{81.73} & \color[HTML]{757171}\textbf{51.98} & \color[HTML]{757171}\textbf{52.08} & \color[HTML]{757171}\textbf{64.46} & \\
  & $\tau^\prime=0.01$ & {70.06}& {70.33}& 39.38  & 86.74  & 37.22& 31.12 & 63.78   & 35.16& 25.43& 16.87 & 4.89 & 34.23  & 73.42& 60.46& 20.56& 49.99 & 44.98& \\
  & $\tau^\prime=0.03$ & 69.38& 64.18 & 36.23  & 86.56  & 36.74& 31.13 & 62.50   & 35.52& 25.75& 14.24 & 5.13 & 33.86  & 73.30& 59.63& 20.15& 49.92 & 44.01& \\
  & \textbf{$\tau^\prime=0.05$}  & {\ul 72.68}& 71.18 & 42.52  & 88.33  & 36.36& 33.91 & 66.78   & 37.10& 26.44& 16.86 & 5.85 & 35.98  & 76.29& 61.66& 24.69& 49.83 & 46.65& \\
\rowcolor[HTML]{faebd7}
{\cellcolor[HTML]{FFFFFF}\multirow{-6}{*}{}}   & {\textbf{$\tau^\prime=0.07$ (Ours)}}& \textbf{74.46}& {81.81}   & {54.45} & 91.88  & 41.06& 53.58 & 74.16   & 47.57& 31.92& \textbf{19.29}& 10.95& 43.20  & \textbf{82.39}   & {\ul 71.53} & 37.32& 51.20 & 54.17& \\
  & $\tau^\prime=0.09$ & 70.66& {\ul 83.79}   & {\ul 56.44}& \textbf{92.23}& {\ul 42.10}  & \textbf{58.26}& {\ul 75.88}& \textbf{49.16}   & \textbf{33.56}& {\ul 16.87}   & \textbf{11.67}  & \textbf{43.62} & {\ul 82.07} & \textbf{72.96}   & \textbf{38.48}& \textbf{56.23} & \textbf{55.25}& \\
{\cellcolor[HTML]{FFFFFF}\multirow{-7}{*}{\rotatebox{90}{Clean}}}  & $\tau^\prime=0.10$  & 70.40& \textbf{85.27}& \textbf{57.27} & {\ul 92.16} & \textbf{42.51} & {\ul 57.85}   & \textbf{76.34}  & {\ul 48.97} & {\ul 32.93} & 15.28 & {\ul 11.19}  & {\ul 43.56}& 81.78& 72.90& {\ul 38.43}& {\ul 56.08}& {\ul 55.18} & \\  \midrule
 
\rowcolor[HTML]{EDEDED}  
{\cellcolor[HTML]{FFFFFF}\multirow{-6}{*}{}}  
   & \color[HTML]{757171}\textbf{CLIP}~\citep{radford2021learning} & \color[HTML]{757171}\textbf{1.26} & \color[HTML]{757171}\textbf{6.47} & \color[HTML]{757171}\textbf{0.33} & \color[HTML]{757171}\textbf{30.70} & \color[HTML]{757171}\textbf{0.70} & \color[HTML]{757171}\textbf{9.73} & \color[HTML]{757171}\textbf{4.80} & \color[HTML]{757171}\textbf{1.11} & \color[HTML]{757171}\textbf{0.11} & \color[HTML]{757171}\textbf{0.10} & \color[HTML]{757171}\textbf{0.00} & \color[HTML]{757171}\textbf{1.00} & \color[HTML]{757171}\textbf{19.11} & \color[HTML]{757171}\textbf{13.56} & \color[HTML]{757171}\textbf{0.37} & \color[HTML]{757171}\textbf{0.00} & \color[HTML]{757171}\textbf{5.58} & \color[HTML]{757171}\textbf{10.28} \\
  & $\tau^\prime=0.01$ & 45.98& 37.00 & 20.52  & 69.35  & 18.07& 12.52 & 37.80   & 20.31& 16.86& \textbf{11.53}& 1.74 & 17.21  & 54.62& 41.67& 7.82 & \textbf{49.16} & 28.89& 35.18   \\
  & $\tau^\prime=0.03$ & 46.30& 33.71 & 18.71  & 68.74  & {\ul 18.08}  & 13.20 & 37.78   & 20.61& 16.70& 11.29 & 1.77 & 17.41  & 55.72& 41.57& 7.46 & {\ul 46.95}& 28.50& 34.60   \\
  & $\tau^\prime=0.05$ & \textbf{48.48}& 37.96 & {\ul 21.59}& 70.98  & 17.72& 14.17 & 40.07   & 20.26& 17.23& {\ul 11.42}   & 1.44 & 18.35  & 58.42& 42.95& 9.59 & 45.69 & 29.77& 36.35   \\
\rowcolor[HTML]{faebd7}
{\cellcolor[HTML]{FFFFFF}\multirow{-6}{*}{}}  & \textbf{$\tau^\prime=0.07$ (Ours)}  & {\ul 45.80}& \textbf{42.32}& \textbf{23.03} & \textbf{73.15}& \textbf{18.26} & 20.52 & \textbf{44.02}  & {\ul 24.54} & \textbf{18.14}& 2.26  & \textbf{2.61}& \textbf{20.15} & \textbf{63.73}   & \textbf{48.66}   & \textbf{12.60}& 29.51 & \textbf{30.58}& \textbf{39.09}  \\
  & $\tau^\prime=0.09$ & 38.58& {\ul 41.00}   & 21.23  & {\ul 71.83} & 16.93& \textbf{21.60}& {\ul 40.67}& \textbf{25.08}   & {\ul 17.87} & 1.58  & {\ul 2.43}   & {\ul 18.98}& {\ul 63.36} & {\ul 47.54} & {\ul 11.37}& 23.08 & {\ul 28.94} & {\ul 37.99} \\
{\cellcolor[HTML]{FFFFFF}\multirow{-7}{*}{\rotatebox{90}{Auto Attack}}}  
 & $\tau^\prime=0.10$  & 37.64& 40.75 & 19.94  & 70.90  & 16.06& {\ul 20.57}   & 40.77   & 24.15& 17.18& 1.24  & 2.28 & 18.04  & 61.53& 46.28& 10.96& 17.82 & 27.88& 37.04  \\
\bottomrule
\end{tabular}}
\label{tab:re-tauprime}
\vspace{-1em}
\end{table}

%% file: Tab/rebuttal-VLM.tex
\begin{table}[]
\centering
\caption{\textbf{Zero-shot adversarial robustness on different vision–language models (VLMs).}
All methods are fine-tuned on TinyImageNet following TeCoA~\citep{yu2024text}, adversarial training uses 2-step PGD~\citep{madry2017towards} with $\epsilon\!=\!1/255$. \textit{Average} is the mean across datasets. \textit{H} is the harmonic mean between Clean and the corresponding robust score. 
% Best and second-best are in \textbf{bold} and \underline{underline}.
}
\resizebox{\textwidth}{!}{
\begin{tabular}{ll|*{16}{r}|rr} 
% \begin{tabular}{>{\color{blue}}l>{\color{blue}}l|*{16}{>{\color{blue}}r}|>{\color{blue}}r>{\color{blue}}r} 
\toprule

\multicolumn{2}{l|}{Methods} &
\multicolumn{1}{c}{\rotatebox{90}{TinyImageNet}} &
\multicolumn{1}{c}{\rotatebox{90}{CIFAR-10}} &
\multicolumn{1}{c}{\rotatebox{90}{CIFAR-100}} &
\multicolumn{1}{c}{\rotatebox{90}{STL10}} &
\multicolumn{1}{c}{\rotatebox{90}{SUN397}} &
\multicolumn{1}{c}{\rotatebox{90}{Food101}} &
\multicolumn{1}{c}{\rotatebox{90}{Oxfordpets}} &
\multicolumn{1}{c}{\rotatebox{90}{Flowers102}} &
\multicolumn{1}{c}{\rotatebox{90}{DTD}} &
\multicolumn{1}{c}{\rotatebox{90}{EuroSAT}} &
\multicolumn{1}{c}{\rotatebox{90}{FGVC Aircraft}} &
\multicolumn{1}{c}{\rotatebox{90}{ImageNet}} &
\multicolumn{1}{c}{\rotatebox{90}{Caltech101}} &
\multicolumn{1}{c}{\rotatebox{90}{Caltech256}} &
\multicolumn{1}{c}{\rotatebox{90}{StanfordCars}} &
\multicolumn{1}{c}{\rotatebox{90}{PCAM}} &
\multicolumn{1}{|c}{\rotatebox{90}{Average}} &
\multicolumn{1}{c}{\rotatebox{90}{H}} \\

% \multicolumn{2}{>{\color{blue}}l|}{Methods} &
% \multicolumn{1}{>{\color{blue}}c}{\rotatebox{90}{TinyImageNet}} &
% \multicolumn{1}{>{\color{blue}}c}{\rotatebox{90}{CIFAR-10}} &
% \multicolumn{1}{>{\color{blue}}c}{\rotatebox{90}{CIFAR-100}} &
% \multicolumn{1}{>{\color{blue}}c}{\rotatebox{90}{STL10}} &
% \multicolumn{1}{>{\color{blue}}c}{\rotatebox{90}{SUN397}} &
% \multicolumn{1}{>{\color{blue}}c}{\rotatebox{90}{Food101}} &
% \multicolumn{1}{>{\color{blue}}c}{\rotatebox{90}{Oxfordpets}} &
% \multicolumn{1}{>{\color{blue}}c}{\rotatebox{90}{Flowers102}} &
% \multicolumn{1}{>{\color{blue}}c}{\rotatebox{90}{DTD}} &
% \multicolumn{1}{>{\color{blue}}c}{\rotatebox{90}{EuroSAT}} &
% \multicolumn{1}{>{\color{blue}}c}{\rotatebox{90}{FGVC Aircraft}} &
% \multicolumn{1}{>{\color{blue}}c}{\rotatebox{90}{ImageNet}} &
% \multicolumn{1}{>{\color{blue}}c}{\rotatebox{90}{Caltech101}} &
% \multicolumn{1}{>{\color{blue}}c}{\rotatebox{90}{Caltech256}} &
% \multicolumn{1}{>{\color{blue}}c}{\rotatebox{90}{StanfordCars}} &
% \multicolumn{1}{>{\color{blue}}c}{\rotatebox{90}{PCAM}} &
% \multicolumn{1}{|>{\color{blue}}c}{\rotatebox{90}{Average}} &
% \multicolumn{1}{>{\color{blue}}c}{\rotatebox{90}{H}} \\
\midrule

CLIP-B/16~\citep{radford2021learning}& Clean& 60.86& 89.49& 66.25& 98.03& 61.05 & 88.51& 88.66 & 70.97& 43.03& 45.83& 24.57& 0.07 & 86.07& 84.99& 58.29& 52.86 & 63.72&\\
& AutoAttack& 0.00 & 0.00& 0.03 & 0.00 & 0.01& 0.00& 0.00& 0.02 & 0.00 & 0.08& 0.06& 0.00 & 0.01 & 0.00 & 0.03 & 0.00& 0.01 & 0.03 \\
\rowcolor[HTML]{faebd7}
{\cellcolor[HTML]{FFFFFF}{+ UCAT (Ours)}} & Clean& 77.52& 82.02& 57.34& 92.66& 43.78 & 54.76& 73.75 & 49.86& 30.16& 24.25& 12.39& 0.06 & 82.21& 73.72& 37.51& 54.50 & 52.91&\\
\rowcolor[HTML]{faebd7}
{\cellcolor[HTML]{FFFFFF}\multirow{-2}{*}{}} & AutoAttack& 50.90& 46.15& 26.05& 76.83& 19.74 & 21.56& 45.08 & 27.13& 17.66& 2.37& 3.84& 0.00 & 65.09& 52.64& 13.02& 20.65 & 30.54& 39.05\\ \midrule

SLIP-B/16~\citep{mu2022slip}& Clean& 36.74& 78.97& 44.22& 94.28& 52.71 & 59.70& 31.67 & 59.67& 21.65& 19.87& 5.79& 38.61& 75.15& 62.71& 5.85 & 48.96 & 46.03&\\
& AutoAttack& 0.04 & 0.00& 0.02 & 0.01 & 0.02& 0.03& 0.00& 0.00 & 0.11 & 0.04& 0.00& 0.02 & 0.04 & 0.03 & 0.00 & 0.01& 0.02 & 0.05 \\
\rowcolor[HTML]{faebd7}
{\cellcolor[HTML]{FFFFFF}{+ UCAT (Ours)}} & Clean& 51.90& 70.14& 38.51& 86.76& 41.86 & 26.59& 25.29 & 38.19& 16.22& 13.50& 3.75& 26.42& 70.12& 51.07& 3.37 & 50.29 & 38.37&\\
\rowcolor[HTML]{faebd7}
{\cellcolor[HTML]{FFFFFF}\multirow{-2}{*}{}} & AutoAttack& 25.52& 34.97& 16.04& 66.73& 19.13 & 8.77& 7.39& 16.34& 8.09 & 1.23& 0.75& 11.58& 49.03& 29.35& 0.57 & 30.92 & 20.40& 26.68 \\ \midrule

CLIP-B/32~\citep{radford2021learning}& Clean& 57.27& 88.05& 60.47& 97.04& 57.27 & 83.89& 87.35 & 65.52& 40.80& 42.50& 20.13& 59.15& 85.33& 81.72& 52.02& 52.24 & 64.42&\\
& AutoAttack& 1.26 & 6.47& 0.33 & 30.70& 0.70& 9.73& 4.80& 1.11 & 0.11 & 0.10& 0.00& 1.00 & 19.11& 13.56& 0.37 & 0.00& 5.58 & 10.28\\
\rowcolor[HTML]{faebd7}
{\cellcolor[HTML]{FFFFFF}{+ UCAT (Ours)}} & Clean& 74.46& 81.81& 54.45& 91.88& 41.06 & 53.58& 74.16 & 47.57& 31.92& 19.29& 10.95& 43.20& 82.39& 71.53& 37.32& 51.20 & 54.17&\\
\rowcolor[HTML]{faebd7}
{\cellcolor[HTML]{FFFFFF}\multirow{-2}{*}{}} & AutoAttack& 45.80& 42.32& 23.03& 73.15& 18.26 & 20.52& 44.02 & 24.54& 18.14& 2.26& 2.61& 20.15& 63.73& 48.66& 12.60& 29.51 & 30.58& 39.09\\
\bottomrule
\end{tabular}}
\label{tab:re-vlm}
\vspace{-1em}
\end{table}

%% file: Tab/rebuttal-at.tex
\begin{table}[!ht]
\caption{\textbf{Comparison with representative adversarial training methods zero-shot adversarial robustness (ZSAR) setting.}
All methods are fine-tuned on TinyImageNet following TeCoA~\citep{yu2024text}, adversarial training uses 2-step PGD~\citep{madry2017towards} with $\epsilon\!=\!1/255$. \textit{Average} is the mean across datasets. \textit{H} is the harmonic mean between Clean and the corresponding robust score. Best and second-best are in \textbf{bold} and \underline{underline}.}
\resizebox{\textwidth}{!}{
\begin{tabular}{ll|*{16}{r}|rr} 
% \begin{tabular}{>{\color{blue}}l>{\color{blue}}l|*{16}{>{\color{blue}}r}|>{\color{blue}}r>{\color{blue}}r} 
\toprule

% \multicolumn{2}{>{\color{blue}}l|}{Methods} &
% \multicolumn{1}{>{\color{blue}}c}{\rotatebox{90}{TinyImageNet}} &
% \multicolumn{1}{>{\color{blue}}c}{\rotatebox{90}{CIFAR-10}} &
% \multicolumn{1}{>{\color{blue}}c}{\rotatebox{90}{CIFAR-100}} &
% \multicolumn{1}{>{\color{blue}}c}{\rotatebox{90}{STL10}} &
% \multicolumn{1}{>{\color{blue}}c}{\rotatebox{90}{SUN397}} &
% \multicolumn{1}{>{\color{blue}}c}{\rotatebox{90}{Food101}} &
% \multicolumn{1}{>{\color{blue}}c}{\rotatebox{90}{Oxfordpets}} &
% \multicolumn{1}{>{\color{blue}}c}{\rotatebox{90}{Flowers102}} &
% \multicolumn{1}{>{\color{blue}}c}{\rotatebox{90}{DTD}} &
% \multicolumn{1}{>{\color{blue}}c}{\rotatebox{90}{EuroSAT}} &
% \multicolumn{1}{>{\color{blue}}c}{\rotatebox{90}{FGVC Aircraft}} &
% \multicolumn{1}{>{\color{blue}}c}{\rotatebox{90}{ImageNet}} &
% \multicolumn{1}{>{\color{blue}}c}{\rotatebox{90}{Caltech101}} &
% \multicolumn{1}{>{\color{blue}}c}{\rotatebox{90}{Caltech256}} &
% \multicolumn{1}{>{\color{blue}}c}{\rotatebox{90}{StanfordCars}} &
% \multicolumn{1}{>{\color{blue}}c}{\rotatebox{90}{PCAM}} &
% \multicolumn{1}{|>{\color{blue}}c}{\rotatebox{90}{Average}} &
% \multicolumn{1}{>{\color{blue}}c}{\rotatebox{90}{H}} \\
\multicolumn{2}{l|}{Methods} &
\multicolumn{1}{c}{\rotatebox{90}{TinyImageNet}} &
\multicolumn{1}{c}{\rotatebox{90}{CIFAR-10}} &
\multicolumn{1}{c}{\rotatebox{90}{CIFAR-100}} &
\multicolumn{1}{c}{\rotatebox{90}{STL10}} &
\multicolumn{1}{c}{\rotatebox{90}{SUN397}} &
\multicolumn{1}{c}{\rotatebox{90}{Food101}} &
\multicolumn{1}{c}{\rotatebox{90}{Oxfordpets}} &
\multicolumn{1}{c}{\rotatebox{90}{Flowers102}} &
\multicolumn{1}{c}{\rotatebox{90}{DTD}} &
\multicolumn{1}{c}{\rotatebox{90}{EuroSAT}} &
\multicolumn{1}{c}{\rotatebox{90}{FGVC Aircraft}} &
\multicolumn{1}{c}{\rotatebox{90}{ImageNet}} &
\multicolumn{1}{c}{\rotatebox{90}{Caltech101}} &
\multicolumn{1}{c}{\rotatebox{90}{Caltech256}} &
\multicolumn{1}{c}{\rotatebox{90}{StanfordCars}} &
\multicolumn{1}{c}{\rotatebox{90}{PCAM}} &
\multicolumn{1}{|c}{\rotatebox{90}{Average}} &
\multicolumn{1}{c}{\rotatebox{90}{H}} \\
\midrule

\rowcolor[HTML]{EDEDED}
{\cellcolor[HTML]{FFFFFF}\multirow{-6}{*}{}}   & \color[HTML]{757171}\textbf{CLIP}~\citep{radford2021learning}   & \color[HTML]{757171}\textbf{57.96} & \color[HTML]{757171}\textbf{88.02} & \color[HTML]{757171}\textbf{60.47} & \color[HTML]{757171}\textbf{97.03} & \color[HTML]{757171}\textbf{57.26} & \color[HTML]{757171}\textbf{83.89} & \color[HTML]{757171}\textbf{87.38} & \color[HTML]{757171}\textbf{65.52} & \color[HTML]{757171}\textbf{40.69} & \color[HTML]{757171}\textbf{42.65} & \color[HTML]{757171}\textbf{20.16} & \color[HTML]{757171}\textbf{59.15} & \color[HTML]{757171}\textbf{85.33} & \color[HTML]{757171}\textbf{81.73} & \color[HTML]{757171}\textbf{51.98} & \color[HTML]{757171}\textbf{52.08} & \color[HTML]{757171}\textbf{64.46} & \\
 & TRADES~\citep{zhang2019theoretically}& 67.73 & 62.05  & 34.38   & 80.81& 26.64 & {\ul 22.68}& 54.43& 24.48& 20.59  & {\ul 16.51}& 3.60& 26.31   & 63.77& 50.89& 14.59 & 49.95   & 38.71  &  \\
 & ACAT~\citep{addepalli2022efficient}  & {\ul 72.80} & 64.72  & 34.71   & 82.41& 30.06 & 19.13  & {\ul 60.40}  & {\ul 26.80}   & 17.29  & 15.62  & {\ul 4.41}& {\ul 28.60} & 64.94& 50.89& {\ul 15.82} & {\ul 50.01}& {\ul 39.91}   &  \\
 & DKL~\citep{cui2024decoupled}   & 70.84 & {\ul 65.31}& {\ul 35.61} & {\ul 82.54} & {\ul 30.11}  & 21.11  & 55.11& 25.94& {\ul 21.17}   & 16.26  & 3.96& 27.18   & {\ul 65.84}   & {\ul 52.00}   & 14.20 & 48.97   & 39.76  &  \\

\rowcolor[HTML]{faebd7}
{\cellcolor[HTML]{FFFFFF}\multirow{-5}{*}{\rotatebox{90}{Clean}}}  & UCAT (Ours)  & \textbf{74.46}  & \textbf{81.81}& \textbf{54.45} & \textbf{91.88}  & \textbf{41.06}   & \textbf{53.58}& \textbf{74.16}  & \textbf{47.57}& \textbf{31.92}& \textbf{19.29}& \textbf{10.95}& \textbf{43.20} & \textbf{82.39}& \textbf{71.53}& \textbf{37.32}  & \textbf{51.20} & \textbf{54.17}&  \\ \midrule

\rowcolor[HTML]{EDEDED}
{\cellcolor[HTML]{FFFFFF}\multirow{-5}{*}{}}   & \color[HTML]{757171}\textbf{CLIP}~\citep{radford2021learning}   & \color[HTML]{757171}\textbf{1.26} & \color[HTML]{757171}\textbf{6.47} & \color[HTML]{757171}\textbf{0.33} & \color[HTML]{757171}\textbf{30.70} & \color[HTML]{757171}\textbf{0.70} & \color[HTML]{757171}\textbf{9.73} & \color[HTML]{757171}\textbf{4.80} & \color[HTML]{757171}\textbf{1.11} & \color[HTML]{757171}\textbf{0.11} & \color[HTML]{757171}\textbf{0.10} & \color[HTML]{757171}\textbf{0.00} & \color[HTML]{757171}\textbf{1.00} & \color[HTML]{757171}\textbf{19.11} & \color[HTML]{757171}\textbf{13.56} & \color[HTML]{757171}\textbf{0.37} & \color[HTML]{757171}\textbf{0.00} & \color[HTML]{757171}\textbf{5.58} & \color[HTML]{757171}\textbf{10.28} \\
 & TRADES~\citep{zhang2019theoretically}& {\ul 52.19} & 41.39  & 22.44   & 67.59& 15.69 & {\ul 12.50}& {\ul 37.42}  & 16.46& {\ul 15.75}   & \textbf{11.77}& 1.23& 16.16   & 51.16& 37.84& 7.29  & {\ul 46.42}& {\ul 28.33}   & 32.72\\
 & ACAT~\citep{addepalli2022efficient}  & 51.85 & 34.54  & 16.85   & 64.84& 14.99 & 8.95   & 36.71& {\ul 16.56}   & 10.96  & 11.41  & 1.59& 15.69   & 48.70& 34.84& {\ul 7.62}  & \textbf{49.89} & 26.62  & 31.94\\
 & DKL~\citep{cui2024decoupled}    & \textbf{{54.71}} & {\ul 41.77}& {\ul 22.29} & {\ul 67.84} & {\ul 17.32}  & 10.73  & 36.06& 16.23& 15.69  & {\ul 11.72}& {\ul 1.65}& {\ul 16.66} & {\ul 51.85}   & {\ul 38.33}   & 6.67  & 43.48   & 28.31  & {\ul 33.07}\\

\rowcolor[HTML]{faebd7}
{\cellcolor[HTML]{FFFFFF}\multirow{-5}{*}{\rotatebox{90}{AutoAttack}}}  & UCAT (Ours)  & {45.80}  & \textbf{42.32}& \textbf{23.03} & \textbf{73.15}  & \textbf{18.26}   & \textbf{20.52}& \textbf{44.02}  & \textbf{24.54}& \textbf{18.14}& 2.26   & \textbf{2.61} & \textbf{20.15} & \textbf{63.73}& \textbf{48.66}& \textbf{12.60}  & 29.51   & \textbf{30.58}& \textbf{39.09} \\
\bottomrule
\end{tabular}}
\label{tab:re-at}
\vspace{-1em}
\end{table}

%% file: ref.bib
@inproceedings{radford2021learning,
  title={Learning transferable visual models from natural language supervision},
  author={Radford, Alec and Kim, Jong Wook and Hallacy, Chris and Ramesh, Aditya and Goh, Gabriel and Agarwal, Sandhini and Sastry, Girish and Askell, Amanda and Mishkin, Pamela and Clark, Jack and others},
  booktitle={International conference on machine learning},
  pages={8748--8763},
  year={2021},
  organization={PmLR}
}

@inproceedings{wang2021understanding,
  title={Understanding the behaviour of contrastive loss},
  author={Wang, Feng and Liu, Huaping},
  booktitle={Proceedings of the IEEE/CVF conference on computer vision and pattern recognition},
  pages={2495--2504},
  year={2021}
}

@inproceedings{zhai2022lit,
  title={Lit: Zero-shot transfer with locked-image text tuning},
  author={Zhai, Xiaohua and Wang, Xiao and Mustafa, Basil and Steiner, Andreas and Keysers, Daniel and Kolesnikov, Alexander and Beyer, Lucas},
  booktitle={Proceedings of the IEEE/CVF conference on computer vision and pattern recognition},
  pages={18123--18133},
  year={2022}
}

@inproceedings{wang2020understanding,
  title={Understanding contrastive representation learning through alignment and uniformity on the hypersphere},
  author={Wang, Tongzhou and Isola, Phillip},
  booktitle={International conference on machine learning},
  pages={9929--9939},
  year={2020},
  organization={PMLR}
}

@inproceedings{jia2021scaling,
  title={Scaling up visual and vision-language representation learning with noisy text supervision},
  author={Jia, Chao and Yang, Yinfei and Xia, Ye and Chen, Yi-Ting and Parekh, Zarana and Pham, Hieu and Le, Quoc and Sung, Yun-Hsuan and Li, Zhen and Duerig, Tom},
  booktitle={International conference on machine learning},
  pages={4904--4916},
  year={2021},
  organization={PMLR}
}

@inproceedings{wortsman2022robust,
  title={Robust fine-tuning of zero-shot models},
  author={Wortsman, Mitchell and Ilharco, Gabriel and Kim, Jong Wook and Li, Mike and Kornblith, Simon and Roelofs, Rebecca and Lopes, Raphael Gontijo and Hajishirzi, Hannaneh and Farhadi, Ali and Namkoong, Hongseok and others},
  booktitle={Proceedings of the IEEE/CVF conference on computer vision and pattern recognition},
  pages={7959--7971},
  year={2022}
}

@inproceedings{he2020momentum,
  title={Momentum contrast for unsupervised visual representation learning},
  author={He, Kaiming and Fan, Haoqi and Wu, Yuxin and Xie, Saining and Girshick, Ross},
  booktitle={Proceedings of the IEEE/CVF conference on computer vision and pattern recognition},
  pages={9729--9738},
  year={2020}
}

@inproceedings{wu2018unsupervised,
  title={Unsupervised feature learning via non-parametric instance discrimination},
  author={Wu, Zhirong and Xiong, Yuanjun and Yu, Stella X and Lin, Dahua},
  booktitle={Proceedings of the IEEE conference on computer vision and pattern recognition},
  pages={3733--3742},
  year={2018}
}

@inproceedings{yeh2022decoupled,
  title={Decoupled contrastive learning},
  author={Yeh, Chun-Hsiao and Hong, Cheng-Yao and Hsu, Yen-Chi and Liu, Tyng-Luh and Chen, Yubei and LeCun, Yann},
  booktitle={European conference on computer vision},
  pages={668--684},
  year={2022},
  organization={Springer}
}

@article{yao2021filip,
  title={Filip: Fine-grained interactive language-image pre-training},
  author={Yao, Lewei and Huang, Runhui and Hou, Lu and Lu, Guansong and Niu, Minzhe and Xu, Hang and Liang, Xiaodan and Li, Zhenguo and Jiang, Xin and Xu, Chunjing},
  journal={arXiv preprint arXiv:2111.07783},
  year={2021}
}

@article{zhou2022learning,
  title={Learning to prompt for vision-language models},
  author={Zhou, Kaiyang and Yang, Jingkang and Loy, Chen Change and Liu, Ziwei},
  journal={International Journal of Computer Vision},
  volume={130},
  number={9},
  pages={2337--2348},
  year={2022},
  publisher={Springer}
}

@article{yu2024text,
  title={Text-guided attention is all you need for zero-shot robustness in vision-language models},
  author={Yu, Lu and Zhang, Haiyang and Xu, Changsheng},
  journal={Advances in Neural Information Processing Systems},
  volume={37},
  pages={96424--96448},
  year={2024}
}

@article{mao2022understanding,
  title={Understanding zero-shot adversarial robustness for large-scale models},
  author={Mao, Chengzhi and Geng, Scott and Yang, Junfeng and Wang, Xin and Vondrick, Carl},
  journal={arXiv preprint arXiv:2212.07016},
  year={2022}
}

@article{schlarmann2024robust,
  title={Robust clip: Unsupervised adversarial fine-tuning of vision embeddings for robust large vision-language models},
  author={Schlarmann, Christian and Singh, Naman Deep and Croce, Francesco and Hein, Matthias},
  journal={arXiv preprint arXiv:2402.12336},
  year={2024}
}

@inproceedings{li2024language,
  title={Language-driven anchors for zero-shot adversarial robustness},
  author={Li, Xiao and Zhang, Wei and Liu, Yining and Hu, Zhanhao and Zhang, Bo and Hu, Xiaolin},
  booktitle={Proceedings of the IEEE/CVF Conference on Computer Vision and Pattern Recognition},
  pages={24686--24695},
  year={2024}
}

@inproceedings{xing2025clip,
  title={Clip is strong enough to fight back: Test-time counterattacks towards zero-shot adversarial robustness of clip},
  author={Xing, Songlong and Zhao, Zhengyu and Sebe, Nicu},
  booktitle={Proceedings of the Computer Vision and Pattern Recognition Conference},
  pages={15172--15182},
  year={2025}
}

@article{zhang2025clipure,
  title={Clipure: Purification in latent space via clip for adversarially robust zero-shot classification},
  author={Zhang, Mingkun and Bi, Keping and Chen, Wei and Guo, Jiafeng and Cheng, Xueqi},
  journal={arXiv preprint arXiv:2502.18176},
  year={2025}
}

@inproceedings{wang2024pre,
  title={Pre-trained model guided fine-tuning for zero-shot adversarial robustness},
  author={Wang, Sibo and Zhang, Jie and Yuan, Zheng and Shan, Shiguang},
  booktitle={Proceedings of the IEEE/CVF conference on computer vision and pattern recognition},
  pages={24502--24511},
  year={2024}
}

@article{shu2022test,
  title={Test-time prompt tuning for zero-shot generalization in vision-language models},
  author={Shu, Manli and Nie, Weili and Huang, De-An and Yu, Zhiding and Goldstein, Tom and Anandkumar, Anima and Xiao, Chaowei},
  journal={Advances in Neural Information Processing Systems},
  volume={35},
  pages={14274--14289},
  year={2022}
}

@article{pang2019rethinking,
  title={Rethinking softmax cross-entropy loss for adversarial robustness},
  author={Pang, Tianyu and Xu, Kun and Dong, Yinpeng and Du, Chao and Chen, Ning and Zhu, Jun},
  journal={arXiv preprint arXiv:1905.10626},
  year={2019}
}

@inproceedings{zhang2019theoretically,
  title={Theoretically principled trade-off between robustness and accuracy},
  author={Zhang, Hongyang and Yu, Yaodong and Jiao, Jiantao and Xing, Eric and El Ghaoui, Laurent and Jordan, Michael},
  booktitle={International conference on machine learning},
  pages={7472--7482},
  year={2019},
  organization={PMLR}
}

@article{telgarsky2013dirichlet,
  title={Dirichlet draws are sparse with high probability},
  author={Telgarsky, Matus},
  journal={arXiv preprint arXiv:1301.4917},
  year={2013}
}

@misc{minka2000estimating,
  title={Estimating a Dirichlet distribution},
  author={Minka, Thomas},
  year={2000},
  publisher={Technical report, MIT}
}

@article{KIUREGHIAN2009105,
	author = {Armen Der Kiureghian and Ove Ditlevsen},
	issn = {0167-4730},
	journal = {Structural Safety},
	number = {2},
	pages = {105-112},
	title = {{Aleatory or epistemic? Does it matter?}},
	volume = {31},
	year = {2009}}

@article{kendall2017uncertainties,
  title={What uncertainties do we need in bayesian deep learning for computer vision?},
  author={Kendall, Alex and Gal, Yarin},
  journal={Advances in neural information processing systems},
  volume={30},
  year={2017}
}

@article{hullermeier2021aleatoric,
  title={Aleatoric and epistemic uncertainty in machine learning: An introduction to concepts and methods},
  author={H{\"u}llermeier, Eyke and Waegeman, Willem},
  journal={Machine learning},
  volume={110},
  number={3},
  pages={457--506},
  year={2021},
  publisher={Springer}
}

@article{charpentier2020posterior,
  title={Posterior network: Uncertainty estimation without ood samples via density-based pseudo-counts},
  author={Charpentier, Bertrand and Z{\"u}gner, Daniel and G{\"u}nnemann, Stephan},
  journal={Advances in neural information processing systems},
  volume={33},
  pages={1356--1367},
  year={2020}
}

@inproceedings{sensoy2020uncertainty,
  title={Uncertainty-aware deep classifiers using generative models},
  author={Sensoy, Murat and Kaplan, Lance and Cerutti, Federico and Saleki, Maryam},
  booktitle={Proceedings of the AAAI conference on artificial intelligence},
  volume={34},
  number={04},
  pages={5620--5627},
  year={2020}
}

@inproceedings{ji2023map,
  title={Map: Multimodal uncertainty-aware vision-language pre-training model},
  author={Ji, Yatai and Wang, Junjie and Gong, Yuan and Zhang, Lin and Zhu, Yanru and Wang, Hongfa and Zhang, Jiaxing and Sakai, Tetsuya and Yang, Yujiu},
  booktitle={Proceedings of the IEEE/CVF conference on computer vision and pattern recognition},
  pages={23262--23271},
  year={2023}
}

@article{malinin2019reverse,
  title={Reverse kl-divergence training of prior networks: Improved uncertainty and adversarial robustness},
  author={Malinin, Andrey and Gales, Mark},
  journal={Advances in neural information processing systems},
  volume={32},
  year={2019}
}

@article{ma2025estimating,
  title={Estimating LLM Uncertainty with Evidence},
  author={Ma, Huan and Chen, Jingdong and Zhou, Joey Tianyi and Wang, Guangyu and Zhang, Changqing},
  journal={arXiv preprint arXiv:2502.00290},
  year={2025}
}

@article{ulmer2021prior,
  title={Prior and posterior networks: A survey on evidential deep learning methods for uncertainty estimation},
  author={Ulmer, Dennis and Hardmeier, Christian and Frellsen, Jes},
  journal={arXiv preprint arXiv:2110.03051},
  year={2021}
}

@article{yoon2024uncertainty,
  title={Uncertainty estimation by density aware evidential deep learning},
  author={Yoon, Taeseong and Kim, Heeyoung},
  journal={arXiv preprint arXiv:2409.08754},
  year={2024}
}

@article{sensoy2018evidential,
  title={Evidential deep learning to quantify classification uncertainty},
  author={Sensoy, Murat and Kaplan, Lance and Kandemir, Melih},
  journal={Advances in neural information processing systems},
  volume={31},
  year={2018}
}

@inproceedings{guo2017calibration,
  title={On calibration of modern neural networks},
  author={Guo, Chuan and Pleiss, Geoff and Sun, Yu and Weinberger, Kilian Q},
  booktitle={International conference on machine learning},
  pages={1321--1330},
  year={2017},
  organization={PMLR}
}

@article{malinin2018predictive,
  title={Predictive uncertainty estimation via prior networks},
  author={Malinin, Andrey and Gales, Mark},
  journal={Advances in neural information processing systems},
  volume={31},
  year={2018}
}

@article{hendrycks2016baseline,
  title={A baseline for detecting misclassified and out-of-distribution examples in neural networks},
  author={Hendrycks, Dan and Gimpel, Kevin},
  journal={arXiv preprint arXiv:1610.02136},
  year={2016}
}

@article{ovadia2019can,
  title={Can you trust your model's uncertainty? evaluating predictive uncertainty under dataset shift},
  author={Ovadia, Yaniv and Fertig, Emily and Ren, Jie and Nado, Zachary and Sculley, David and Nowozin, Sebastian and Dillon, Joshua and Lakshminarayanan, Balaji and Snoek, Jasper},
  journal={Advances in neural information processing systems},
  volume={32},
  year={2019}
}

@article{madry2017towards,
  title={Towards deep learning models resistant to adversarial attacks},
  author={Madry, Aleksander and Makelov, Aleksandar and Schmidt, Ludwig and Tsipras, Dimitris and Vladu, Adrian},
  journal={arXiv preprint arXiv:1706.06083},
  year={2017}
}

@inproceedings{croce2020reliable,
  title={Reliable evaluation of adversarial robustness with an ensemble of diverse parameter-free attacks},
  author={Croce, Francesco and Hein, Matthias},
  booktitle={International conference on machine learning},
  pages={2206--2216},
  year={2020},
  organization={PMLR}
}

@article{goodfellow2014explaining,
  title={Explaining and harnessing adversarial examples},
  author={Goodfellow, Ian J and Shlens, Jonathon and Szegedy, Christian},
  journal={arXiv preprint arXiv:1412.6572},
  year={2014}
}

@inproceedings{carlini2017towards,
  title={Towards evaluating the robustness of neural networks},
  author={Carlini, Nicholas and Wagner, David},
  booktitle={2017 ieee symposium on security and privacy (sp)},
  pages={39--57},
  year={2017},
  organization={Ieee}
}

@incollection{kurakin2018adversarial,
  title={Adversarial examples in the physical world},
  author={Kurakin, Alexey and Goodfellow, Ian J and Bengio, Samy},
  booktitle={Artificial intelligence safety and security},
  pages={99--112},
  year={2018},
  publisher={Chapman and Hall/CRC}
}

@inproceedings{deng2009imagenet,
  title={Imagenet: A large-scale hierarchical image database},
  author={Deng, Jia and Dong, Wei and Socher, Richard and Li, Li-Jia and Li, Kai and Fei-Fei, Li},
  booktitle={2009 IEEE conference on computer vision and pattern recognition},
  pages={248--255},
  year={2009},
  organization={Ieee}
}

@misc{krizhevsky2009learning,
  title={Learning multiple layers of features from tiny images.(2009)},
  author={Krizhevsky, Alex and Hinton, Geoffrey and others},
  year={2009}
}

@inproceedings{coates2011analysis,
  title={An analysis of single-layer networks in unsupervised feature learning},
  author={Coates, Adam and Ng, Andrew and Lee, Honglak},
  booktitle={Proceedings of the fourteenth international conference on artificial intelligence and statistics},
  pages={215--223},
  year={2011},
  organization={JMLR Workshop and Conference Proceedings}
}

@inproceedings{fei2004learning,
  title={Learning generative visual models from few training examples: An incremental bayesian approach tested on 101 object categories},
  author={Fei-Fei, Li and Fergus, Rob and Perona, Pietro},
  booktitle={2004 conference on computer vision and pattern recognition workshop},
  pages={178--178},
  year={2004},
  organization={IEEE}
}

@techreport{griffin2007caltech,
  title={Caltech-256 object category dataset},
  author={Griffin, Gregory and Holub, Alex and Perona, Pietro and others},
  year={2007},
  institution={Technical Report 7694, California Institute of Technology Pasadena}
}

@INPROCEEDINGS{6248092,
  author={{Parkhi, Omkar M and Vedaldi, Andrea and Zisserman, Andrew and Jawahar, C. V.}},
  booktitle={{2012 IEEE Conference on Computer Vision and Pattern Recognition}}, 
  title={{Cats and dogs}}, 
  year={2012},
  volume={},
  number={},
  pages={3498-3505}}

@inproceedings{krause20133d,
  title={3d object representations for fine-grained categorization},
  author={Krause, Jonathan and Stark, Michael and Deng, Jia and Fei-Fei, Li},
  booktitle={Proceedings of the IEEE international conference on computer vision workshops},
  pages={554--561},
  year={2013}
}

@inproceedings{bossard2014food,
  title={Food-101--mining discriminative components with random forests},
  author={Bossard, Lukas and Guillaumin, Matthieu and Van Gool, Luc},
  booktitle={European conference on computer vision},
  pages={446--461},
  year={2014},
  organization={Springer}
}

@inproceedings{nilsback2008automated,
  title={Automated flower classification over a large number of classes},
  author={Nilsback, Maria-Elena and Zisserman, Andrew},
  booktitle={2008 Sixth Indian conference on computer vision, graphics \& image processing},
  pages={722--729},
  year={2008},
  organization={IEEE}
}

@article{maji2013fine,
  title={Fine-grained visual classification of aircraft},
  author={Maji, Subhransu and Rahtu, Esa and Kannala, Juho and Blaschko, Matthew and Vedaldi, Andrea},
  journal={arXiv preprint arXiv:1306.5151},
  year={2013}
}

@inproceedings{xiao2010sun,
  title={Sun database: Large-scale scene recognition from abbey to zoo},
  author={Xiao, Jianxiong and Hays, James and Ehinger, Krista A and Oliva, Aude and Torralba, Antonio},
  booktitle={2010 IEEE computer society conference on computer vision and pattern recognition},
  pages={3485--3492},
  year={2010},
  organization={IEEE}
}

@inproceedings{cimpoi2014describing,
  title={Describing textures in the wild},
  author={Cimpoi, Mircea and Maji, Subhransu and Kokkinos, Iasonas and Mohamed, Sammy and Vedaldi, Andrea},
  booktitle={Proceedings of the IEEE conference on computer vision and pattern recognition},
  pages={3606--3613},
  year={2014}
}

@inproceedings{veeling2018rotation,
  title={Rotation equivariant CNNs for digital pathology},
  author={Veeling, Bastiaan S and Linmans, Jasper and Winkens, Jim and Cohen, Taco and Welling, Max},
  booktitle={International Conference on Medical image computing and computer-assisted intervention},
  pages={210--218},
  year={2018},
  organization={Springer}
}

@article{helber2019eurosat,
  title={Eurosat: A novel dataset and deep learning benchmark for land use and land cover classification},
  author={Helber, Patrick and Bischke, Benjamin and Dengel, Andreas and Borth, Damian},
  journal={IEEE Journal of Selected Topics in Applied Earth Observations and Remote Sensing},
  volume={12},
  number={7},
  pages={2217--2226},
  year={2019},
  publisher={IEEE}
}

@inproceedings{lin2014microsoft,
  title={Microsoft coco: Common objects in context},
  author={Lin, Tsung-Yi and Maire, Michael and Belongie, Serge and Hays, James and Perona, Pietro and Ramanan, Deva and Doll{\'a}r, Piotr and Zitnick, C Lawrence},
  booktitle={European conference on computer vision},
  pages={740--755},
  year={2014},
  organization={Springer}
}

@inproceedings{mao2021composite,
  title={Composite adversarial attacks},
  author={Mao, Xiaofeng and Chen, Yuefeng and Wang, Shuhui and Su, Hang and He, Yuan and Xue, Hui},
  booktitle={Proceedings of the AAAI conference on artificial intelligence},
  volume={35},
  number={10},
  pages={8884--8892},
  year={2021}
}

@inproceedings{liu2022practical,
  title={Practical evaluation of adversarial robustness via adaptive auto attack},
  author={Liu, Ye and Cheng, Yaya and Gao, Lianli and Liu, Xianglong and Zhang, Qilong and Song, Jingkuan},
  booktitle={Proceedings of the IEEE/CVF Conference on Computer Vision and Pattern Recognition},
  pages={15105--15114},
  year={2022}
}

@inproceedings{mu2022slip,
  title={Slip: Self-supervision meets language-image pre-training},
  author={Mu, Norman and Kirillov, Alexander and Wagner, David and Xie, Saining},
  booktitle={European conference on computer vision},
  pages={529--544},
  year={2022},
  organization={Springer}
}

@article{addepalli2022efficient,
  title={Efficient and effective augmentation strategy for adversarial training},
  author={Addepalli, Sravanti and Jain, Samyak and others},
  journal={Advances in Neural Information Processing Systems},
  volume={35},
  pages={1488--1501},
  year={2022}
}

@article{cui2024decoupled,
  title={Decoupled kullback-leibler divergence loss},
  author={Cui, Jiequan and Tian, Zhuotao and Zhong, Zhisheng and Qi, Xiaojuan and Yu, Bei and Zhang, Hanwang},
  journal={Advances in Neural Information Processing Systems},
  volume={37},
  pages={74461--74486},
  year={2024}
}

@inproceedings{stutz2020confidence,
  title={Confidence-calibrated adversarial training: Generalizing to unseen attacks},
  author={Stutz, David and Hein, Matthias and Schiele, Bernt},
  booktitle={International conference on machine learning},
  pages={9155--9166},
  year={2020},
  organization={PMLR}
}

@inproceedings{dong2025improving,
  title={Improving Zero-Shot Adversarial Robustness in Vision-Language Models by Closed-form Alignment of Adversarial Path Simplices},
  author={Dong, Junhao and Koniusz, Piotr and Zhang, Yifei and Zhu, Hao and Liu, Weiming and Qu, Xinghua and Ong, Yew-Soon},
  booktitle={Forty-second International Conference on Machine Learning},
  year={2025}
}

@inproceedings{dong2025robustifying,
  title={Robustifying zero-shot vision language models by subspaces alignment},
  author={Dong, Junhao and Koniusz, Piotr and Feng, Liaoyuan and Zhang, Yifei and Zhu, Hao and Liu, Weiming and Qu, Xinghua and Ong, Yew-Soon},
  booktitle={Proceedings of the IEEE/CVF International Conference on Computer Vision},
  pages={21037--21047},
  year={2025}
}

@inproceedings{sheng2025r,
  title={R-TPT: Improving Adversarial Robustness of Vision-Language Models through Test-Time Prompt Tuning},
  author={Sheng, Lijun and Liang, Jian and Wang, Zilei and He, Ran},
  booktitle={Proceedings of the Computer Vision and Pattern Recognition Conference},
  pages={29958--29967},
  year={2025}
}

@inproceedings{li2024one,
  title={One prompt word is enough to boost adversarial robustness for pre-trained vision-language models},
  author={Li, Lin and Guan, Haoyan and Qiu, Jianing and Spratling, Michael},
  booktitle={Proceedings of the IEEE/CVF Conference on Computer Vision and Pattern Recognition},
  pages={24408--24419},
  year={2024}
}

@inproceedings{wang2025tapt,
  title={Tapt: Test-time adversarial prompt tuning for robust inference in vision-language models},
  author={Wang, Xin and Chen, Kai and Zhang, Jiaming and Chen, Jingjing and Ma, Xingjun},
  booktitle={Proceedings of the Computer Vision and Pattern Recognition Conference},
  pages={19910--19920},
  year={2025}
}

@inproceedings{tong2025zero,
  title={On the Zero-shot Adversarial Robustness of Vision-Language Models: A Truly Zero-shot and Training-free Approach},
  author={Tong, Baoshun and Lai, Hanjiang and Pan, Yan and Yin, Jian},
  booktitle={Proceedings of the Computer Vision and Pattern Recognition Conference},
  pages={19921--19930},
  year={2025}
}

@article{zhang2019you,
  title={You only propagate once: Accelerating adversarial training via maximal principle},
  author={Zhang, Dinghuai and Zhang, Tianyuan and Lu, Yiping and Zhu, Zhanxing and Dong, Bin},
  journal={Advances in neural information processing systems},
  volume={32},
  year={2019}
}

@article{shafahi2019adversarial,
  title={Adversarial training for free!},
  author={Shafahi, Ali and Najibi, Mahyar and Ghiasi, Mohammad Amin and Xu, Zheng and Dickerson, John and Studer, Christoph and Davis, Larry S and Taylor, Gavin and Goldstein, Tom},
  journal={Advances in neural information processing systems},
  volume={32},
  year={2019}
}

@article{wong2020fast,
  title={Fast is better than free: Revisiting adversarial training},
  author={Wong, Eric and Rice, Leslie and Kolter, J Zico},
  journal={arXiv preprint arXiv:2001.03994},
  year={2020}
}

@inproceedings{rice2020overfitting,
  title={Overfitting in adversarially robust deep learning},
  author={Rice, Leslie and Wong, Eric and Kolter, Zico},
  booktitle={International conference on machine learning},
  pages={8093--8104},
  year={2020},
  organization={PMLR}
}

@article{wu2020adversarial,
  title={Adversarial weight perturbation helps robust generalization},
  author={Wu, Dongxian and Xia, Shu-Tao and Wang, Yisen},
  journal={Advances in neural information processing systems},
  volume={33},
  pages={2958--2969},
  year={2020}
}

@article{levine2023enabling,
  title={Enabling calibration in the zero-shot inference of large vision-language models},
  author={LeVine, Will and Pikus, Benjamin and Raja, Pranav and Gil, Fernando Amat},
  journal={arXiv preprint arXiv:2303.12748},
  year={2023}
}

@inproceedings{murugesan2024robust,
  title={Robust calibration of large vision-language adapters},
  author={Murugesan, Balamurali and Silva-Rodr{\'\i}guez, Julio and Ayed, Ismail Ben and Dolz, Jose},
  booktitle={European Conference on Computer Vision},
  pages={147--165},
  year={2024},
  organization={Springer}
}

@inproceedings{esmaeilpour2022zero,
  title={Zero-shot out-of-distribution detection based on the pre-trained model clip},
  author={Esmaeilpour, Sepideh and Liu, Bing and Robertson, Eric and Shu, Lei},
  booktitle={Proceedings of the AAAI conference on artificial intelligence},
  volume={36},
  number={6},
  pages={6568--6576},
  year={2022}
}

@inproceedings{wang2019improving,
  title={Improving adversarial robustness requires revisiting misclassified examples},
  author={Wang, Yisen and Zou, Difan and Yi, Jinfeng and Bailey, James and Ma, Xingjun and Gu, Quanquan},
  booktitle={International conference on learning representations},
  year={2019}
}

@article{yu2026complementary,
  title={Complementary Text-Guided Attention for Zero-Shot Adversarial Robustness},
  author={Yu, Lu and Zhang, Haiyang and Xu, Changsheng},
  journal={IEEE Transactions on Pattern Analysis and Machine Intelligence},
  year={2026},
  publisher={IEEE}
}
